\documentclass[11pt, letterpaper, logo, onecolumn, numbering]{rexomni}
\usepackage[all]{hypcap}
\usepackage[numbers]{natbib}
\usepackage{hyperref}[citecolor=lightblue]

\hypersetup{
    colorlinks = true,
    citecolor = {lightblue},
    linkcolor = {lightblue},
    urlcolor = {lightblue}
}

\usepackage{algorithm}
\usepackage{algorithmicx}
\usepackage{algpseudocode}
\usepackage{microtype}
\usepackage{graphicx}
\usepackage{float}
\usepackage{bigstrut}

\usepackage{amsmath}
\usepackage{amssymb}
\usepackage{mathtools}
\usepackage{amsthm}
\usepackage{mathrsfs}
\usepackage{nicefrac}
\usepackage{dsfont}
\usepackage{enumitem}
\usepackage{subcaption}
\usepackage{graphicx}
\usepackage{float}
\usepackage[utf8]{inputenc} %
\usepackage[T1]{fontenc}    %
\usepackage{hyperref}       %
\usepackage{url}            %
\usepackage{booktabs}       %
\usepackage{amsfonts}       %
\usepackage{nicefrac}       %
\usepackage{microtype}      %
\usepackage{fdsymbol}
\usepackage{wrapfig}
\usepackage{lipsum}
\usepackage{stackengine}
\usepackage[font=small,labelfont=bf]{caption}
\usepackage{color}
\usepackage{adjustbox}
\usepackage{rotating}
\usepackage{makecell}

\usepackage{xspace}
\usepackage{tikz}
\usepackage{multirow}    
\usepackage{array}       
\usepackage{colortbl}
\usepackage{tabularx}

\usepackage{pifont}   

\newtcolorbox{AIbox}[2][]{aibox,title=#2,#1}
\definecolor{lightblue}{rgb}{0.22,0.45,0.70}%
\definecolor{Gray}{gray}{0.95}
\definecolor{Cornsilk}{rgb}{1.0, 0.97, 0.86}

\usepackage{amsmath}

\usepackage[all]{hypcap}

\usepackage[capitalize,noabbrev]{cleveref}
\crefname{section}{Section}{\S\S}
\Crefname{section}{Section}{\S\S}
\crefname{table}{Table}{Tables}
\crefname{figure}{Figure}{Figures}
\crefname{algorithm}{Algorithm}{}
\crefname{equation}{eq.}{}
\crefname{appendix}{Appendix}{}
\crefformat{section}{Section #2#1#3}

\usepackage{multicol}
\usepackage{titlesec}

\definecolor{deltaBg}{RGB}{220,230,255} 

\usepackage{listings}

\title{\bf Detect Anything via Next Point Prediction}

\author{Qing Jiang$^{\ddagger}$, Junan Huo, Xingyu Chen, Yuda Xiong, Zhaoyang Zeng, \\ \vspace{-0.5em}
\textbf{Yihao Chen}, \textbf{Tianhe Ren},  \textbf {Junzhi Yu}, \textbf{Lei Zhang}${^\dagger}$ \\
International Digital Economy Academy (IDEA)
}

\correspondingauthor{
    $^\dagger$Corresponding author: leizhang@idea.edu.cn; $^\ddagger$Project Lead: jiangqing@idea.edu.cn. This work was done when Qing and Junan were interns at IDEA, and Xingyu was on an academic visit at IDEA.
}

\begin{document}
\begin{abstract}
    Object detection has long been dominated by traditional coordinate regression-based models, such as YOLO, DETR, and Grounding DINO. Although recent efforts have attempted to leverage MLLMs to tackle this task, they face challenges like low recall rate, duplicate predictions, coordinate misalignment, etc. In this work, we bridge this gap and propose \textbf{Rex-Omni}, a 3B-scale MLLM that achieves state-of-the-art object perception performance. On benchmarks like COCO and LVIS, Rex-Omni attains performance comparable to or exceeding regression-based models (e.g., DINO, Grounding DINO) in a zero-shot setting. This is enabled by three key designs: \textbf{1) Task Formulation}: we use special tokens to represent quantized coordinates from 0 to 999, reducing the model's learning difficulty and improving token efficiency for coordinate prediction; \textbf{2) Data Engines}: we construct multiple data engines to generate high-quality grounding, referring, and pointing data, providing semantically rich supervision for training; \textbf{3) Training Pipelines}: we employ a two-stage training process, combining supervised fine-tuning on 22 million data with GRPO-based reinforcement post-training. This RL post-training leverages geometry-aware rewards to effectively bridge the discrete-to-continuous coordinate prediction gap, improve box accuracy, and mitigate undesirable behaviors like duplicate predictions that stem from the teacher-guided nature of the initial SFT stage. Beyond conventional detection, Rex-Omni's inherent language understanding enables versatile capabilities such as object referring, pointing, visual prompting, GUI grounding, spatial referring, OCR and key-pointing, all systematically evaluated on dedicated benchmarks. We believe that Rex-Omni paves the way for more versatile and language-aware visual perception systems.
\end{abstract}

\maketitle
\section{Introduction}
\label{sec:intro}

Object detection~\cite{girshick2015fast, ren2016faster, redmon2016you, redmon2018yolov3, carion2020end, zhang2022dino, tian2019fcos, liu2016ssd, liu2022dab, zhao2024detrs, khanam2024yolov11, tan2020efficientdet, duan2019centernet} has long been a foundational task in computer vision due to its broad applications. The field has progressed from early CNN-based architectures, such as YOLO~\cite{redmon2016you} and Faster R-CNN~\cite{ren2016faster}, to Transformer-based models like DETR~\cite{carion2020end} and DINO~\cite{zhang2022dino}, while the task itself has evolved from traditional closed-set detection to open-set detection~\cite{liu2023grounding, li2022grounded, jiang2024t, cheng2024yolo, minderer2023scaling, ren2024dino, VLP:MDETR, minderer2022simple, minderer2023scaling} to better handle emerging real-world challenges.

\begin{figure}[t]     
    \includegraphics[width=1.0\textwidth]{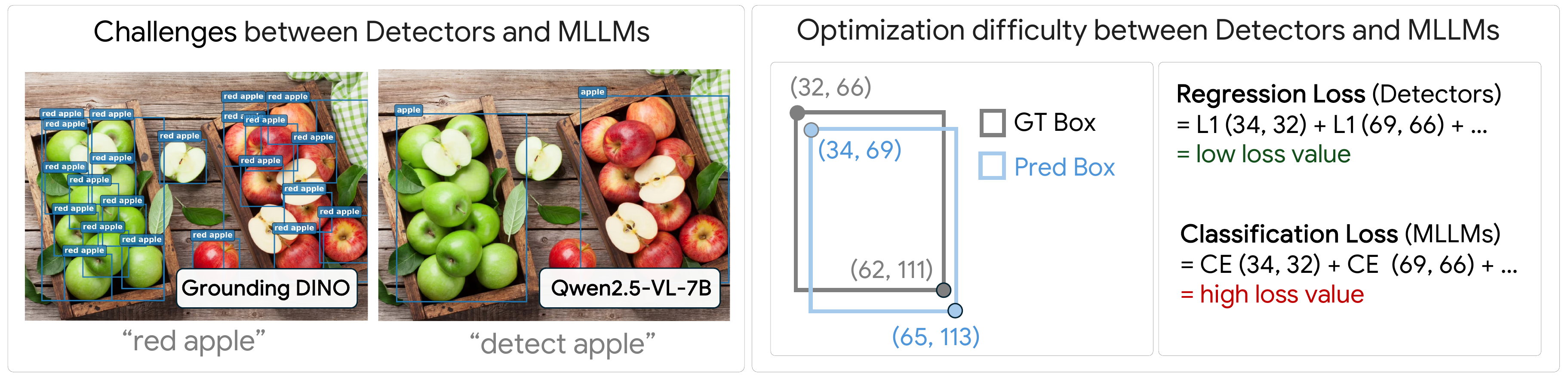}
    \caption{1) Detectors excel in localization but lack language understanding. MLLMs understand language well but struggle with localization. 2) Differences in optimization difficulty between detectors and MLLMs.}
    \label{fig:intro}
\end{figure}

A paramount objective in object detection is to develop models capable of identifying arbitrary objects and concepts. A common approach to this problem is open vocabulary object detection, where models such as Grounding DINO~\cite{liu2023grounding} and DINO-X~\cite{ren2024dino} leverage text encoders (e.g., BERT~\cite{devlin2018bert} or CLIP~\cite{radford2021clip}) to represent object categories and perform category-level open-set detection. Despite their effectiveness, these methods are fundamentally constrained by their relatively shallow language understanding, which restricts their ability to handle complex semantic descriptions (In Figure~\ref{fig:intro}, Grounding DINO detected all apples, despite the input prompt being \textit{red apple}.). Consequently, these methods are inherently limited in fully addressing this objective.

In contrast, multimodal large language models (MLLMs)~\cite{gpt4v,  VLM:Gemini, lu2025ovis2, wang2025internvl3, chen2022pali, agrawal2024pixtral, li2024aria, deitke2024molmo, wang2024qwen2} benefit from the strong language understanding capabilities of their underlying LLMs, presenting a promising avenue for integrating advanced language comprehension into object detection. A common MLLM-based approach~\cite{chen2023shikra, you2023ferret, wang2023cogvlm, zhan2025griffon, zhang2024ferret, jiang2024chatrex, ma2024groma, jiang2025referringperson, zhu2025internvl3, guo2025seed1, bai2025qwen2} is to represent coordinates as discrete tokens~\cite{chen2021pix2seq} and predict bounding boxes through next-token prediction. While conceptually elegant, existing MLLM-based methods have rarely matched the performance of traditional regression-based detectors on benchmarks such as COCO. As exemplified in Figure \ref{fig:intro}, even advanced MLLMs like Qwen2.5-VL~\cite{bai2025qwen2} struggles with precise object localization, in addition to facing limitations such as low recall rates, coordinate drift, and duplicate predictions.

We argue that the performance disparity in MLLM-based object detection primarily arises from two fundamental challenges inherent in their current formulation and training. First, MLLMs typically treat coordinate prediction as a discrete classification task, directly generating absolute coordinate values and relying on cross-entropy loss for supervision. While traditional regression-based models benefit from continuous, geometry-aware losses (e.g., L1, GIoU) that are directly sensitive to small geometric offsets, MLLMs face a significant learning difficulty in accurately mapping a fixed set of discrete tokens to the continuous pixel space. As illustrated in Figure \ref{fig:intro}, even minor pixel misalignments in discrete coordinate predictions can result in disproportionately large cross-entropy losses, hindering precise localization. This inherent challenge underscores the need for effective strategies to reduce coordinate learning complexity and provide extensive data for this mapping. 

Secondly, MLLMs commonly employ Supervised Fine-tuning (SFT) for teacher-guided next-token prediction training~\cite{radford2018improving}. While efficient, this paradigm creates a fundamental mismatch between training and inference. During SFT, the model is always conditioned on a ground-truth prefix, namely teacher forcing, meaning it is never exposed to its own, potentially imperfect, predictions. This training setup fails to capture the model's true performance in an autonomous generation setting.
This inherently prevents the model from developing robust behavioral awareness. Consequently, during free-form inference without this direct guidance, the model often struggles to regulate its own output structure. This leads to anomalous coordinate sequence generation, such as spurious duplicate predictions or object omissions, which undermine its overall performance. Addressing these two intertwined challenges is crucial for advancing MLLM-based object detection.

To overcome these inherent limitations and unlock the full potential of MLLMs for precise and versatile object perception, we propose \textbf{Rex-Omni}, a 3B-scale MLLM that achieves competitive performance with traditional detectors while distinctly excelling in language understanding capabilities. We address the aforementioned challenges through three core design principles:
\vspace{-1em}

\begin{itemize}
    \item \textbf{Task Formulation}: We unify visual perception tasks under a coordinate prediction framework, wherein each task is formulated as generating a sequence of coordinates. Specifically, pointing predicts one point, detection employs two points to form a bounding box, polygons use four or more points to represent object contours, and keypoint tasks output multiple semantic points. We adopt a quantized coordinate representation, where each coordinate value is mapped to one of 1,000 discrete tokens corresponding to values from 0 to 999. This approach significantly reduces the coordinate learning complexity and eases optimization, concurrently enhancing the efficiency of spatial representation.

    \item \textbf{Data Engines}: To facilitate the model's learning of the mapping between 1,000 discrete coordinate tokens and pixel-level positions, and to foster robust comprehension of complex natural language expressions, we design multiple specialized data engines for grounding, referring, and pointing tasks. These engines generate high-quality, semantically rich visual supervision signals for coordinate prediction.

    \item \textbf{Training Pipelines}:  We adopt a two-stage training paradigm. In the first stage, we perform supervised fine-tuning on 22 million data to teach the model basic coordinate prediction skills. In the second stage, we apply GRPO-based~\cite{deepseekmath} reinforcement post-training with three geometry-aware reward functions. This reinforcement phase serves two purposes: it enhances the precision of coordinate predictions through continuous geometric supervision, and crucially, it mitigates undesired behaviors (such as duplicate predictions) that arise from the teacher-guided nature of the initial SFT stage.
\end{itemize}

After this two-stage training, Rex-Omni achieves superior performance across a diverse range of perception tasks, as shown in Figure \ref{fig:teaser}., including object detection, object referring, visual prompting, GUI grounding, layout grounding, OCR, pointing, keypointing, and spatial referring. All of these tasks are achieved through direct prediction of coordinate points. To quantitatively assess its performance, Rex-Omni is first evaluated on COCO~\cite{lin2014microsoft}, a core benchmark for object detection. In a zero-shot setting (without training on COCO data), Rex-Omni demonstrates superior F1-score performance compared to traditional coordinate regression-based models (e.g., DINO-ResNet50, Grounding DINO) and other MLLMs (e.g., SEED1.5-VL~\cite{guo2025seed1}). Beyond COCO, Rex-Omni's performance is further benchmarked across diverse tasks, such as long-tailed detection, referring expression comprehension, dense object detection, GUI grounding, and OCR. Rex-Omni consistently outperforms both traditional detectors and MLLMs, thereby establishing a unified framework that combines precise localization with robust language understanding.

In summary, Rex-Omni represents a significant step towards unifying robust language understanding with precise visual perception. By carefully integrating principled task formulations, advanced data engines, and a sophisticated two-stage training pipeline, we demonstrate that MLLMs have the profound potential to define the next generation of object detection models, offering unprecedented versatility and a truly language-aware approach to visual perception systems.
\section{Task Formulation}
\label{sec:method}
In this section, we present Rex-Omni's task formulation design, covering its coordinate representation, the specific output formats for different tasks, and the details of its model architecture.
\subsection{Coordinate Formulation}

\begin{figure*}[t]
    \centering
    \includegraphics[width=\textwidth]{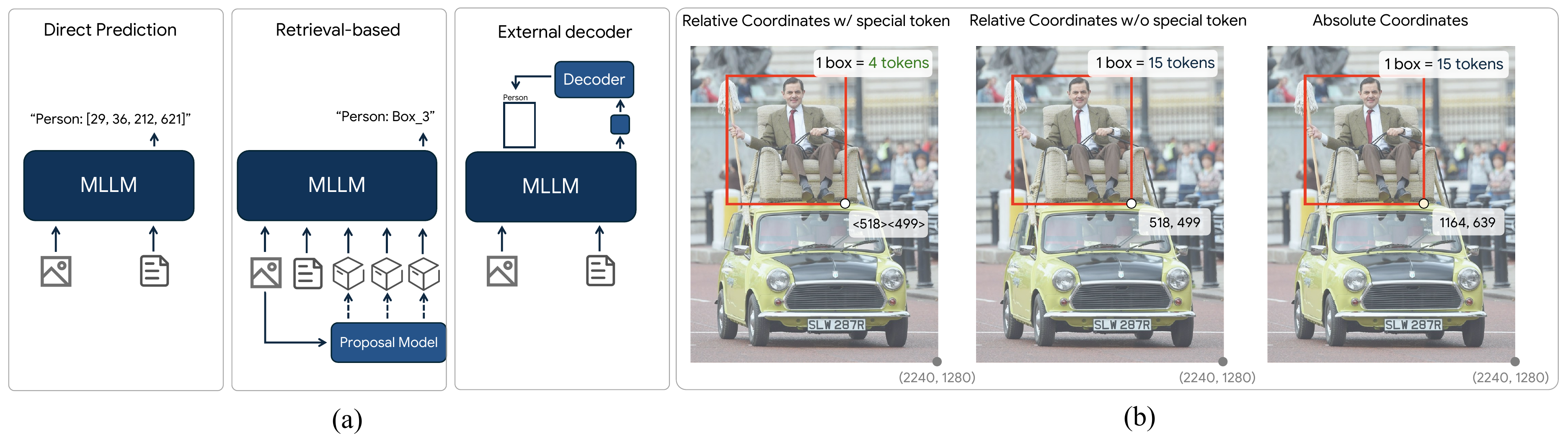}
    \caption{Design philosophy of Rex-Omni for coordinate prediction. It illustrates our chosen approach: (a) adopting a direct coordinate prediction strategy, and (b) employing a quantized relative coordinate format represented by special tokens for efficient and robust spatial encoding.}
    \label{fig:philosophy}
\end{figure*}

We begin by defining the output formulation for coordinate prediction. Existing approaches to utilizing MLLMs for this task can be broadly categorized into three paradigms, as illustrated in Figure \ref{fig:philosophy}a: \textbf{1) Direct Coordinate Prediction:} Inspired by the Pix2Seq~\cite{chen2021pix2seq} paradigm, these method~\cite{chen2023shikra, you2023ferret, wang2023cogvlm, zhan2025griffon, zhang2024ferret} treats coordinate values as discrete tokens within the language model's vocabulary, enabling the model to directly generate coordinate outputs;
\textbf{2) Retrieval-based Methods:} This approach~\cite{jiang2024chatrex, ma2024groma, jiang2025rex, jiang2025referringperson} incorporates an additional proposal module. The LLM is trained to predict the index of a candidate region or bounding box, thereby representing the output as a retrieval task over predefined proposals;
\textbf{3) External decoder:} In this strategy~\cite{zhang2023nextchatlmmchatdetection, wu2024visionllm, lai2024lisa, liu2025seg}, the LLM predicts special tokens, whose corresponding embeddings are then passed to an external decoder responsible for producing the final coordinates. We adopt the direct coordinate prediction strategy for Rex-Omni, motivated by its simplicity, flexibility, and the advantage of not relying on external modules or additional supervision.

Within the direct coordinate prediction paradigm, several variations exist, as illustrated in Figure \ref{fig:philosophy}b:
\textbf{1) Relative coordinates with special tokens:} Coordinates are quantized to values between 0 and 999, with each coordinate represented by a special token in the LLM's vocabulary. The model is thus trained to predict these 1,000 tokens as representations for coordinates. A representative model is Pix2Seq~\cite{chen2021pix2seq}.
\textbf{2) Relative coordinates without special tokens:} Coordinates are similarly quantized to 1,000 bins; however, they are represented by multiple atomic tokens rather than a single special token. A representative model is SEED1.5-VL~\cite{guo2025seed1}.
\textbf{3) Absolute coordinates:} This method uses absolute coordinates, where a coordinate value such as 1921 is tokenized into individual digits (1, 9, 2, 1).  A representative model is Qwen2.5-VL~\cite{bai2025qwen2}. We choose relative coordinates with special tokens modeling approach for two primary reasons: First, selecting relative coordinates over absolute coordinates inherently reduces learning complexity by confining the classification task to a bounded range of 1,000 categories. Second, the use of dedicated special tokens for coordinates significantly reduces the required token length per coordinate. For instance, a bounding box is represented by only four special tokens, in contrast to 15 atomic tokens (including separators) without such a scheme. This significantly improves token efficiency and inference speed, especially in dense object scenes.

\subsection{Input Format}

Rex-Omni adopts a unified text-based interface for all visual perception tasks. Each task is expressed as a natural language query that specifies the target objects or relationships to be identified in the image. This design allows the model to seamlessly integrate diverse vision-language tasks under a single instruction-driven framework.

\vspace{0.3em}
\noindent\textbf{Text Prompts.}  
For most tasks, the model receives an image paired with a text prompt formulated in natural language. The text prompt can describe one or multiple. When multiple targets are specified, their corresponding categories or referring expressions are concatenated using commas. For example:

\begin{tcolorbox}[
  colback=lightblue!5,
  colframe=blue!20!black,
  coltitle=black,
  colbacktitle=lightblue!20,
  title=Example of a text prompt for multi-object detection,
  fonttitle=\bfseries,
  boxrule=1.0pt
]
Please detect pigeon, person, truck, snow in this image. Return the output in box format.
\end{tcolorbox}

\noindent For different tasks, we design distinct query styles to guide the model for generation.

\vspace{0.3em}
\noindent\textbf{Visual Prompts.}  
While text prompts offer strong generalization and interpretability, they face limitations when dealing with objects that lack clear linguistic descriptions—particularly rare or visually complex categories. As shown in prior work such as T-Rex2~\cite{jiang2025t}, certain objects are inherently difficult to express through text alone. To address this, Rex-Omni supports visual prompting, allowing users to provide bounding boxes as an additional and intuitive form of input.

Unlike existing methods~\cite{jiang2025t, ren2024dino, jiang2023t} that treat visual prompts as feature-matching problems by extracting embeddings from the indicated region and comparing them to detection queries, Rex-Omni adopts a unified text-based interface. Given a visual prompt in box format, the corresponding region is first converted into quantized coordinate tokens. The model is then guided through natural language instructions to identify all objects that share the same category as the indicated region. This design seamlessly integrates visual prompting into the generative text framework, enabling the model to reason about visual correspondence through language.

\begin{tcolorbox}[
colback=lightblue!5,
colframe=blue!20!black,
coltitle=black,
colbacktitle=lightblue!20,
title=An example of visual prompting in Rex-Omni,
fonttitle=\bfseries,
boxrule=1.0pt
]
Here are some example boxes specifying the location of several objects in the image:
{"object1": ["<12><412><339><568>", "<92><55><179><378>"]}.
Please detect all objects with the same category and return their bounding boxes in [x0, y0, x1, y1] format.
\end{tcolorbox}

\subsection{Output Format for Each Task}  
The output for each visual task is uniformly represented as a structured token sequence that includes descriptive phrases, coordinate tokens, and special tokens for demarcation, organized as follows::
\begin{tcolorbox}[
  colback=lightblue!5,       
  colframe=blue!20!black,     
  coltitle=black,             
  colbacktitle=lightblue!20, 
  title=Basic output format of Rex-Omni,
  fonttitle=\bfseries,
  boxrule=1.0pt
]
<|object\_ref\_start|>\textbf{PHRASE}<|object\_ref\_end|><|box\_start|> \textbf{COORDS}<|box\_end|>
\end{tcolorbox}

\noindent Here, \textbf{PHRASE} denotes the category or description of the object represented by the coordinate sequence, and \textbf{COORDS} refers to the sequence of coordinates.  Rex-Omni is bulid upon Qwen2.5-VL-3B and we retain Qwen2.5-VL's original special tokens for task formatting, including the phrase start token (<object\_ref\_start>), phrase end token (<object\_ref\_end>), coordinate start token (<box\_start>), and coordinate end token (<box\_end>).

\noindent For tasks involving outputting boxes, such as object detection, COORDS consists of a sequence of coordinates in the format of [x0, y0, x1, y1], sorted by x0 in ascending order. For example:

\begin{tcolorbox}[
  colback=lightblue!5,       
  colframe=blue!20!black,     
  coltitle=black,             
  colbacktitle=lightblue!20, 
  title=An example of a task for outputting bounding boxes,
  fonttitle=\bfseries,
  boxrule=1.0pt
]
<|object\_ref\_start|>person<|object\_ref\_end|><|box\_start|><12><42><512><612>, <24><66><172><623>, ...<|box\_end|>, ... (more phrases)
\end{tcolorbox}

\noindent For tasks involving outputting points, such as object pointing, COORDS is composed of a sequence of [x0, y0] pairs. For example:

\begin{tcolorbox}[
  colback=lightblue!5,       
  colframe=blue!20!black,     
  coltitle=black,             
  colbacktitle=lightblue!20, 
  title=An example of a task for outputting points,
  fonttitle=\bfseries,
  boxrule=1.0pt
]
<|object\_ref\_start|>button<|object\_ref\_end|><|box\_start|><100><150>,<200><250>, ...<|box\_end|>, ... (more phrases)
\end{tcolorbox}

\noindent For tasks involving outputting polygons, such as OCR, COORDS consists of a sequence of coordinates in the format [x0, y0, x1, y1, x2, y2, ...]. For example:

\begin{tcolorbox}[
  colback=lightblue!5,       
  colframe=blue!20!black,     
  coltitle=black,             
  colbacktitle=lightblue!20, 
  title=An example of a task for outputting polygons,
  fonttitle=\bfseries,
  boxrule=1.0pt
]
<|object\_ref\_start|>text<|object\_ref\_end|><|box\_start|><10><20>...<|box\_end|>, ... (more phrases)
\end{tcolorbox}

\noindent For the keypointing task, we output a structured JSON format that includes both the bounding box of the object and its associated keypoints.

\begin{tcolorbox}[
  colback=lightblue!5,       
  colframe=blue!20!black,     
  coltitle=black,             
  colbacktitle=lightblue!20, 
  title=An example of keypoint detection task,
  fonttitle=\bfseries,
  boxrule=1.0pt
]
\{"person1": \{"box": <0><123><42><256>, "keypoints": \{"left eye": <32><43>, "right eye":  <66><55>, ...\}\}, \{"person2": \{"box": <51><116><72><522>, "keypoints": \{"left eye": <342><23>, "right eye":  <16><571>, ...\}\}\}
\end{tcolorbox}

\noindent For the simultaneous detection of multiple phrases, the predicted outputs corresponding to different phrases are concatenated using commas. If a particular phrase refers to an object that is not present in the image, the corresponding COORDS field is replaced with \textbf{None}.

\subsection{Model Architecture}

\begin{figure*}[t]
    \centering
    \includegraphics[width=0.8\textwidth]{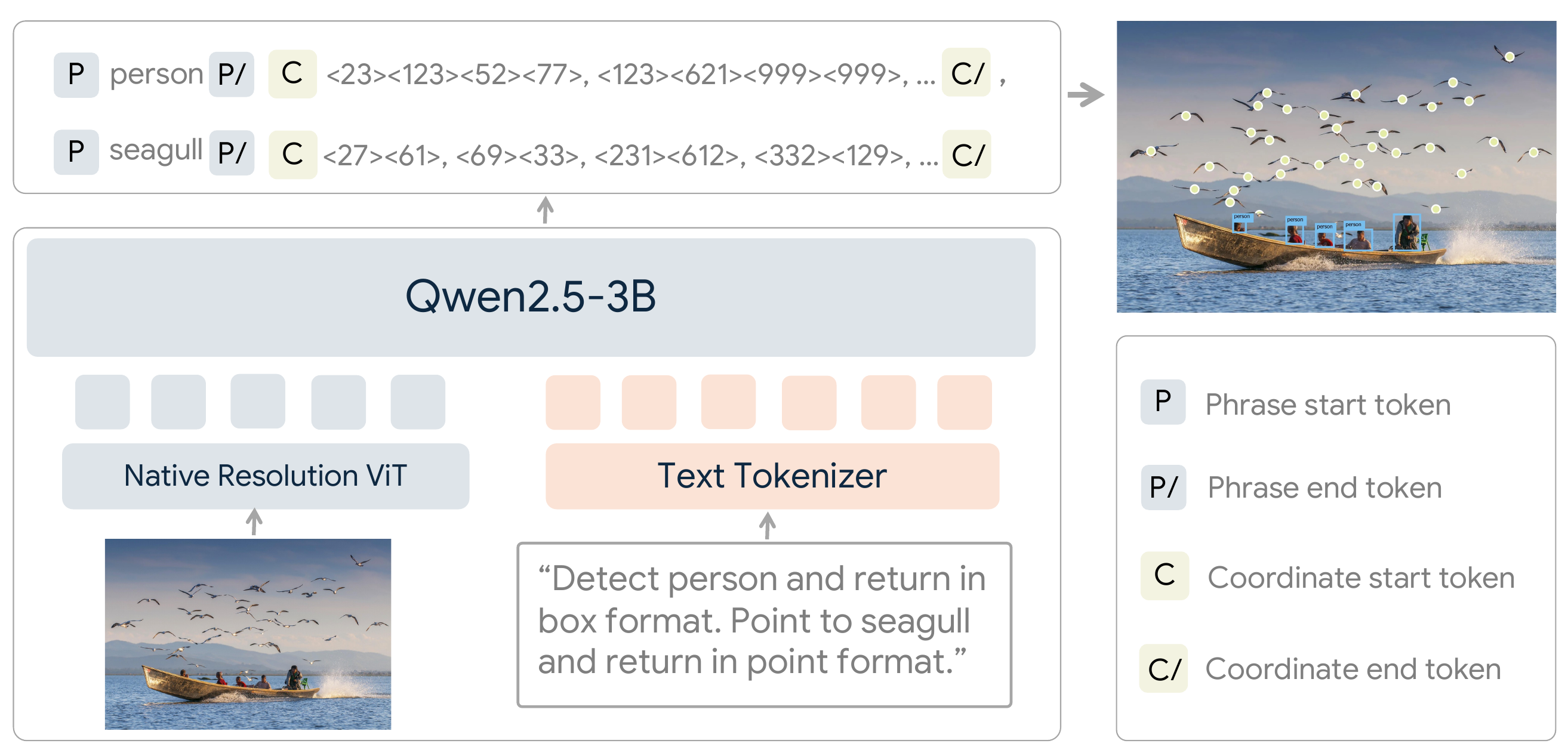}
    \caption{Overview of the Rex-Omni Model Architecture. Rex-Omni is constructed upon the Qwen2.5-VL-3B backbone with minimal architectural modifications. Notably, the last 1,000 tokens of the original vocabulary are repurposed to serve as dedicated special tokens, representing quantized coordinate values from 0 to 999.}
    \label{fig:model}
\end{figure*}

As shown in Figure~\ref{fig:model}, Rex-Omni is built upon the Qwen2.5-VL-3B-Instruct model with minimal architectural modifications. While the original Qwen2.5-VL employs an absolute coordinate encoding scheme, we adapt the model to support relative coordinate representations without introducing additional parameters. Specifically, we repurpose the final 1,000 tokens of the model's vocabulary to serve as special tokens, each corresponding to a quantized coordinate ranging from 0 to 999.

\section{Training Data}
To equip Rex-Omni with both precise coordinate prediction capabilities and strong language understanding, we utilize two sources of training data: publicly available datasets and automatically annotated data generated by our custom-designed data engines.
\begin{table*}[t]
\centering
  \resizebox{1.0\linewidth}{!}{
    \begin{tabular}{c|c|c|c}
\hline
Task             & Output Format & Question Template Example                                                                                                                                                                                           & Datasets                                                                                                                                                                                               \\ \hline
Object Detection & Box           & Detect {[}PHRASE{]} in this image                                                                                                                                                                                   & \begin{tabular}[c]{@{}c@{}}APTv2~\cite{yang2023aptv2}, BDD100K~\cite{yu2020bdd100k}, DeepFashion~\cite{liu2016deepfashion}\\  DOTAv2~\cite{xia2018dota}, EgoObjects~\cite{zhu2023egoobjects}, FAIR-1M~\cite{sun2022fair1m}\\ HumanParts~\cite{li2018detector},  ImageNet-Part~\cite{he2022partimagenet} NuImages~\cite{caesar2020nuscenes}\\ PACO~\cite{ramanathan2023paco},  OpenImages~\cite{Datasets:OpenImages}, O365~\cite{shao2019objects365}\\ V3Det~\cite{wang2023v3det}, VisDrone~\cite{du2019visdrone}\end{tabular} \\ \hline
Object Referring & Box           & Detect {[}PHRASE{]} in this image                                                                                                                                                                                   & HumanRef, RefCOCOg                                                                                                                                                                                     \\ \hline
Visual Prompting & Box           & \begin{tabular}[c]{@{}c@{}}Given reference boxes {[}BOX{]} indicating one or more \\ objects,  find all objects with the same category\end{tabular}                                                                 & \begin{tabular}[c]{@{}c@{}}O365~\cite{shao2019objects365}, OpenImages~\cite{Datasets:OpenImages}, HierText~\cite{long2022towards} \\ CrowdHuman~\cite{shao2018crowdhuman}, SA-1B~\cite{TransF:SAM}, VisDrone~\cite{du2019visdrone} \\ FSCD147~\cite{nguyen2022few}\end{tabular}                                                                                         \\ \hline
OCR              & Box / Polygon & Detect all the text in box/polygon format and recognize them                                                                                                                                                        & \begin{tabular}[c]{@{}c@{}}Art~\cite{chng2019icdar2019}, HierText~\cite{long2022towards}, ICDAR2013~\cite{lucas2005icdar} \\ ICDAR2015~\cite{karatzas2015icdar},  LSVT~\cite{sun2019chinese}, RCTW~\cite{shi2017icdar2017}\\ ReCTS~\cite{zhang2019icdar}\\ SROIE~\cite{huang2019icdar}, TextOCR~\cite{singh2021textocr}\\ IDLOCR~\cite{biten2022ocr}, WildReceipt~\cite{sun2021spatial}\end{tabular}                                                      \\ \hline
Layout Groudning & Box           & Detect {[}PHRASE{]} in this image                                                                                                                                                                                   & \begin{tabular}[c]{@{}c@{}}DocLayNet~\cite{pfitzmann2022doclaynet}, PubLayNet~\cite{zhong2019publaynet}, TableBank~\cite{li2020tablebank} \\ M6Doc~\cite{cheng2023m6doc}, CDLA~\cite{li2021cdla}, TabRecSet~\cite{yang2023large}\end{tabular}                                                                                                     \\ \hline
GUI Grounding    & Box / Point   & Detect/Point to element {[}PHRASE{]}                                                                                                                                                                                & Os-Atlas~\cite{wu2024atlas}, UI-RefExp~\cite{bai2021uibert}, ShowUI~\cite{lin2024showui}                                                                                                                                                                            \\ \hline
Pointing         & Point         & Point to {[}PHRASE{]}                                                                                                                                                                                               & Pixmo-point~\cite{deitke2024molmo}                                                                                                                                                                                            \\ \hline
Affordance       & Point         & Point to {[}PHRASE{]}                                                                                                                                                                                               & AGD20K~\cite{luo2022learning}                                                                                                                                                                                                 \\ \hline
Spatial Referring       & Point         & Point to {[}PHRASE{]}                                                                                                                                                                                               & RefSpatial~\cite{zhou2025roborefer}                                                                                                                                                                                                 \\ \hline
KeyPointing      & Box \& Point  & \begin{tabular}[c]{@{}c@{}}Can you detect each {[}PHRASE{]} in the image \\ in box format, and then provide the coordinates of \\ its {[}KEYPOINT{]} as {[}x0, y0{]}? Output the \\ answer in JSON format.\end{tabular} & \begin{tabular}[c]{@{}c@{}}AP10K~\cite{yu2021ap}, APT36K~\cite{yang2022apt}, COCO-Keypoint~\cite{Datasets:MSCOCO} \\ MacaquePose~\cite{labuguen2021macaquepose}, HumanArt~\cite{ju2023human}, MPII~\cite{andriluka20142d}\\ OCHuman~\cite{zhang2019pose2seg}, CrowdPose~\cite{li2019crowdpose}\end{tabular}                                                                               \\ \hline
\end{tabular}}
  \caption{Publicly available training datasets used by Rex-Omni, covering tasks such as object detection, referring, prompting, OCR, grounding, pointing, affordance, and keypointing, with outputs including boxes, points, polygons, and JSON-formatted keypoints.}
  \label{tab:Public_dataset}
  \vspace{-2mm}
  \end{table*} 

\subsection{Public Datasets}
In Table \ref{tab:Public_dataset}, we enumerate the publicly available datasets leveraged for Rex-Omni's training across various subtasks, including Object Detection, Object Referring, Visual Prompting, OCR, Layout Grounding, GUI Grounding, Pointing, Affordance Grounding, Spatial Referring, and Keypointing. For each of these tasks, a set of question templates was defined to construct corresponding question-answer (QA) pairs. In total, approximately 8.9 million public data samples were utilized.

\subsection{Data Engines}

Effective training of Rex-Omni necessitates learning a fine-grained mapping between its 1,000 quantized coordinate tokens and the continuous pixel space of images. This capability demands a substantially larger volume of high-quality training data than what is conventionally available in existing public datasets. Moreover, while many public datasets offer category-level annotations, those providing richer, instance-level semantic grounding (e.g., referring expressions) remain scarce in both scale and diversity. To address these limitations, we developed a dedicated suite of data engines engineered to generate high-quality, large-scale training data specifically tailored for fine-grained spatial reasoning and complex language grounding tasks.

\begin{figure*}[t]
    \centering
    \includegraphics[width=\textwidth]{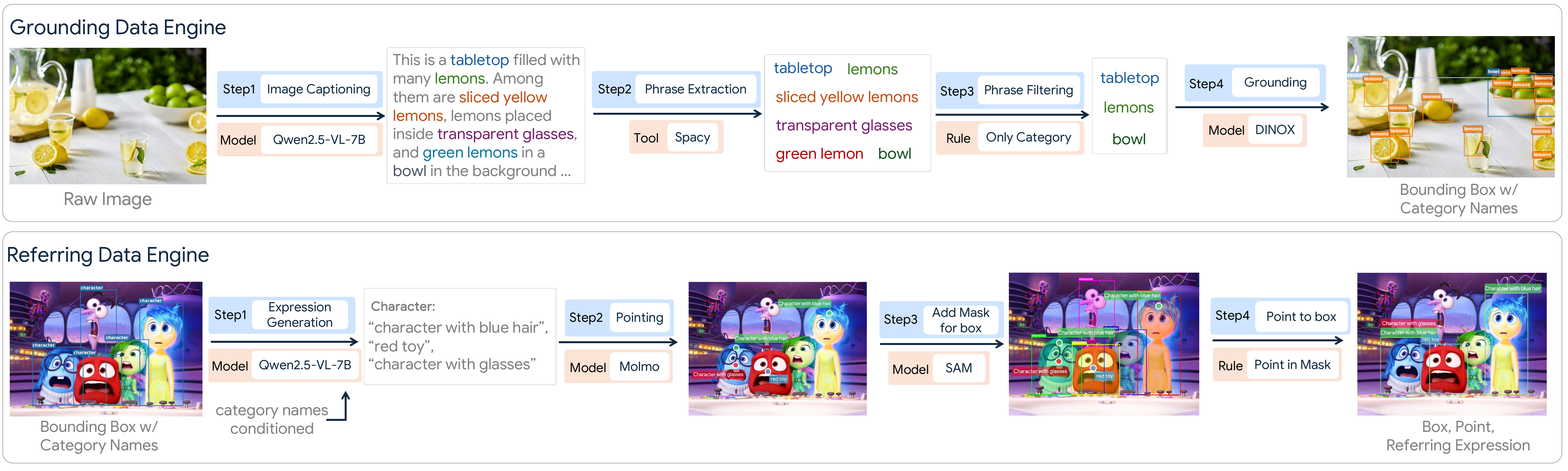}
    \caption{Pipelines of our two primary data engines. The figure illustrates the processes of the Grounding Data Engine (top) and the Referring Data Engine (bottom), which are custom-designed to produce extensive, high-quality grounding and referring data for Rex-Omni's training.
}
    \label{fig:data_engine}
\end{figure*}

\subsubsection{Grounding Data Engine}

A common strategy for constructing large-scale detection datasets is to develop a grounding data engine~\cite{jiang2024t, ren2024dino, ren2024grounding, cheng2024yolo, peng2023kosmos2}, which typically involves generating image captions, extracting candidate phrases, and using a grounding model (e.g., Grounding DINO) to assign bounding boxes to those phrases. In contrast to prior approaches, we introduce a phrase filtering stage into the pipeline to improve annotation quality. Specifically, our annotation process consists of the following four stages:

\begin{itemize}
    \item \textbf{Image Captioning:} We begin by generating descriptive captions for each image using Qwen2.5-VL-7B-Instruct. These captions provide natural language descriptions of the visual content, typically covering multiple objects within the scene.
    \item \textbf{Phrase Extraction:} We then apply the SpaCy\footnote{\url{https://spacy.io/}} NLP toolkit to extract noun phrases from the generated captions. These phrases may include basic class names (e.g., tabletop, lemon) as well as more specific descriptions (e.g., sliced yellow lemons, green lemons).
    \item \textbf{Phrase Filtering:} This step marks a key departure from prior approaches. To minimize data ambiguity, we remove noun phrases containing descriptive attributes such as adjectives (e.g., green lemon is discarded, while lemon is retained). The rationale is that current grounding models struggle to accurately interpret such descriptive expressions, often detecting all instances of a category regardless of the modifier. For instance, the phrase green lemon may incorrectly trigger detections of all lemons, thereby introducing significant labeling errors.
    \item \textbf{Phrase Grounding:} Finally, we use DINO-X \cite{ren2024dino}, an open-vocabulary object detector, to produce bounding boxes corresponding to the filtered phrases.
\end{itemize}

For this data engine, images are primarily sourced from the COYO~\cite{kakaobrain2022coyo-700m} and SA-1B~\cite{TransF:SAM} datasets. We apply rigorous preprocessing, including discarding low-resolution images and filtering content labeled as NSFW. This process yields a curated dataset of approximately 3 million images, each annotated with high-quality grounding labels.

\subsubsection{Referring Data Engine}

Unlike detection or grounding data, which primarily emphasize object category names, referring data necessitate semantically richer natural language descriptions, exemplified by phrases like “a man in a yellow shirt”. The RexSeek~\cite{jiang2025referringperson} study underscores that high-quality referring annotations should accommodate a single referring expression mapping to multiple instances, thereby fostering the model's ability to learn flexible and context-aware reference grounding. However, RexSeek's reliance on manual annotation renders it labor-intensive and inherently unscalable. To address this limitation, we design a fully automated referring data engine capable of generating large-scale referring data without human supervision.

\begin{itemize}
    \item \textbf{Expression Generation:} Given an image annotated with bounding boxes and corresponding category labels, we prompt Qwen2.5-VL-7B with the image and category information to generate a set of referring expressions. Each expression is designed to naturally describe an object category present in the image, mimicking human-like descriptions.
    \item \textbf{Pointing:} For each generated referring expression, we employ Molmo~\cite{deitke2024molmo}, a state-of-the-art referring model, to produce the corresponding spatial point. Although Molmo outputs only point-level predictions, it exhibits strong performance in understanding and grounding referring expressions.
    \item \textbf{Mask Generation:} We apply SAM~\cite{TransF:SAM} to generate a mask for each ground-truth bounding box in the image.
    \item \textbf{Point-to-Box Association:} Each point produced by Molmo is aligned with a SAM-generated mask. When a point lies within a mask, the corresponding bounding box is linked to the referring expression, thereby grounding the language in the object region.
\end{itemize}

For this data engine, we use images from O365~\cite{shao2019objects365}, OpenImages~\cite{Datasets:OpenImages}, and additional data generated by our Grounding Data Engine. Through this pipeline, we obtain approximately 3 million images with automatically generated referring annotations.

\subsubsection{Other Data Engines}
In addition to grounding and referring data, we develop two relatively lightweight data engines to generate datasets for pointing and OCR tasks.

\begin{itemize}
    \item \textbf{Pointing Data Engine:} Point-level supervision offers an efficient alternative to bounding boxes, particularly when object boundaries are ambiguous or difficult to delineate (e.g., edges, whitespace, or fine structures). To derive point annotations from box-level supervision, we adopt a geometry-aware strategy. Given a bounding box, SAM is first used to obtain the corresponding segmentation mask. We then compute the minimum-area enclosing rotated rectangle of the mask and take the intersection of its diagonals as the candidate point. If this point lies within the mask, it is designated as the point annotation for the box. Through this conversion, we obtain approximately 5 million point-level samples from existing detection datasets as well as from the outputs of our grounding and referring data engines.
    
    \item \textbf{OCR Data Engine:} PaddleOCR\footnote{\url{https://github.com/PaddlePaddle/PaddleOCR}} is utilized to annotate images containing textual content, extracting both polygonal boundaries of text regions and their corresponding transcriptions. For each extracted polygon, the minimum enclosing axis-aligned rectangle is subsequently computed to serve as its bounding box representation. Images are sourced from the COYO dataset, yielding approximately 2 million OCR-annotated samples.
\end{itemize}

In total, combining publicly available datasets and data generated by our annotation pipelines, we obtain 22 million high-quality annotated images for training.
\section{Training Pipelines}
\begin{figure*}[t]
    \centering
    \includegraphics[width=\textwidth]{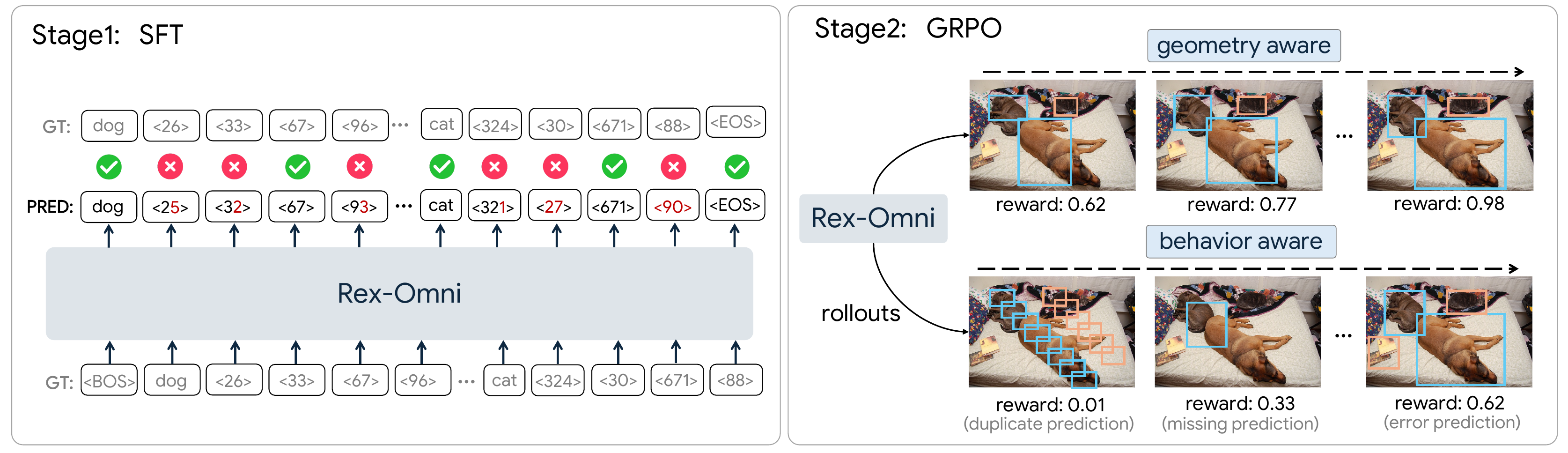}
    \caption{Overview of Rex-Omni's two-stage training pipeline. The first stage involves supervised fine-tuning (SFT) on 22 million samples to establish fundamental coordinate prediction capabilities. This is followed by GRPO-based reinforcement post-training, which leverages geometry-aware rewards and behavior-aware optimization to refine precision and correct SFT-induced behavioral deficiencies.}
    \label{fig:training_pipelines}
\end{figure*}

We employ a two-stage training strategy, depicted in Figure \ref{fig:training_pipelines}. In the first stage, supervised fine-tuning (SFT) is performed on 22 million annotated samples using a teacher-guided approach, enabling the model to acquire fundamental coordinate prediction capabilities. In the second stage, we apply reinforcement learning based on the GRPO framework, which further refines the model's performance by combining geometry-aware rewards with behavior-aware optimization, thus addressing the limitations of the SFT stage and enhancing overall prediction quality.

\subsection{Stage1: Supervised Fine-Tuning}
Since the model predicts coordinates in the form of quantized tokens ranging from 0 to 999, it must first learn how to accurately map these discrete values back to continuous pixel locations within the image. This corresponds to a 1,000 way classification problem over spatial positions, which requires substantial supervision to achieve reliable performance. Therefore, we begin training with a teacher-guided supervised fine-tuning stage on large-scale annotated data, enabling the model to acquire the fundamental ability to interpret and predict spatial coordinates.

We adopt the following online strategy to construct SFT conversation data:
\vspace{-1em}
\begin{itemize}
    \item \textbf{Conversation Templates:} For each training task, we construct multiple question templates with GPT-4o to mimic real user scenarios. These templates include placeholders for \textbf{PHRASE} keywords, which are replaced with actual phrases from the data during training.
    \item \textbf{Multi-Phrase Queries:} In practical settings, users may wish to detect multiple object categories within a single image. To reflect this, if an image contains $N$ annotated phrases, we randomly sample between 1 and $N$ phrases to form the training query.
    \item \textbf{Visual Prompting Training:} Following T-Rex2~\cite{jiang2025t}, for each training sample consisting of an image and its annotated category-specific bounding boxes, we simulate visual prompting scenarios. Specifically, for each category present in the image, we randomly sample between 1 and $N$ bounding boxes, where $N$ denotes the maximum number of annotated instances for that category. These sampled boxes are treated as visual prompts and converted into quantized coordinate tokens consistent with our coordinate formulation. The model is then instructed, through natural language queries, to detect all objects of the same category as indicated by the given visual prompts.

\end{itemize}

\noindent We adopt the standard cross-entropy loss for training. The model is trained on 8 nodes, each equipped with 8 A100 GPUs, and the total training time is approximately 8 days. All model parameters are updated during training. We use separate learning rates for different components: 2e-6 for the vision encoder, and 2e-5 for both the projection layer and the LLM. Optimization is performed using the AdamW~\cite{loshchilov2017adamw} optimizer with a learning rate warm-up of 3\% and a weight decay of 0.01. Following the architecture of Qwen2.5-VL, Rex-Omni also employs a native resolution Vision Transformer as its vision encoder. We constrain the number of input pixels to range from a minimum of 16 × 28 × 28 to a maximum of 2560 × 28 × 28. Given a ViT patch size of 28, this limits the number of image tokens between 16 and 2560.

\subsection{Stage2: Reinforcement Post-Training}

\subsubsection{Limitations of SFT}
\label{sec:sft_limitation}
While SFT allows the model to quickly acquire basic coordinate prediction capabilities by leveraging massive amounts of labeled data, it presents two key limitations: 

\textbf{Geometric Discretization Issue.} Using cross-entropy loss for coordinate prediction inherently introduces a discretization problem. Coordinates are represented as categorical tokens (from \texttt{<0>} to \texttt{<999>}), and the model is trained to classify each token exactly. However, this formulation is misaligned with the continuous nature of geometry in spatial tasks. For example, if the ground-truth token is \texttt{<33>} but the model predicts \texttt{<32>}, the difference in pixel space may be negligible, yet the CE loss penalizes it as a completely incorrect prediction. Conversely, if the ground truth is \texttt{<0><0><100><100>} but the model predicts \texttt{<0><0><100><1000>}, only one token is misclassified. In this case, the CE loss remains relatively small, even though the resulting bounding box is severely misaligned and the geometric error is substantial.

\textbf{Behavioral Regulation Deficiency.} In the SFT stage, teacher-forced training relies on full ground-truth sequences for efficient parallel learning. This setup fixes the number of predicted boxes to the ground-truth count, preventing the model from autonomously learning how many objects to predict. Consequently, during inference the model often fails to regulate output quantity, leading to two typical errors: (1) predicting fewer boxes than required (missed detections), or (2) predicting more boxes than necessary (repetitive detections with identical or slightly shifted coordinates). These behaviors reflect the model’s lack of effective output regulation.

\subsubsection{GRPO-based Post-Training}
To address the geometry and behavior-related limitations of SFT, we adopt a reinforcement post-training strategy based on GRPO~\cite{deepseekmath}. GRPO enables the model to explore its own output space and improve through reward-guided optimization. Given an image and a question $(I, x)$, the model samples a group of $G$ complete responses $\{o_1, o_2, \dots, o_G\}$ from the current policy $\pi_\theta$. Each response consists of a full reasoning trace and a final set of predicted coordinates or boxes, depending on the task. For each output $o_i$, we compute a scalar reward $r_i$, which is normalized across the group to obtain the \textit{relative advantage}:

\begin{equation}
A_i = \frac{r_i - \mathrm{mean}(r_1, \dots, r_G)}{\mathrm{std}(r_1, \dots, r_G)}.
\end{equation}

\noindent These group-based advantages provide fine-grained credit assignment among diverse outputs, encouraging the model to prefer more accurate and non-redundant predictions. The GRPO objective is formulated as a clipped policy gradient with KL regularization:

\begin{equation}
\mathcal{J}_{\text{GRPO}}(\theta) = \frac{1}{G} \sum_{i=1}^G \frac{1}{|o_i|} \sum_{t=1}^{|o_i|} \left[\min \left(\rho_{i,t} \hat{A}_{i,t}, \text{clip}(\rho_{i,t}, 1 - \epsilon, 1 + \epsilon)\hat{A}_{i,t} \right) - \beta \mathbb{D}_{\text{KL}}[\pi_\theta \| \pi_{\text{ref}} \right],
\end{equation}

\noindent where $\rho_{i,t}$ is the importance sampling ratio, and $\pi_{\text{ref}}$ is the model frozen after the SFT stage. The KL penalty ensures training stability by preventing excessive divergence from the reference model.

This framework naturally mitigates both geometry and behavior limitations: 1) rewards can be made \textit{geometry-aware}, such as IoU or L1 distance metrics, directly encouraging accurate spatial alignment beyond token-level correctness; and 2) by allowing \textit{variable-length outputs}, the model can learn to avoid repetition or over-generation. Repetitive or redundant predictions receive lower rewards, leading to more concise and behaviorally aligned responses.

\subsubsection{Geometry-aware Rewards}

To provide informative feedback on the spatial quality of predictions, we design three geometry-aware reward functions tailored to different tasks: box IoU reward, point-in-mask reward, and point-in-box reward. These reward types reflect the structural correctness of the predicted outputs with respect to ground-truth annotations.

\textbf{Box IoU Reward.} This reward is applied to tasks requiring bounding box predictions, including object detection, grounding, referring, and OCR. The reward encourages both accurate localization and correct object-category alignment.

\noindent Given a set of predicted boxes $\hat{B} = \{\hat{b}_1, \dots, \hat{b}_m\}$ and the ground-truth boxes $B^* = \{b^*_1, \dots, b^*_n\}$, we perform a ground-truth-guided matching. For each ground-truth box $b^*_j$, we find the predicted box $\hat{b}_i$ that maximizes the IoU with $b^*_j$:

\begin{equation}
\text{IoU}(b^*_j, \hat{b}_i) = \max_{\hat{b}_i \in \hat{B}} \text{IoU}(b^*_j, \hat{b}_i).
\end{equation}

\noindent If the category label of $\hat{b}_i$ matches that of $b^*_j$, we assign the IoU value as the reward $r_j$ for that ground-truth box. Otherwise, $r_j = 0$. Let $R = \{r_1, \dots, r_n\}$ be the reward set for all GT boxes. We then compute recall and precision as follows:

\begin{equation}
\text{Recall} = \frac{\sum_{j=1}^{n} r_j}{n}, \quad
\text{Precision} = \frac{\sum_{j=1}^{n} r_j}{m}, \quad
r^{\text{IoU}} = \frac{2 \cdot \text{Precision} \cdot \text{Recall}}{\text{Precision} + \text{Recall} + \epsilon},
\end{equation}

\noindent where $\epsilon$ is a small constant to prevent division by zero. This formulation rewards both spatial accuracy and label correctness. It penalizes unmatched or misclassified predictions and balances over- and under-prediction through the F1-style reward signal.

\textbf{Point-in-Mask Reward.} This reward is applied to tasks where the model localizes objects via point predictions, such as pointing-based detection, grounding, and referring. It evaluates whether a predicted point lies within the object mask.

\noindent Given a set of ground-truth bounding boxes $B^* = \{b^*_1, \dots, b^*_n\}$, we apply SAM to extract a binary mask $M_j$ for each ground-truth box $b^*_j$. Let $\hat{P} = \{\hat{p}_1, \dots, \hat{p}_m\}$ denote the predicted points, each associated with a category label. For each ground-truth mask $M_j$, we determine whether there exists a predicted point $\hat{p}_i$ that lies inside $M_j$:
\begin{equation}
\exists \hat{p}_i \in \hat{P},\quad \text{s.t.} \quad \hat{p}_i \in M_j.
\end{equation}

\noindent If such a point exists and its associated category label matches that of $M_j$, we assign a reward of 1 to the corresponding ground-truth instance; otherwise, the reward is 0. Precision, recall, and F1 reward are then computed using the same formulation as in the Box IoU Reward.

\textbf{Point-in-Box Reward.} This reward is specifically designed for the GUI Grounding task, where the model is expected to predict a point indicating the clickable position (e.g., a button) on a graphical user interface. If the predicted point falls within the ground-truth bounding box of the target GUI element, a reward of 1 is assigned; otherwise, the reward is 0. This simple binary reward effectively encourages precise point-level interaction behavior required in GUI scenarios.

\subsubsection{Implementation Details}

We sample 66K data from the SFT dataset to serve as training data for the GRPO stage. We reuse the same dialogue templates from the SFT phase. The GRPO training is conducted on 8 A100 GPUs for approximately 24 hours. We set the rollout size to 8, the KL penalty coefficient $\beta$ to 0.01, and use a batch size of 64. All model parameters are updated during this stage.

\section{Benchmark Results}
\label{sec:experiment}
This section presents the evaluation of Rex-Omni across multiple visual perception tasks, such as common, long-tailed, and dense object detection, referring object detection, and object pointing. For each task, we outline the benchmark datasets, experimental settings, and evaluation metrics.

\subsection{Common Object Detection}
\label{sec:coco}
Common object detection refers to the task of detecting objects from a predefined set of categories that frequently appear in real-world scenarios. The goal of this task is to evaluate the model's baisc ability to accurately identify and localize these common objects.

\begin{table*}[t]
    \centering
    \resizebox{1.0\textwidth}{!}{
    \begin{tabular}{cc|cc|cccccccccc}
\toprule
\multirow{2}{*}{Type} & \multirow{2}{*}{Method} & \multirow{2}{*}{Zero-Shot} & \multirow{2}{*}{\begin{tabular}[c]{@{}c@{}}Score \\ Thresh.\end{tabular}} & \multicolumn{10}{c}{COCO} \\
\cline{5-14}
 &  &  &  & \multicolumn{1}{c|}{mAP} & \begin{tabular}[c]{@{}c@{}}R@IoU\\ 0.5\end{tabular} & \begin{tabular}[c]{@{}c@{}}P@IoU\\ 0.5\end{tabular} & \multicolumn{1}{c|}{\begin{tabular}[c]{@{}c@{}}F1@IoU\\ 0.5\end{tabular}} & \begin{tabular}[c]{@{}c@{}}R@IoU\\ 0.95\end{tabular} & \begin{tabular}[c]{@{}c@{}}P@IoU\\ 0.95\end{tabular} & \multicolumn{1}{c|}{\begin{tabular}[c]{@{}c@{}}F1@IoU\\ 0.95\end{tabular}} & \begin{tabular}[c]{@{}c@{}}R@\\ mIoU\end{tabular} & \begin{tabular}[c]{@{}c@{}}P@\\ mIoU\end{tabular} & \begin{tabular}[c]{@{}c@{}}F1@\\ mIoU\end{tabular} \\
\midrule
\multirow{7}{*}{Closed-set}
& Faster RCNN-R50          & No  & 0.42 & \multicolumn{1}{c|}{38.4} & 60.4 & 60.7 & \multicolumn{1}{c|}{60.6} & 4.9  & 12.6 & \multicolumn{1}{c|}{7.1}  & 43.2 & 54.2 & 48.1 \\
& DETR-R50                 & No  & 0.78 & \multicolumn{1}{c|}{41.5} & 59.6 & 73.9 & \multicolumn{1}{c|}{65.9} & 10.6 & 19.0 & \multicolumn{1}{c|}{13.6} & 42.9 & 55.3 & 48.3 \\
& DyHead-R50               & No  & 0.24 & \multicolumn{1}{c|}{45.9} & 58.1 & 76.6 & \multicolumn{1}{c|}{66.1} & 11.9 & 20.6 & \multicolumn{1}{c|}{15.0} & 44.8 & 60.1 & 51.3 \\
& DAB-DETR-R50             & No  & 0.31 & \multicolumn{1}{c|}{44.4} & 59.2 & 77.4 & \multicolumn{1}{c|}{67.1} & 10.5 & 18.5 & \multicolumn{1}{c|}{13.4} & 43.8 & 58.8 & 50.2 \\
& Deformable-DETR-R50      & No  & 0.34 & \multicolumn{1}{c|}{49.4} & 62.6 & 78.5 & \multicolumn{1}{c|}{69.7} & 14.8 & 17.7 & \multicolumn{1}{c|}{17.7} & 48.7 & 62.4 & 54.7 \\
& DINO-R50                 & No  & 0.30 & \multicolumn{1}{c|}{51.7} & 62.6 & 76.5 & \multicolumn{1}{c|}{68.8} & 17.8 & 25.8 & \multicolumn{1}{c|}{21.1} & 50.0 & 62.4 & 55.6 \\
& DINO-Swin-L              & No  & 0.32 & \multicolumn{1}{c|}{59.6} & \textbf{69.8} & \textbf{82.5} & \multicolumn{1}{c|}{\textbf{75.6}} & \textbf{22.2} & \textbf{29.7} & \multicolumn{1}{c|}{\textbf{25.4}} & \textbf{57.0} & \textbf{68.2} & \textbf{62.1} \\
\midrule
Open-set & Grounding DINO-Swin-T & Yes & 0.37 & \multicolumn{1}{c|}{51.5} & 62.8 & 78.4 & \multicolumn{1}{c|}{69.8} & 20.5 & 26.2 & \multicolumn{1}{c|}{23.0} & 54.6 & 58.8 & 56.6 \\
\midrule
\multirow{10}{*}{MLLM}
& DeepSeek-VL2-Tiny         & UNK & - & \multicolumn{1}{c|}{-} & 29.4 & 55.0 & \multicolumn{1}{c|}{38.2} & 5.9  & 13.0 & \multicolumn{1}{c|}{8.1}  & 20.0 & 38.6 & 26.3 \\
& OVIS2.5-9B                & UNK & - & \multicolumn{1}{c|}{-} & 60.0 & 45.3 & \multicolumn{1}{c|}{51.6} & 8.4  & 8.0  & \multicolumn{1}{c|}{8.2}  & 39.3 & 31.0 & 34.6 \\
& Mimo-VL-7B                & UNK & - & \multicolumn{1}{c|}{-} & 56.2 & 56.9 & \multicolumn{1}{c|}{56.5} & 6.1  & 7.4  & \multicolumn{1}{c|}{6.7}  & 35.3 & 36.4 & 35.9 \\
& OVIS2.5-2B                & UNK & - & \multicolumn{1}{c|}{-} & 63.1 & 50.7 & \multicolumn{1}{c|}{56.2} & 10.4 & 10.1 & \multicolumn{1}{c|}{10.3} & 42.6 & 35.4 & 38.7 \\
& DeepSeek-VL2-Small        & UNK & - & \multicolumn{1}{c|}{-} & 52.1 & 73.3 & \multicolumn{1}{c|}{60.9} & 12.5 & \textbf{18.5} & \multicolumn{1}{c|}{14.9} & 39.1 & 55.4 & 45.9 \\
& Qwen2.5-VL-7B             & UNK & - & \multicolumn{1}{c|}{-} & 55.4 & 75.8 & \multicolumn{1}{c|}{64.0} & 10.6 & 15.5 & \multicolumn{1}{c|}{12.6} & 39.5 & 54.2 & 45.7 \\
& Qwen2.5-VL-3B             & UNK & - & \multicolumn{1}{c|}{-} & 55.7 & 77.2 & \multicolumn{1}{c|}{64.7} & 12.7 & \underline{18.3} & \multicolumn{1}{c|}{15.0} & 41.0 & 56.8 & 47.6 \\
& SEED1.5-VL                & Yes & - & \multicolumn{1}{c|}{-} & 65.3 & \textbf{78.6} & \multicolumn{1}{c|}{\underline{71.3}} & 12.7 & 16.4 & \multicolumn{1}{c|}{14.3} & 46.8 & \textbf{56.9} & \underline{51.4} \\
\rowcolor{lightblue!10}
& Rex-Omni-SFT              & Yes & - & \multicolumn{1}{c|}{-} & \underline{66.4} & 70.1 & \multicolumn{1}{c|}{68.2} & \textbf{14.8} & 17.0 & \multicolumn{1}{c|}{\underline{15.8}} & \underline{48.7} & 52.2 & 50.4 \\
\rowcolor{lightblue!10}
& Rex-Omni                  & Yes & - & \multicolumn{1}{c|}{-} & \textbf{68.1} & \underline{76.3} & \multicolumn{1}{c|}{\textbf{72.0}} & \underline{14.5} & 17.5 & \multicolumn{1}{c|}{\textbf{15.9}} & \textbf{49.8} & \underline{56.5} & \textbf{52.9} \\
\bottomrule
\end{tabular}
    }
    \caption{Evaluation results on the COCO benchmark for common object detection. Rex-Omni-SFT refers to the model after the first-stage supervised fine-tuning (SFT), while Rex-Omni is the final model after the second-stage GRPO-based reinforcement post-training. "UNK" signifies that the information was not reported in the respective papers.}
    \label{tab:common_object_detection}
\end{table*}

\textbf{Benchmark}: We conduct our evaluation on the COCO~\cite{Datasets:MSCOCO} dataset, one of the most widely used benchmarks in the field of object detection. The dataset includes 5,000 test images and spans 80 distinct object categories, representing a broad range of common objects.

\textbf{Evaluation Settings}: We evaluate two variants of our proposed model: \textbf{Rex-Omni-SFT}, which undergoes only the first stage of supervised fine-tuning, and the full \textbf{Rex-Omni} model, which undergoes both SFT and the subsequent GRPO-based reinforcement post-training. We compare these variants with three types of models: 1) Closed-set detection models trained on COCO, including Faster RCNN~\cite{ren2016faster}, DETR~\cite{carion2020end}, DyHead~\cite{dai2021dynamic}, DAB-DETR~\cite{liu2022dab}, Deformable-DETR~\cite{zhu2020deformable} and DINO~\cite{zhang2022dino}; 2) Open-set detection model Grounding DINO~\cite{liu2023grounding} that is not trained on COCO,  and 3) Multimodal large language models (MLLMs) including DeepSeek-VL2~\cite{wu2024deepseek}, Ovis2.5~\cite{lu2025ovis2}, MiMo-VL~\cite{coreteam2025mimovltechnicalreport}, Qwen2.5-VL~\cite{bai2025qwen2} and SEED1.5-VL~\cite{guo2025seed1}. For closed-set detection models, we input images and retain only the predicted bounding boxes whose categories match the ground-truth (GT) labels in each image. For open-set models, we provide all GT categories as text prompts and keep the corresponding results. For MLLMs, we adopt two prompting strategies: (1) querying one GT category at a time (e.g., “Detect dog in this image”), and (2) querying all GT categories simultaneously (e.g., “Detect dog, cat, person in this image”). Although the latter is more practical in real-world scenarios, most MLLMs exhibit a performance drop when handling multiple categories simultaneously. Therefore, except for SEED1.5-VL and Rex-Omni, we use the single-category strategy. All evaluations of Rex-Omni (both SFT and full versions) are conducted with a sampling temperature of 0 to minimize randomness.

\textbf{Metric}: In object detection, the standard metric is Average Precision (AP), which relies on confidence scores to compute precision and recall at varying thresholds. However, multimodal models often lack reliable confidence estimation, rendering AP unsuitable. We therefore adopt Recall, Precision, and F1 score as evaluation metrics. Given predicted and ground-truth boxes, Recall and Precision are computed per category and then averaged, while F1 is taken as their harmonic mean. Following COCO conventions, Intersection over Union (IoU) is evaluated at thresholds from 0.5 to 0.95 (step size 0.05), and results are reported at IoU=0.5, IoU=0.95, and the mean across thresholds. For fair comparison with MLLMs, we further compute F1 scores across confidence thresholds ranging from 0 to 1 (step size 0.01) for both closed- and open-set detection models, reporting the highest F1 as the final performance.

\begin{figure}[t]     
    \includegraphics[width=1.0\textwidth]{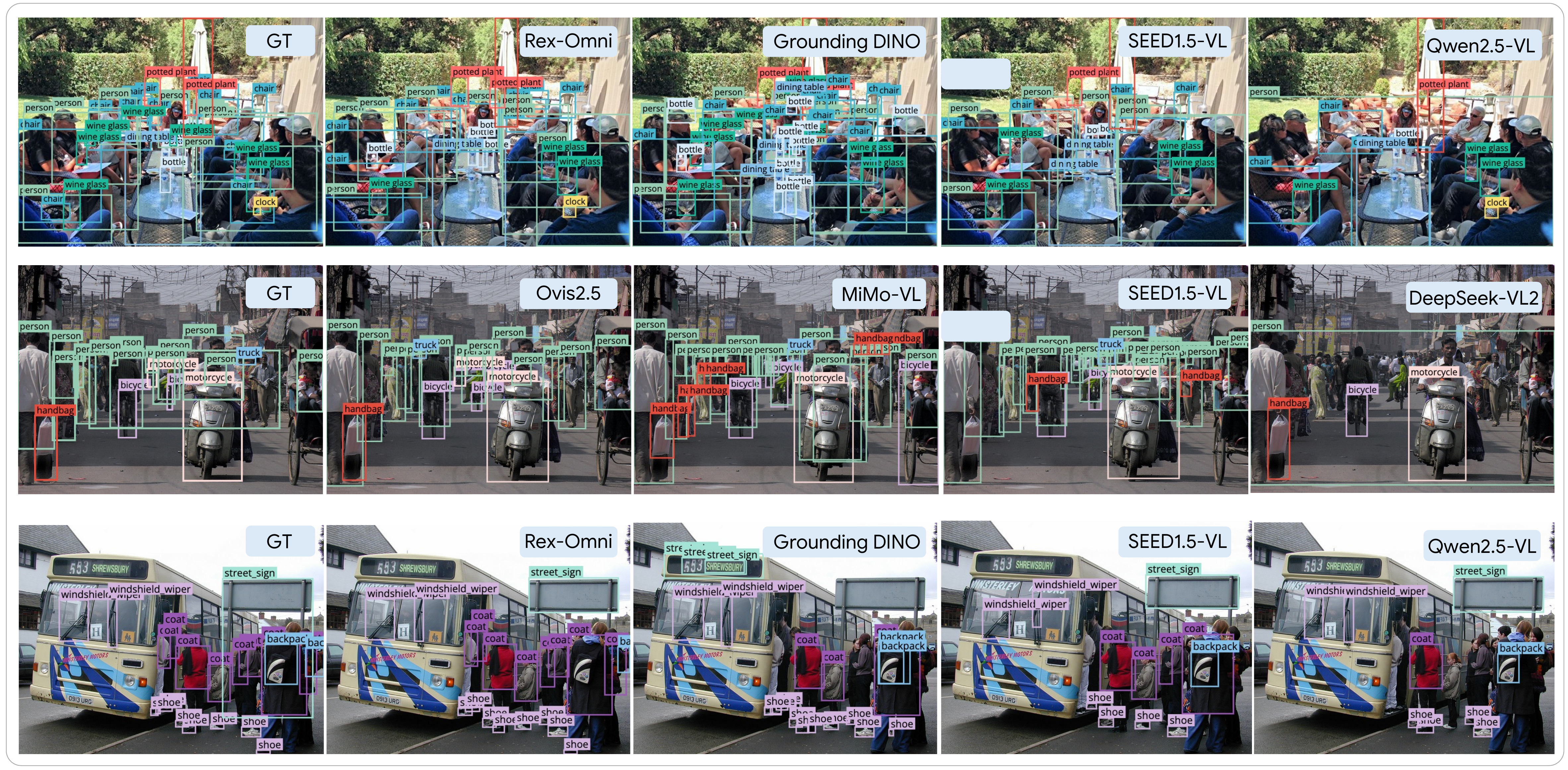}
    \caption{Visualization of detection predictions from different models on common and long-tailed object detection benchmarks, using COCO and LVIS, respectively.}
    \label{fig:compare_common_longtail}
\end{figure}

\textbf{Results}: The results are presented in Table \ref{tab:common_object_detection}. Firstly, among MLLMs, Rex-Omni surpasses existing approaches, including SEED1.5-VL, which previously held state-of-the-art detection performance. At an IoU threshold of 0.5, Rex-Omni demonstrates superior performance, outperforming both the open-set detection model Grounding DINO-SwinT and the closed-set detection model DINO-R50. Crucially, Rex-Omni achieves this in a zero-shot setting (without training on COCO data), thereby indicating that MLLM-based detection methods can indeed surpass traditional regression-based models when highly precise bounding box localization is not the sole critical factor. However, at a stricter IoU threshold of 0.95, Rex-Omni's performance, while still strong, only marginally outperforms DAB-DETR, suggesting that MLLMs may still lag behind conventional regression-based models in scenarios demanding extremely precise bounding box tightness.

Nevertheless, despite this nuanced limitation, the achieved performance is generally sufficient for a wide range of practical applications. We show some visualization results in Figure \ref{fig:compare_common_longtail}. Furthermore, a significant improvement is observed with GRPO post-training, where the full Rex-Omni model substantially outperforms its SFT-only variant (Rex-Omni-SFT). This clearly highlights the effectiveness of our reinforcement learning strategy. 

\subsection{Long-tailed Object Detection}
Long-tailed object detection tackles the challenge of recognizing categories with highly imbalanced instance distributions, where most categories are sparsely represented. This task requires models to generalize effectively and robustly detect rare objects in complex real-world scenarios.

\textbf{Benchmark: }We evaluate on the widely used LVIS~\cite{gupta2019lvis} dataset. LVIS comprises 1,203 categories, significantly more than COCO's 80, and features 19,626 test images.  Its categories are derived from WordNet synsets and are intentionally distributed to mimic real-world frequencies, resulting in a natural long-tail distribution where many categories have very few instances. 

\textbf{Evaluation Settings and Metrics: }We assess the performance of open-set detection models and MLLMs, following the same evaluation settings and metrics as described in Section~\ref{sec:coco} for COCO.

\begin{table*}[t]
    \centering
    \resizebox{1.0\textwidth}{!}{
        \begin{tabular}{cc|cc|ccccccccc}
\toprule
\multirow{2}{*}{Type} & \multirow{2}{*}{Method} & \multirow{2}{*}{Zero-Shot} & \multirow{2}{*}{\begin{tabular}[c]{@{}c@{}}Score\\ Thresh.\end{tabular}} & \multicolumn{9}{c}{LVIS} \\
\cline{5-13}
 &  &  &  & \begin{tabular}[c]{@{}c@{}}R@IoU\\ 0.5\end{tabular} & \begin{tabular}[c]{@{}c@{}}P@IoU\\ 0.5\end{tabular} & \multicolumn{1}{c|}{\begin{tabular}[c]{@{}c@{}}F1@IoU\\ 0.5\end{tabular}} & \begin{tabular}[c]{@{}c@{}}R@IoU\\ 0.95\end{tabular} & \begin{tabular}[c]{@{}c@{}}P@IoU\\ 0.95\end{tabular} & \multicolumn{1}{c|}{\begin{tabular}[c]{@{}c@{}}F1@IoU\\ 0.95\end{tabular}} & \begin{tabular}[c]{@{}c@{}}R@\\ mIoU\end{tabular} & \begin{tabular}[c]{@{}c@{}}P@\\ mIoU\end{tabular} & \begin{tabular}[c]{@{}c@{}}F1@\\ mIoU\end{tabular} \\
\midrule
Open-set                & Grounding DINO-Swin-T    & Yes & 0.21 & 39.1 & 61.2 & \multicolumn{1}{c|}{47.7} & 16.9 & \textbf{34.8} & \multicolumn{1}{c|}{\textbf{22.7}} & 31.9 & 49.6 & 38.8 \\
\midrule
\multirow{10}{*}{MLLM}
& DeepSeek-VL2-Tiny        & UNK & - & 22.4 & 55.7 & \multicolumn{1}{c|}{32.0} & 7.2  & 25.7 & \multicolumn{1}{c|}{11.2} & 14.2 & 41.9 & 21.2 \\
& MiMo-VL-7B               & UNK & - & 43.3 & 57.8 & \multicolumn{1}{c|}{49.5} & 6.5  & 13.7 & \multicolumn{1}{c|}{8.8}  & 25.4 & 41.1 & 31.4 \\
& OVIS2.5-2B               & UNK & - & 53.2 & 55.6 & \multicolumn{1}{c|}{54.4} & 13.3 & 19.4 & \multicolumn{1}{c|}{15.8} & 34.0 & 41.6 & 37.4 \\
& OVIS2.5-9B               & UNK & - & 52.2 & 54.1 & \multicolumn{1}{c|}{53.1} & 11.2 & 19.9 & \multicolumn{1}{c|}{14.4} & 32.2 & 40.1 & 35.8 \\
& Qwen2.5-VL-7B            & UNK & - & 47.1 & 74.3 & \multicolumn{1}{c|}{57.7} & 12.7 & 29.0 & \multicolumn{1}{c|}{17.6} & 32.0 & 54.3 & 40.2 \\
& Qwen2.5-VL-3B            & UNK & - & 44.8 & 73.9 & \multicolumn{1}{c|}{55.8} & 14.3 & \underline{29.8} & \multicolumn{1}{c|}{19.3} & 31.9 & 54.6 & 40.3 \\
& DeepSeek-VL2-Small       & UNK & - & 46.1 & 72.2 & \multicolumn{1}{c|}{56.2} & 15.8 & 31.2 & \multicolumn{1}{c|}{\underline{21.0}} & 33.7 & 55.0 & 41.8 \\
& SEED1.5-VL               & Yes & - & \underline{54.7} & \textbf{82.0} & \multicolumn{1}{c|}{\textbf{65.6}} & 15.0 & 28.1 & \multicolumn{1}{c|}{19.5} & \underline{38.5} & \textbf{59.3} & \underline{46.7} \\
\rowcolor{lightblue!10}
& Rex-Omni-SFT             & Yes & - & 52.0 & 71.6 & \multicolumn{1}{c|}{60.3} & \textbf{16.6} & 27.4 & \multicolumn{1}{c|}{20.7} & 37.7 & 53.3 & 44.2 \\
\rowcolor{lightblue!10}
& Rex-Omni                 & Yes & - & \textbf{54.7} & \underline{78.1} & \multicolumn{1}{c|}{\underline{64.3}} & \underline{16.5} & 27.6 & \multicolumn{1}{c|}{20.7} & \textbf{39.6} & \underline{57.5} & \textbf{46.9} \\
\bottomrule
\end{tabular}
    }
    \caption{Performance evaluation on the LVIS dataset for long-tailed object detection. All reported metrics adhere to the same methodology described for COCO evaluation in Section \ref{sec:coco}.}
    \label{tab:long_tailed_object_detection}
\end{table*}

\textbf{Results: }The results are presented in Table~\ref{tab:long_tailed_object_detection}. On LVIS, MLLMs generally outperform traditional open-set detectors such as Grounding DINO, owing to the stronger linguistic reasoning ability of their LLM components compared to conventional text encoders (e.g., CLIP or BERT). This advantage facilitates better generalization to low-frequency categories.

In the zero-shot setting, Rex-Omni achieves competitive performance, with an F1 score at IoU=0.5 second only to SEED1.5-VL, likely due to the latter’s larger model size and stronger language understanding. Notably, Rex-Omni attains state-of-the-art results on the mIoU metric, reflecting its superior bounding box precision across thresholds. Moreover, the substantial improvement from Rex-Omni-SFT to the full Rex-Omni model underscores the effectiveness of GRPO-based reinforcement post-training in enhancing object localization. Qualitative results are shown in Figure~\ref{fig:compare_common_longtail} and Figure~\ref{fig:app_common_longtailed}.

\subsection{Dense and Tiny Object Detection}
\label{sec:dense}
Dense and tiny object detection is crucial for applications such as remote sensing and object counting, requiring accurate localization of numerous small objects in crowded scenes. For MLLMs, this task is particularly challenging: it not only demands precise, extended coordinate predictions sensitive to subtle pixel variations, but also exposes the absence of multi-scale feature mechanisms (e.g., feature pyramids~\cite{lin2017feature}) that traditional regression-based detectors exploit to handle scale diversity. As a result, MLLMs often suffer from issues such as duplicate predictions and coordinate offsets in dense and tiny object detection scenarios.
 
\begin{table*}[t]
    \centering
    \resizebox{1.0\textwidth}{!}{
        \begin{tabular}{cc|c|ccc|ccc}
\toprule
\multirow{2}{*}{Type} & \multirow{2}{*}{Method} & \multirow{2}{*}{\begin{tabular}[c]{@{}c@{}}Score \\ Threshhold\end{tabular}}
& \multicolumn{3}{c|}{Dense200} & \multicolumn{3}{c}{VisDrone} \\
\cline{4-9}
 &  &  & \begin{tabular}[c]{@{}c@{}}F1@IoU\\ 0.5\end{tabular} & \begin{tabular}[c]{@{}c@{}}F1@IoU\\ 0.95\end{tabular} & \begin{tabular}[c]{@{}c@{}}F1@IoU\\ mIoU\end{tabular} & \begin{tabular}[c]{@{}c@{}}F1@IoU\\ 0.5\end{tabular} & \begin{tabular}[c]{@{}c@{}}F1@IoU\\ 0.95\end{tabular} & \begin{tabular}[c]{@{}c@{}}F1@IoU\\ mIoU\end{tabular} \\
\midrule
Open-set & Grounding DINO-Swin-T & 0.25 & 36.9 & \textbf{19.7} & 33.1 & 55.2 & \textbf{3.9} & \textbf{38.5} \\
\midrule
\multirow{10}{*}{MLLM}
& DeepSeek-VL2-Tiny  & - & 2.2  & 0.3 & 1.5 & 4.3  & 0.1 & 1.8 \\
& OVIS2.5-9B         & - & 14.0 & 0.0 & 5.1 & 15.8 & 0.1 & 6.5 \\
& OVIS2.5-2B         & - & 17.9 & 0.0 & 6.7 & 21.0 & 0.1 & 9.2 \\
& MiMo-VL-7B         & - & 29.7 & 0.4 & 15.9 & 27.7 & 0.3 & 14.3 \\
& Qwen2.5-VL-3B      & - & 0.8  & 0.1 & 0.5 & 31.5 & 1.9 & 20.4 \\
& Qwen2.5-VL-7B      & - & 1.1  & 0.1 & 0.6 & 34.5 & 1.6 & 21.7 \\
& DeepSeek-VL2-Small & - & 16.0 & 3.9 & 12.7 & 35.8 & 1.7 & 23.3 \\
& SEED1.5-VL         & - & \underline{76.9} & 5.3 & \underline{53.2} & \underline{55.9} & 0.6 & 27.4 \\
\rowcolor{lightblue!10}
& Rex-Omni-SFT       & - & 60.2 & \underline{10.6} & 46.4 & 55.6 & \underline{1.9} & 32.4 \\
\rowcolor{lightblue!10}
& Rex-Omni           & - & \textbf{78.4} & 10.3 & \textbf{58.3} & \textbf{61.6} & 1.5 & \underline{35.8} \\
\bottomrule
\end{tabular}
    }
    \caption{Evaluation Results on dense object detection benchmark VisDrone and Dense200. We report the same metric used in COCO evaluation at Section \ref{sec:coco}.}
    \label{tab:dense_object_detection}
\end{table*}

\begin{figure}[!h]     
    \includegraphics[width=1.0\textwidth]{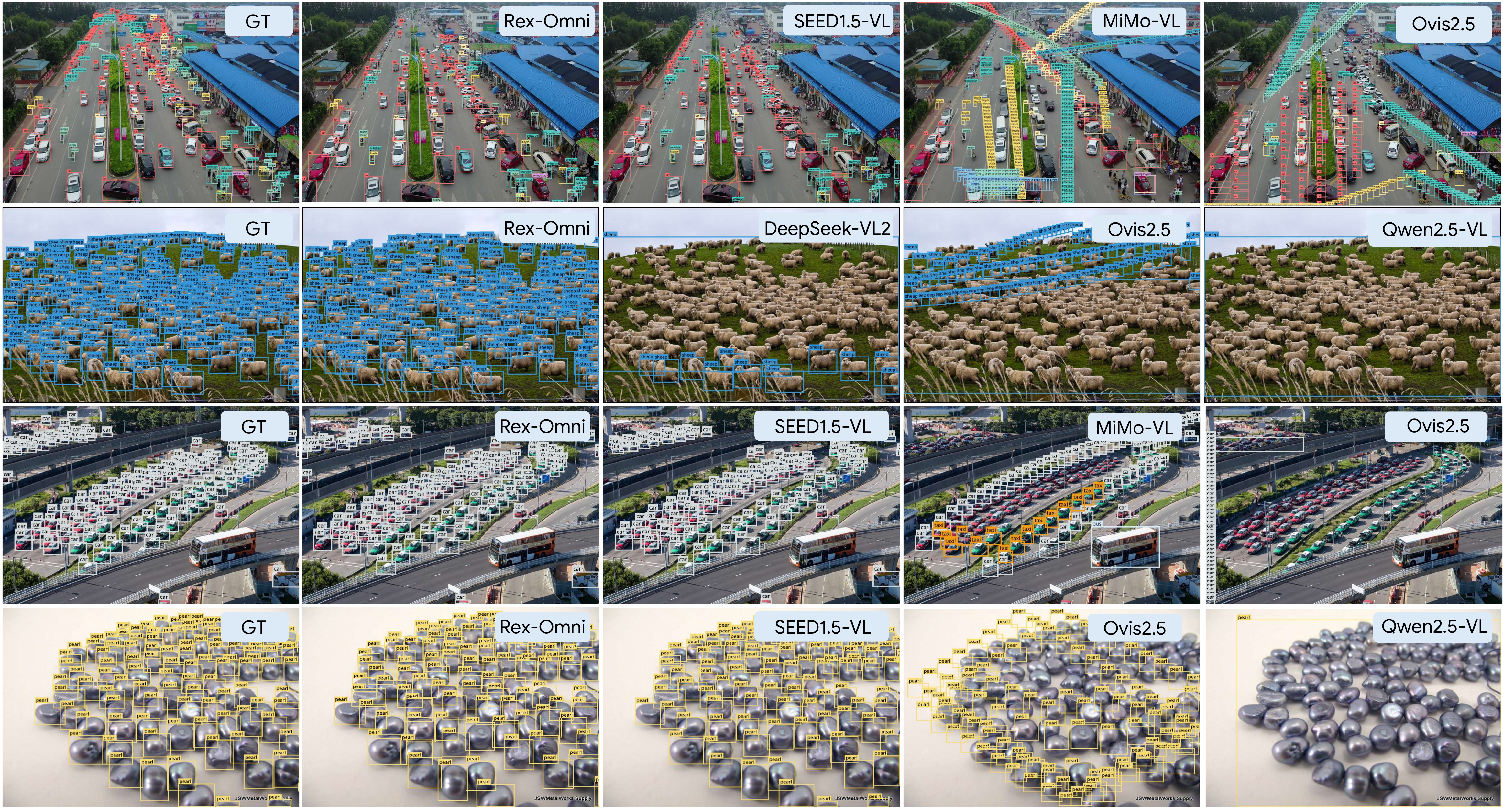}
    \caption{Visualization of dense and tiny object detection predictions. This figure presents a qualitative comparison of various models on the VisDrone and Dense200 datasets.}
    \label{fig:compare_dense}
\end{figure}

\noindent\textbf{Benchmark, settings and metrics: }We evaluate open-set detection models and MLLMs on two distinct datasets tailored for dense and tiny object detection. The first dataset, VisDrone~\cite{du2019visdrone}, comprises 1,610 aerial traffic images spanning 10 categories, with individual boxes measuring approximately 30.7×32.4 pixels. Additionally, we introduce Dense200, a manually collected dataset consisting of 200 densely annotated images covering 109 categories. In Dense200, each image contains an average of 91.2 bounding boxes, with an average size of  66.8×64.5 pixels. Together, these datasets pose significant challenges due to the combination of small object sizes and high object density, demanding precise spatial reasoning and accurate localization. The evaluation settings and metrics are the same as those used in Section \ref{sec:coco} for COCO evaluation.

\noindent\textbf{Results}: The results are reported in Table~\ref{tab:dense_object_detection}, with representative visualizations in Figure~\ref{fig:compare_dense} and Figure~\ref{fig:app_dense}. As anticipated in Section \ref{sec:sft_limitation}, MLLMs struggle with dense and tiny object detection, with most models showing poor performance. We identify two critical failure modes: (1) \textbf{Large-box prediction}, where a single oversized bounding box erroneously covers multiple adjacent objects, and (2) \textbf{Structured duplicate predictions}, where repeated coordinates with minimal offsets are generated instead of distinct object boxes.

We attribute these issues to the SFT stage. Teacher-forced training on full ground-truth sequences restricts the model’s ability to regulate its own output structure. Without such guidance at inference, the model fails to decide object counts or avoid redundant predictions. Notably, we also observed these problematic repetitive predictions in our SFT-only variant. Crucially, after GRPO-based reinforcement post-training, these duplication issues largely disappear, compellingly demonstrating the effectiveness of our two-stage pipeline in correcting SFT-induced deficiencies and enabling more coherent, accurate predictions in dense and tiny object scenarios.

\begin{table*}[t]
    \centering
    \resizebox{1.0\textwidth}{!}{
        \begin{tabular}{cc|c|ccc|ccc|ccc}
\toprule
\multirow{2}{*}{Type} & \multirow{2}{*}{Method} & \multirow{2}{*}{\begin{tabular}[c]{@{}c@{}}Score \\ Thresh.\end{tabular}} 
& \multicolumn{3}{c|}{HumanRef} 
& \multicolumn{3}{c|}{RefCOCOg val} 
& \multicolumn{3}{c}{RefCOCOg test} \\
\cline{4-12}
 &  &  & \begin{tabular}[c]{@{}c@{}}F1@IoU\\ 0.5\end{tabular} & \begin{tabular}[c]{@{}c@{}}F1@IoU\\ 0.95\end{tabular} & \begin{tabular}[c]{@{}c@{}}F1@IoU\\ mIoU\end{tabular} 
 & \begin{tabular}[c]{@{}c@{}}F1@IoU\\ 0.5\end{tabular} & \begin{tabular}[c]{@{}c@{}}F1@IoU\\ 0.95\end{tabular} & \begin{tabular}[c]{@{}c@{}}F1@IoU\\ mIoU\end{tabular} 
 & \begin{tabular}[c]{@{}c@{}}F1@IoU\\ 0.5\end{tabular} & \begin{tabular}[c]{@{}c@{}}F1@IoU\\ 0.95\end{tabular} & \begin{tabular}[c]{@{}c@{}}F1@IoU\\ mIoU\end{tabular} \\
\midrule
Open-set & Grounding DINO-Swin-T & 0.25 
& 28.0 & 16.5 & 25.2 
& 52.9 & 20.9 & 45.9 
& 53.8 & 22.9 & 46.8 \\
\midrule
\multirow{10}{*}{MLLM}
& DeepSeek-VL2-Tiny & - 
& 39.1 & 16.9 & 31.4 
& 67.4 & 16.1 & 50.5 
& 69.3 & 16.9 & 52.1 \\
& OVIS2.5-2B & - 
& 70.6 & 12.3 & 50.0 
& 87.4 & 29.3 & 73.4 
& 87.6 & 30.5 & 73.8 \\
& OVIS2.5-9B & - 
& 73.1 & 12.4 & 52.8 
& \underline{88.8} & 23.5 & 72.1 
& \textbf{88.7} & 24.2 & 72.6 \\
& Qwen2.5-VL-3B & - 
& 66.7 & 46.8 & 60.5 
& 83.5 & 30.7 & 69.2 
& 83.8 & 31.8 & 70.1 \\
& MiMo-VL-7B & - 
& 77.6 & 26.4 & 63.4 
& 84.9 & 14.4 & 65.3 
& 84.6 & 14.9 & 65.5 \\
& Qwen2.5-VL-7B & - 
& 72.9 & 42.9 & 64.1 
& 86.2 & 27.2 & 70.0 
& 85.7 & 28.4 & 70.4 \\
& DeepSeek-VL2-Small & - 
& 72.0 & 46.5 & 64.7 
& \textbf{92.4} & \textbf{45.6} & \textbf{81.4} 
& \textbf{91.8} & \textbf{47.0} & \textbf{81.6} \\
& SEED1.5-VL & - 
& \textbf{88.2} & 60.0 & \textbf{81.6} 
& 84.7 & 30.9 & 71.9 
& 85.2 & 32.1 & 73.2 \\
\rowcolor{lightblue!10}
& Rex-Omni-SFT & - 
& 83.3 & \underline{64.3} & 77.9 
& 84.9 & 34.2 & 71.7 
& 85.2 & 35.2 & 72.4 \\
\rowcolor{lightblue!10}
& Rex-Omni & - 
& \underline{85.4} & \textbf{65.4} & \underline{79.9} 
& 86.6 & \underline{35.3} & \underline{73.6} 
& 86.8 & \underline{36.6} & \underline{74.3} \\
\bottomrule
\end{tabular}
    }
    \caption{Evaluation results on referring expression comprehension benchmarks, including HumanRef and RefCOCOg.}
    \label{tab:referring_detection}
\end{table*}

\begin{figure}[!h]     
    \includegraphics[width=1.0\textwidth]{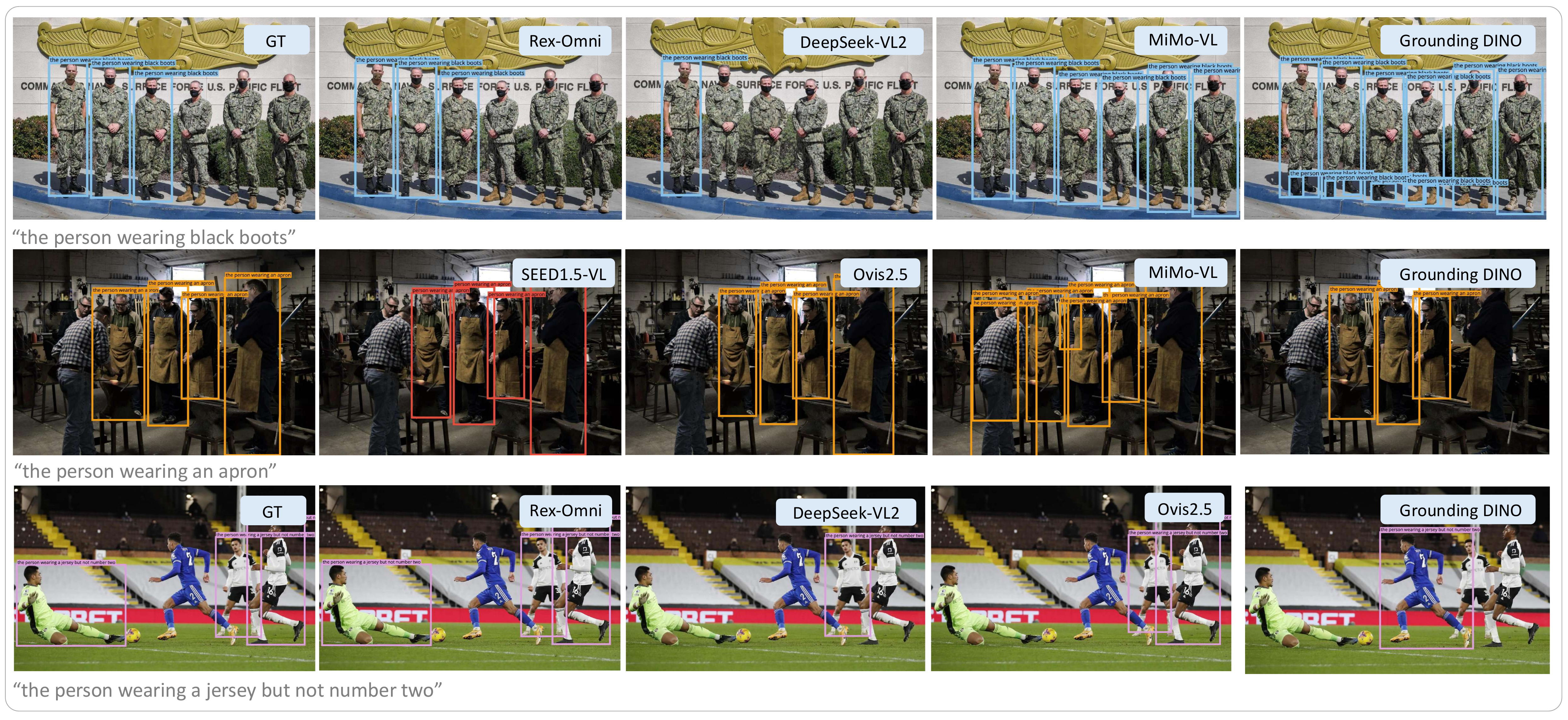}
    \captionsetup{justification=centering}
    \caption{Visualization of model predictions on referring object detection benchmarks.}
    \label{fig:compare_referring}
\end{figure}

\subsection{Referring Object Detection}
Referring object detection requires a model to identify and localize objects described by a natural language expression. Unlike standard object detection, which focuses on category-level recognition, this task demands fine-grained language understanding and strong alignment between linguistic descriptions and visual content.

\textbf{Benchmark: }The evaluation is conducted on two established public benchmarks:
\textbf{1) RefCOCOg (val/test):} RefCOCOg~\cite{Datasets:REFCOCOG} is built on COCO images and includes 4,889 validation and 9,577 test referring expressions. Each expression maps to a single ground-truth bounding box, making this benchmark relatively straightforward for evaluation.
\textbf{2) HumanRef:} HumanRef~\cite{jiang2025referringperson} is a human-annotated benchmark focused on people, comprising 6,000 test expressions organized into six subsets: attribute, position, interaction, reasoning, celebrity, and rejection. We use the first five subsets (5,000 images) for evaluation. Unlike RefCOCOg, a single expression in HumanRef may correspond to multiple ground-truth boxes, averaging two per expression. This design poses greater challenges, requiring both fine-grained language understanding and robust visual perception.

\textbf{Evaluation Settings and Metrics}: We evaluate open-set detection models and MLLMs using the same settings and metrics as in Section~\ref{sec:coco} for COCO, with one exception: during testing, models are queried with a single referring expression at a time. For the open-set detector Grounding DINO, we adopt its official demo confidence threshold of 0.25

\textbf{Results}: The results are presented in Table \ref{tab:referring_detection}. Open-set detection models notably struggle with this task, as evidenced by Grounding DINO’s consistent underperformance across benchmarks. In stark contrast, MLLMs, leveraging their inherent strong language understanding capabilities, consistently excel at this task. On HumanRef, Rex-Omni achieves competitive results, ranking second only to SEED1.5-VL. This indicates that, while Rex-Omni (3B parameters) possesses sufficient language understanding for effective REC, larger models like SEED1.5-VL benefit from greater capacity for more nuanced reasoning. Overall, Rex-Omni’s strong performance across all datasets demonstrates its ability to align natural language with visual content, making it highly practical for real-world referring scenarios. Visualization examples are shown in Figure~\ref{fig:compare_referring} and and Figure~\ref{fig:app_referring}.

\begin{table*}[t]
    \centering
    \resizebox{\textwidth}{!}{
       \begin{tabular}{c|c|cccc|cccc|cccc|cccc}
\toprule
\multirow{2}{*}{Type} & \multirow{2}{*}{Method} 
& \multicolumn{4}{c|}{FSC147-test} 
& \multicolumn{4}{c|}{Dense200} 
& \multicolumn{4}{c|}{COCO} 
& \multicolumn{4}{c}{LVIS} \\ 
\cline{3-18}
 &  & \begin{tabular}[c]{@{}c@{}}F1@\\ 0.5\end{tabular} 
 & \begin{tabular}[c]{@{}c@{}}F1@\\ 0.95\end{tabular} 
 & \begin{tabular}[c]{@{}c@{}}F1@\\ mIoU\end{tabular} 
 & MAE 
 & \begin{tabular}[c]{@{}c@{}}F1@\\ 0.5\end{tabular} 
 & \begin{tabular}[c]{@{}c@{}}F1@\\ 0.95\end{tabular} 
 & \begin{tabular}[c]{@{}c@{}}F1@\\ mIoU\end{tabular} 
 & MAE 
 & \begin{tabular}[c]{@{}c@{}}F1@\\ 0.5\end{tabular} 
 & \begin{tabular}[c]{@{}c@{}}F1@\\ 0.95\end{tabular} 
 & \begin{tabular}[c]{@{}c@{}}F1@\\ mIoU\end{tabular} 
 & MAE 
 & \begin{tabular}[c]{@{}c@{}}F1@\\ 0.5\end{tabular} 
 & \begin{tabular}[c]{@{}c@{}}F1@\\ 0.95\end{tabular} 
 & \begin{tabular}[c]{@{}c@{}}F1@\\ mIoU\end{tabular} 
 & MAE \\ 
\midrule
\multirow{3}{*}{Counting} 
& BMNet+~\cite{shi2022representcomparelearnsimilarityaware} & - & - & - & 14.6 & - & - & - & - & - & - & - & - & - & - & - & - \\
& CountTR~\cite{liu2023countrtransformerbasedgeneralisedvisual} & - & - & - & 12.0 & - & - & - & - & - & - & - & - & - & - & - & - \\
& DAVE~\cite{pelhan2024davedetectandverifyparadigm} & - & - & - & 8.7 & - & - & - & - & - & - & - & - & - & - & - & - \\
\midrule
Open-set 
& T-Rex2~\cite{jiang2024t} 
& \textbf{91.5} & \textbf{47.5} & \textbf{73.3} & 10.9 
& \textbf{93.9} & \textbf{67.1} & \textbf{88.1} & \textbf{6.5} 
& 72.3 & \textbf{19.9} & 57.8 & \textbf{4.0} 
& \underline{71.1} & \textbf{34.6} & \textbf{58.8} & \textbf{3.4} \\
\midrule
\multirow{2}{*}{MLLM} & \cellcolor{lightblue!10} Rex-Omni-SFT 
& \cellcolor{lightblue!10}\underline{87.0} &\cellcolor{lightblue!10} \cellcolor{lightblue!10}\underline{10.5} & \cellcolor{lightblue!10}\underline{64.6} & \cellcolor{lightblue!10}\underline{7.8} 
& \cellcolor{lightblue!10}65.1 & \cellcolor{lightblue!10}11.1 &\cellcolor{lightblue!10} 50.0 &\cellcolor{lightblue!10} 50.9 
&\cellcolor{lightblue!10} \underline{73.6} & \cellcolor{lightblue!10}\underline{19.8} & \cellcolor{lightblue!10}\underline{58.4} &\cellcolor{lightblue!10} 11.7 
& \cellcolor{lightblue!10}70.2 & \cellcolor{lightblue!10}\underline{26.3} & \cellcolor{lightblue!10}56.0 &\cellcolor{lightblue!10} 11.7 \\
 & \cellcolor{lightblue!10}Rex-Omni 
& \cellcolor{lightblue!10}86.0 & \cellcolor{lightblue!10}10.5 & \cellcolor{lightblue!10}62.8 & \cellcolor{lightblue!10}\textbf{7.0} 
& \cellcolor{lightblue!10}\underline{79.5} & \cellcolor{lightblue!10}\underline{10.8} & \cellcolor{lightblue!10}\underline{59.2} & \cellcolor{lightblue!10}\underline{20.1} 
& \cellcolor{lightblue!10}\textbf{79.1} & \cellcolor{lightblue!10}19.4 & \cellcolor{lightblue!10}\textbf{61.3} & \cellcolor{lightblue!10}\underline{4.5} 
& \cellcolor{lightblue!10}\textbf{74.4} & \cellcolor{lightblue!10}24.8 & \cellcolor{lightblue!10}\underline{57.0} & \cellcolor{lightblue!10}\underline{5.2} \\
\bottomrule
\end{tabular}
    }
    \caption{Evaluation results for the visual prompting task across COCO, LVIS, Dense200, and FSC147 datasets. Performance is assessed using F1-score for detection and Mean Absolute Error (MAE) for object counting.}
    \label{tab:visual_prompting}
\end{table*}

\subsection{Visual Prompting}

While text prompts are widely used in many tasks, they have inherent limitations, particularly when certain objects are difficult to describe using language alone. In such scenarios, visual prompting can offer a more effective approach for object detection. In this section, we define visual prompting as a task where, given an image alongside several example bounding boxes within it, the model is required to detect all other objects belonging to the same category as those indicated by the examples.

\textbf{Benchmark and Evaluation Setting:}
We evaluate visual prompting on the object counting dataset FSC147~\cite{ranjan2021learning}, as well as on object detection benchmarks COCO, LVIS, and Dense200. The FSC147 dataset consists of 1,190 images, each containing a dense set of objects from a single category along with three example bounding boxes, which are used as visual prompts for detection. For COCO, LVIS, and Dense200, we follow the T-Rex2~\cite{jiang2024t} methodology, where for each ground-truth category in an image, one bounding box is randomly sampled as the visual prompt for that category. To interface with Rex-Omni, the coordinates of the selected visual prompt box are converted into special tokens and embedded into the query, e.g., “Given reference boxes <12><52><212><337> indicating one or more objects, find all objects of the same category in the image.”

\textbf{Metric:}
We primarily adopt the F1-score as described in Section~\ref{sec:coco} for object detection. Additionally, we introduce the Mean Absolute Error (MAE) metric to evaluate the model’s object counting ability. MAE is computed as the absolute difference between predicted and ground-truth object counts, averaged across the entire dataset, thereby providing an additional measure of the model’s capability to accurately count objects in dense scenes.

\begin{figure}[!h]     
    \centering
    \includegraphics[width=\textwidth]{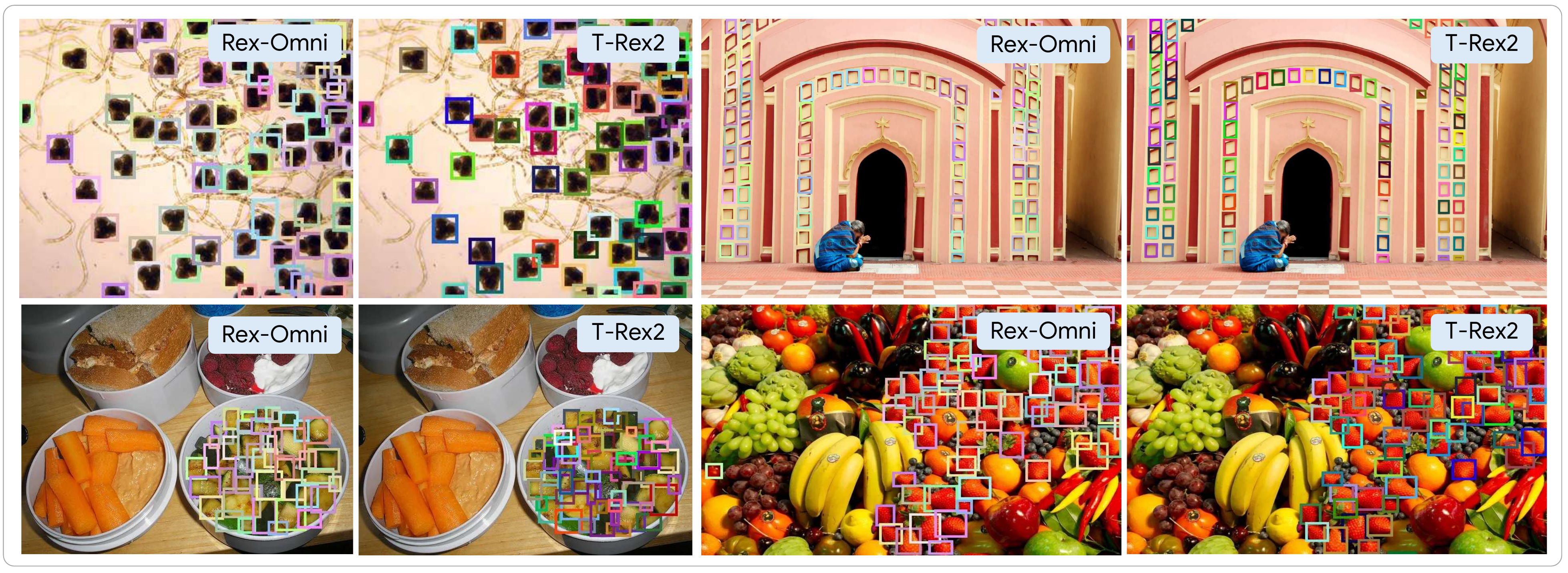}
    \captionsetup{justification=centering}
    \caption{Qualitative comparison of visual prompting predictions between T-Rex2 and Rex-Omni.}
    \label{fig:compare_visual_prompt}
\end{figure}

\textbf{Results:}
While Rex-Omni’s overall performance still falls short of the traditional expert model T-Rex2, it demonstrates strong visual prompting capabilities. In particular, Rex-Omni performs well in both dense scenes and long-tailed scenarios, highlighting its effectiveness in addressing high object density and severe class imbalance. Representative visualization results are shown in Figure~\ref{fig:compare_visual_prompt}.

\subsection{Object Pointing}
The object pointing task requires models to predict precise point coordinates for specified target objects. Unlike bounding boxes, point annotations offer greater flexibility in localization, as models can indicate an object’s center or any representative position within its boundaries.

\textbf{Benchmarks, Evaluation Settings and Metric: }To evaluate object pointing, we integrate datasets previously used for box-based detection, including COCO, LVIS, Dense200, VisDrone, RefCOCOg, and HumanRef. This collection covers a broad range of visual scenarios, from common and long-tailed objects to dense small objects and complex referring expressions. The evaluation protocol follows that of the earlier detection tasks. For most models, except SEED1.5-VL and Rex-Omni, each test image is queried with a single ground-truth category at a time. We adopt the same F1-score–based evaluation metric as in object detection, with one modification in the matching criterion. For each ground-truth bounding box, we generate a segmentation mask using SAM, and a predicted point is considered correct if it falls within the corresponding mask. Recall, Precision, and F1 are then computed analogously to standard box-based evaluation.

\begin{table*}[t]
    \centering
    \resizebox{1.0\textwidth}{!}{
       \begin{tabular}{c|cc|cc|ccc}
\toprule
                         & COCO                              & LVIS                              & Dense200                    & VisDrone      & HumanRef      & RefCOCOg val  & RefCOCOg test \\ \cline{2-8} 
\multirow{-2}{*}{Method} & F1@Point                          & F1@Point                          & F1@Point                    & F1@Point      & F1@Point      & F1@Point      & F1@Point      \\ \midrule
OVIS2.5-2B               & 73.4                              & 52.8                              & 36.4                        & 23.8          & 72.5          & 83.1          & 83.1          \\
Qwen2.5-VL-3B            & 65.9                              & 48.3                              & 4.3                         & 13.9          & 64.1          & 77.4          & 77.8          \\
Qwen2.5-VL-7B            & 61.1                              & 56.5                              & 2.0                         & 14.2          & 65.1          & 78.9          & 79.4          \\
OVIS2.5-9B               & 72.6                              & 61.7                              & 35.0                        & 18.8          & 62.3          & \textbf{85.0} & {\underline{84.5}}    \\
Molmo-7B-D               & 77.3                              & 40.3                              & 33.1                        & 29.2          & 70.0          & 83.7          & 83.6          \\
SEED1.5-VL               & {\color[HTML]{000000} {\underline{78.2}}} & {\color[HTML]{000000} {\underline{70.7}}} & {\color[HTML]{000000} 72.1} & {\underline{56.7}}    & {\underline{83.1}}    & 83.6          & 84.2          \\
  \rowcolor{lightblue!10}  Rex-Omni-SFT             & 76.0                              & 66.7                              & {\underline{72.9}}                  & 49.5          & 82.1          & 83.3          & 83.9          \\
 \rowcolor{lightblue!10} Rex-Omni                 & \textbf{80.5}                     & \textbf{70.8}                     & \textbf{82.5}               & \textbf{58.9} & \textbf{83.8} & {\underline{84.7}}    & \textbf{85.1} \\ \bottomrule
\end{tabular}
    }
    \centering
    \caption{Performance evaluation for the object pointing task across a diverse range of benchmarks (COCO, LVIS, Dense200, VisDrone, RefCOCOg, HumanRef). F1-scores are used as the primary metric.}
    \label{tab:pointing}
    \vspace{-1em}
\end{table*}

\begin{figure}[t]     
    \centering
    \includegraphics[width=\textwidth]{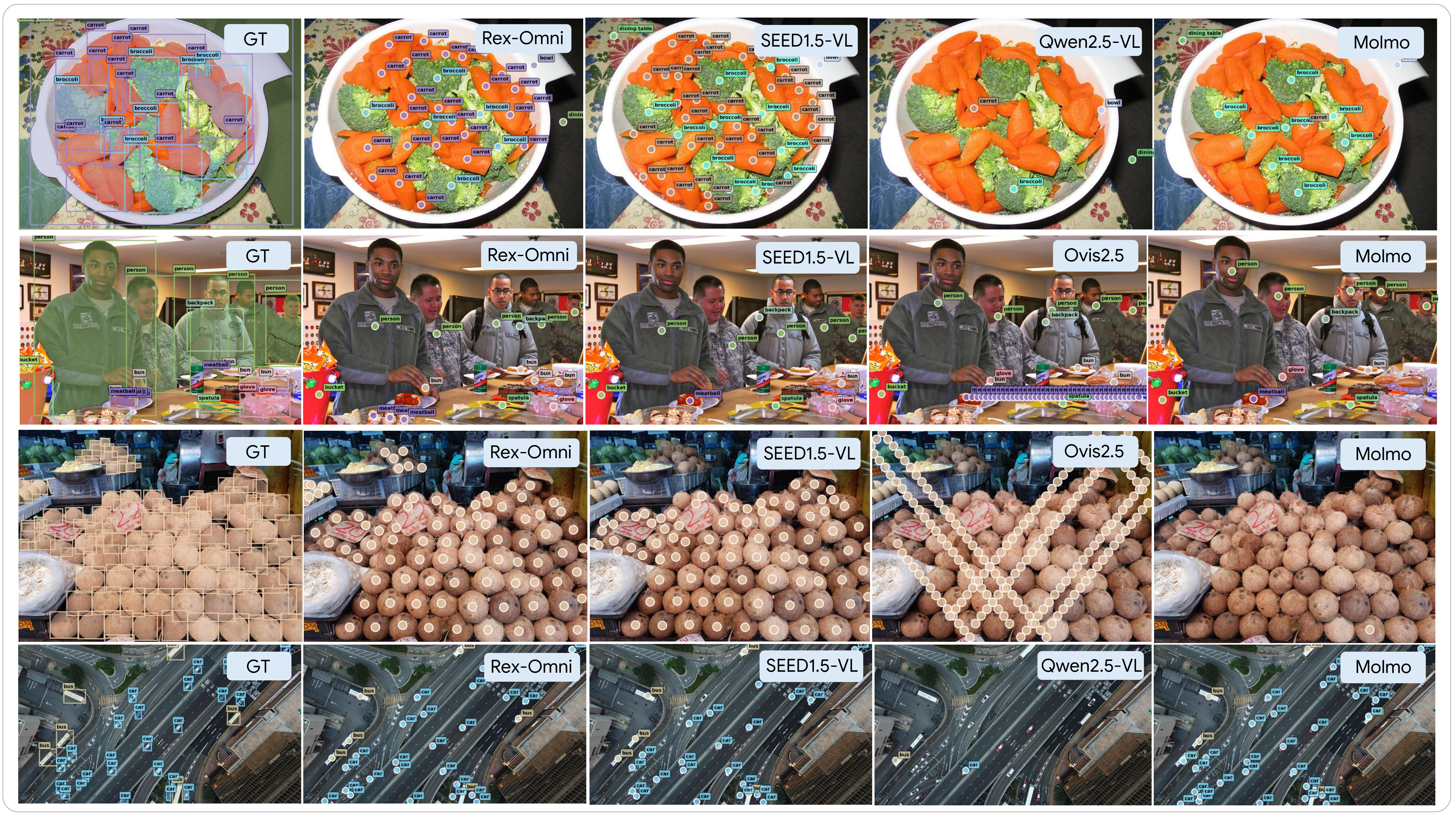}
    \captionsetup{justification=centering}
    \caption{Qualitative comparison of object pointing predictions from different models.}
    \label{fig:compare_pointing}
\end{figure}

\textbf{Results: }The performance of all evaluated models is reported in Table~\ref{tab:pointing}. While most MLLMs achieve reasonable pointing accuracy on common object categories, they struggle with dense or small-scale instances, particularly on Dense200 and VisDrone. Rex-Omni attains the highest F1-scores across both general and challenging datasets, highlighting its strong spatial localization ability. Representative visualizations are shown in Figure~\ref{fig:compare_pointing} and and Figure~\ref{fig:app_pointing}.

\begin{table*}[t]
    \centering
    \resizebox{1.0\textwidth}{!}{
       \begin{tabular}{c|ccccccc|ccccccccccccc}
\toprule
\multirow{3}{*}{Method} & \multicolumn{7}{c|}{ScreenSpot-V2}                                                                                   & \multicolumn{13}{c}{ScreenSpot-Pro}                                                                                                                                                                           \\ \cline{2-21} 
                        & \multicolumn{2}{c}{Mobile}    & \multicolumn{2}{c}{Desktop}   & \multicolumn{2}{c}{Web}       & \multirow{2}{*}{Avg} & \multicolumn{2}{c}{Dev.}      & \multicolumn{2}{c}{Creative}  & \multicolumn{2}{c}{CAD}       & \multicolumn{2}{c}{Sci.}      & \multicolumn{2}{c}{Office}    & \multicolumn{2}{c}{OS}        & Avg           \\ \cline{2-7} \cline{9-21} 
                        & Text          & Icon          & Text          & Icon          & Text          & Icon          &                      & Text          & Icon          & Text          & Icon          & Text          & Icon          & Text          & Icon          & Text          & Icon          & Text          & Icon          &               \\ \midrule
UI-TARS-2B              & 95.2          & 79.1          & 90.7          & 68.6          & 87.2          & 78.3          & 84.7                 & 47.4          & 4.1           & 42.9          & 6.3           & 17.8          & 4.7           & 56.9          & 17.3          & 50.3          & 17.0          & 21.5          & 5.6           & 27.7          \\
Qwen2.5-VL-3B           & 93.4          & 73.5          & 88.1          & 58.6          & {\underline{88.0}}    & 71.4          & 80.9                 & 38.3          & 3.4           & 40.9          & 4.9           & 22.3          & 6.3           & 44.4          & 10.0          & 48.0          & 17.0          & 33.6          & 4.5           & 25.9          \\
UI-R1-3B                & 84.3          & \textbf{96.2} & 75.4          & \textbf{89.2} & 63.6          & \textbf{92.3} & 85.4                 & 22.7          & 4.1           & 27.3          & 3.5           & 11.2          & 6.3           & 42.4          & 11.8          & 32.2          & 11.3          & 13.1          & 4.5           & 17.8          \\
InfiGUI-R1-3B           & -             & -             & -             & -             & -             & -             & -                    & 51.3          & {\underline{12.4}}    & 44.9          & 7.0           & 33.0          & \textbf{14.1} & 58.3          & {\underline{20.0}}    & \textbf{65.5} & 28.3          & \textbf{43.9} & {\underline{12.4}}    & 35.7          \\
SE-GUI-3B               & -             & -             & -             & -             & -             & -             & -                    & 55.8          & 7.6           & 47.0          & 4.9           & \textbf{38.1} & {\underline{12.5}}    & \textbf{61.8} & 16.4          & 59.9          & 24.5          & 40.2          & {\underline{12.4}}    & 35.9          \\
JEDI-3B                 & \textbf{96.6} & 81.5          & {\underline{96.9}}    & 78.6          & \textbf{88.5} & {\underline{83.7}}    & \textbf{88.6}        & 61.0          & \textbf{13.8} & \textbf{53.5} & 8.4           & {\underline{27.4}}    & 9.4           & 54.2          & 18.2          & {\underline{64.4}}    & \textbf{32.1} & {\underline{38.3}}    & 9.0           & {\underline{36.1}}    \\
\rowcolor{lightblue!10} Rex-Omni-SFT            & {\underline{95.5}}    & 80.6          & \textbf{97.4} & 77.1          & 85.5          & 76.4          & 86.4                 & 54.6          & 10.3          & 46.5          & {\underline{10.5}}    & 22.3          & 7.8           & 55.6          & 19.1          & 55.9          & 20.8          & 37.4          & 11.2          & 32.6          \\
\rowcolor{lightblue!10}Rex-Omin-GRPO           & 93.3          & {\underline{84.3}}    & 96.4          & {\underline{84.3}}    & 86.8          & 80.5          & {\underline{88.4}}           & \textbf{61.7} & 9.7           & {\underline{52.5}}    & \textbf{12.6} & 22.3          & 9.4           & {\underline{59.0}}    & \textbf{26.4} & 63.3          & {\underline{28.3}}    & 24.1          & \textbf{15.7} & \textbf{36.8} \\ \bottomrule
\end{tabular}
    }
    \captionsetup{justification=centering}
    \caption{Evaluation results for GUI Grounding task on the ScreenSpot-V2 and ScreenSpot Pro datasets.}
    \label{tab:gui_grounidng}
\end{table*}

\subsection{GUI Grounding}
Graphical User Interface (GUI) grounding evaluates a model’s ability to localize specific UI elements based on natural language queries. This task is critical for applications such as intelligent agents, automated UI interaction, and software testing, as it requires seamless integration of visual perception and language understanding.

\textbf{Benchmarks, Evaluation Setting and Metric: }We evaluate models on two datasets: ScreenSpot-V2~\cite{wu2024atlas} and ScreenSpot-Pro~\cite{li2025screenspot}. ScreenSpot-V2 encompasses mobile, desktop, and web scenarios, featuring a diverse array of UI layouts across 1,272 images. ScreenSpot-Pro, conversely, focuses on ultra-high-resolution interfaces, specifically designed to test the model’s precision in localizing UI elements under highly challenging visual conditions, comprising 1,581 images. Rex-Omni is assessed using its point-based prediction capability, outputting a point within the target UI element for each query. Following standard protocols, we report accuracy, considering a prediction correct if the point falls within the ground-truth bounding box.

\textbf{Results: }As shown in Table \ref{tab:gui_grounidng}, Rex-Omni consistently demonstrates strong performance on GUI grounding tasks. Specifically, among 3B-parameter models, Rex-Omni achieves the highest accuracy across both ScreenSpot V2 and ScreenSpot Pro. This underscores its superior capability to seamlessly integrate robust language understanding with fine-grained visual localization, even in diverse and ultra-high-resolution UI scenarios.

\subsection{Layout Grounding}
Layout grounding requires models to localize and interpret the spatial relationships among elements in a document, such as titles, paragraphs, sections, and figures. This task is crucial for applications like document layout analysis and web page understanding, as it demands not only object detection but also reasoning over structural arrangement and semantic relationships.

\textbf{Benchmark, Evaluation Setting, and Metrics:}
We evaluate our model on the DocLayNet~\cite{pfitzmann2022doclaynet} and M6Doc~\cite{cheng2023m6doc} datasets. DocLayNet is collected from PDF documents and includes 11 categories such as footnotes, pictures, tables, and titles, with a test set consisting of 6,480 images.The M6Doc dataset is considerably more complex, encompassing data from diverse domains (e.g., scientific articles, textbooks, test papers, magazines, newspapers, notes, books) with a total of 74 categories across 2,724 test images. For evaluation, we treat this task as an object detection problem, following the same evaluation protocol used for common object detection on COCO.

\begin{figure}[t]     
    \centering
    \includegraphics[width=\textwidth]{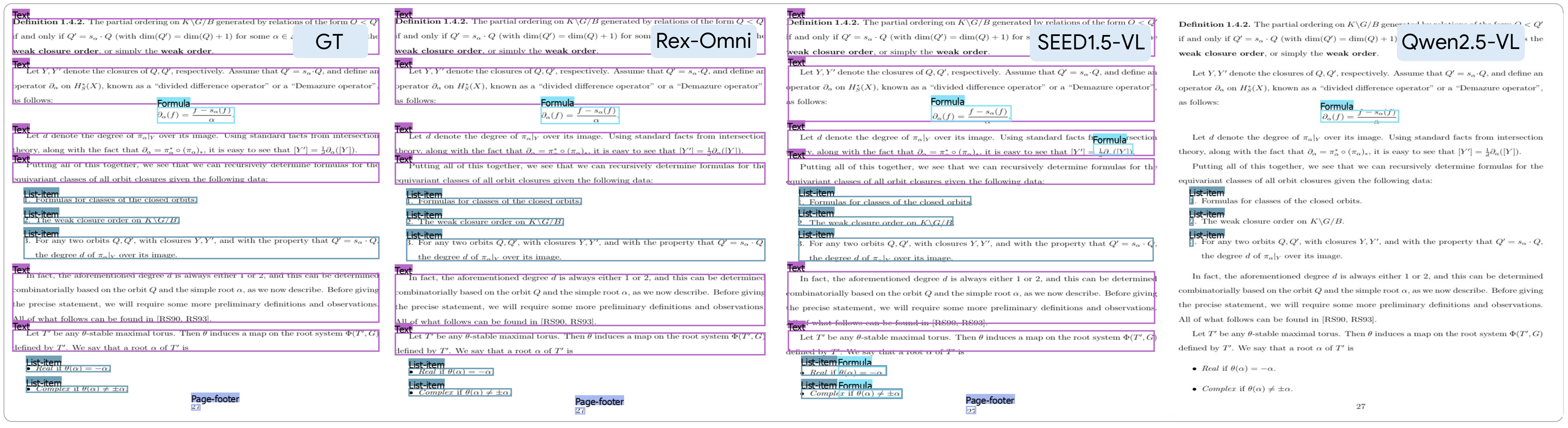}
    \caption{Qualitative comparison of layout grounding predictions from different models. The figure illustrates the models' ability to localize and interpret various layout elements.}
    \label{fig:compare_layout_grounding}
\end{figure}

\textbf{Results:}
The results are presented in Table~\ref{tab:layout_grounding}. Rex-Omni outperforms other MLLMs by a large margin on layout grounding. While there remains a performance gap compared to closed-set models, Rex-Omni’s ability to handle open-set layout grounding provides a unique advantage. Unlike closed-set models, which are limited to predefined categories, Rex-Omni demonstrates a unique capability to generalize to unseen domains and novel layout structures, establishing it as a more versatile and adaptable solution for real-world layout understanding tasks. Representative visualization results are provided in Figure \ref{fig:compare_layout_grounding} and Figure~\ref{fig:app_layout}.

\begin{table*}[t]
    \centering
    \resizebox{0.9\textwidth}{!}{
        \begin{tabular}{cc|c|ccc|ccc}
\midrule
\multirow{2}{*}{Type} & \multirow{2}{*}{Method} & \multirow{2}{*}{\begin{tabular}[c]{@{}c@{}}Score \\ Thresh.\end{tabular}} & \multicolumn{3}{c|}{DocLayNet} & \multicolumn{3}{c}{M6Doc} \\ \cline{4-9} 
 &  &  & \begin{tabular}[c]{@{}c@{}}F1@IoU\\ 0.5\end{tabular} & \begin{tabular}[c]{@{}c@{}}F1@IoU\\ 0.95\end{tabular} & \begin{tabular}[c]{@{}c@{}}F1@IoU\\ mIoU\end{tabular} & \begin{tabular}[c]{@{}c@{}}F1@IoU\\ 0.5\end{tabular} & \begin{tabular}[c]{@{}c@{}}F1@IoU\\ 0.95\end{tabular} & \begin{tabular}[c]{@{}c@{}}F1@IoU\\ mIoU\end{tabular} \\ \midrule
Closed-Set & DocLayout-YOLO~\cite{zhao2024doclayout} & 0.3 & \textbf{91.2} & \textbf{52.1} & \textbf{81.1} & - & - & - \\ \midrule
\multirow{4}{*}{MLLM}  & Qwen2.5-VL-3B & - & 17.5 & 2.9 & 9.1 & 13.3 & 2.5 & 8.4 \\
 & Qwen2.5-VL-7B & - & 25.6 & 5.1 & 13.4 & 24.0 & 4.1 & 15.0 \\
 & SEED1.5-VL & - & 54.9 & 4.3 & 28.7 & 48.0 & 3.4 & 28.0 \\
 \rowcolor{lightblue!10} & Rex-Omni-SFT & - & 85.9 & 27.2 & 70.7 & {\underline{74.5}} & {\underline{16.2}} & {\underline{54.2}} \\
 \rowcolor{lightblue!10} & Rex-Omni &  & {\underline{89.5}} & {\underline{28.4}} & {\underline{70.7}} & \textbf{76.3} & \textbf{18.7} & \textbf{55.6} \\ \midrule
\end{tabular}
    }
    \centering
    \caption{Performance comparison of different models on the DocLayNet and M6Doc datasets for layout grounding. F1-scores (at IoU=0.5, IoU=0.95, mIoU) are reported.}
    \label{tab:layout_grounding}
\end{table*}

\begin{table*}[t]
    \centering
    \resizebox{1.0\textwidth}{!}{
        \begin{tabular}{cc|ccc|ccc|ccc|ccc}
\toprule
\multirow{2}{*}{\begin{tabular}[c]{@{}c@{}}Output\\ Format\end{tabular}} & \multirow{2}{*}{Method} & \multicolumn{3}{c|}{HierText} & \multicolumn{3}{c|}{ICDAR2015} & \multicolumn{3}{c|}{TotalText} & \multicolumn{3}{c}{SROIE} \\
\cline{3-14}
 &  & \begin{tabular}[c]{@{}c@{}}F1@IoU\\ 0.5\end{tabular} & \begin{tabular}[c]{@{}c@{}}F1@IoU\\ 0.95\end{tabular} & \begin{tabular}[c]{@{}c@{}}F1@IoU\\ mIoU\end{tabular} & \begin{tabular}[c]{@{}c@{}}F1@IoU\\ 0.5\end{tabular} & \begin{tabular}[c]{@{}c@{}}F1@IoU\\ 0.95\end{tabular} & \begin{tabular}[c]{@{}c@{}}F1@IoU\\ mIoU\end{tabular} & \begin{tabular}[c]{@{}c@{}}F1@IoU\\ 0.5\end{tabular} & \begin{tabular}[c]{@{}c@{}}F1@IoU\\ 0.95\end{tabular} & \begin{tabular}[c]{@{}c@{}}F1@IoU\\ mIoU\end{tabular} & \begin{tabular}[c]{@{}c@{}}F1@IoU\\ 0.5\end{tabular} & \begin{tabular}[c]{@{}c@{}}F1@IoU\\ 0.95\end{tabular} & \begin{tabular}[c]{@{}c@{}}F1@IoU\\ mIoU\end{tabular} \\
\midrule
\multirow{4}{*}{BBOX}
& PaddleOCRv5~\cite{cui2025paddleocr} & 45.2 & \textbf{3.4} & \textbf{30.5} & 38.2 & \textbf{1.2} & 25.6 & 40.2 & 0.7 & 25.7 & \textbf{77.7} & \textbf{5.6} & \textbf{58.6} \\
& SEED1.5-VL & 27.1 & 0.2 & 12.0 & \underline{38.6} & 0.0 & 18.7 & 35.0 & 0.3 & 19.5 & 51.9 & 0.8 & 28.1 \\

& \cellcolor{lightblue!10}Rex-Omni-SFT & \cellcolor{lightblue!10}23.5 & \cellcolor{lightblue!10}0.5 &\cellcolor{lightblue!10} 13.7 & \cellcolor{lightblue!10}31.4 &\cellcolor{lightblue!10} 0.1 & \cellcolor{lightblue!10}18.7 & \cellcolor{lightblue!10}38.1 & \cellcolor{lightblue!10}\underline{1.5} & \cellcolor{lightblue!10}25.0 & \cellcolor{lightblue!10}46.5 & \cellcolor{lightblue!10}0.9 & \cellcolor{lightblue!10}28.6 \\
& \cellcolor{lightblue!10}Rex-Omni & \cellcolor{lightblue!10}\textbf{45.9} & \cellcolor{lightblue!10}1.4 & \cellcolor{lightblue!10}28.0 & \cellcolor{lightblue!10}\textbf{45.2} & \cellcolor{lightblue!10}0.3 & \cellcolor{lightblue!10}\textbf{28.1} & \cellcolor{lightblue!10}\textbf{56.6} & \cellcolor{lightblue!10}\textbf{3.9} & \cellcolor{lightblue!10}\textbf{40.6} & \cellcolor{lightblue!10}72.0 & \cellcolor{lightblue!10}\underline{1.5} & \cellcolor{lightblue!10}44.8 \\
\midrule
\multirow{3}{*}{POLY}
& PaddleOCRv5 & 41.5 & \underline{1.1} & \textbf{26.3} & 36.4 & 0.2 & 23.3 & 34.1 & 0.0 & 18.4 & 70.5 & \textbf{2.3} & \textbf{50.2} \\
& \cellcolor{lightblue!10}Rex-Omni-SFT &\cellcolor{lightblue!10} \textbf{43.2} & \cellcolor{lightblue!10}\underline{0.3} &\cellcolor{lightblue!10} 26.2 & \cellcolor{lightblue!10}43.2 & \cellcolor{lightblue!10}\textbf{0.3} & \cellcolor{lightblue!10}26.2 & \cellcolor{lightblue!10}50.3 & \cellcolor{lightblue!10}0.1 & \cellcolor{lightblue!10}\textbf{25.7} & \cellcolor{lightblue!10}\textbf{73.8} & \cellcolor{lightblue!10}0.2 & \cellcolor{lightblue!10}39.7 \\
& \cellcolor{lightblue!10}Rex-Omni &\cellcolor{lightblue!10} 40.2 & \cellcolor{lightblue!10}0.1 & \cellcolor{lightblue!10}20.2 & \cellcolor{lightblue!10}\textbf{50.7} & \cellcolor{lightblue!10}0.0 & \cellcolor{lightblue!10}\textbf{28.5} & \cellcolor{lightblue!10}\textbf{52.8} & \cellcolor{lightblue!10}\textbf{0.1} & \cellcolor{lightblue!10}25.6 & \cellcolor{lightblue!10}60.3 & \cellcolor{lightblue!10}0.0 & \cellcolor{lightblue!10}19.2 \\
\bottomrule
\end{tabular}
    }
    \caption{Performance comparison of various models on the OCR task, evaluated using F1 score for text detection and recognition accuracy.}
    \label{tab:ocr}
\end{table*}

\begin{figure}[t]     
    \centering
    \includegraphics[width=\textwidth]{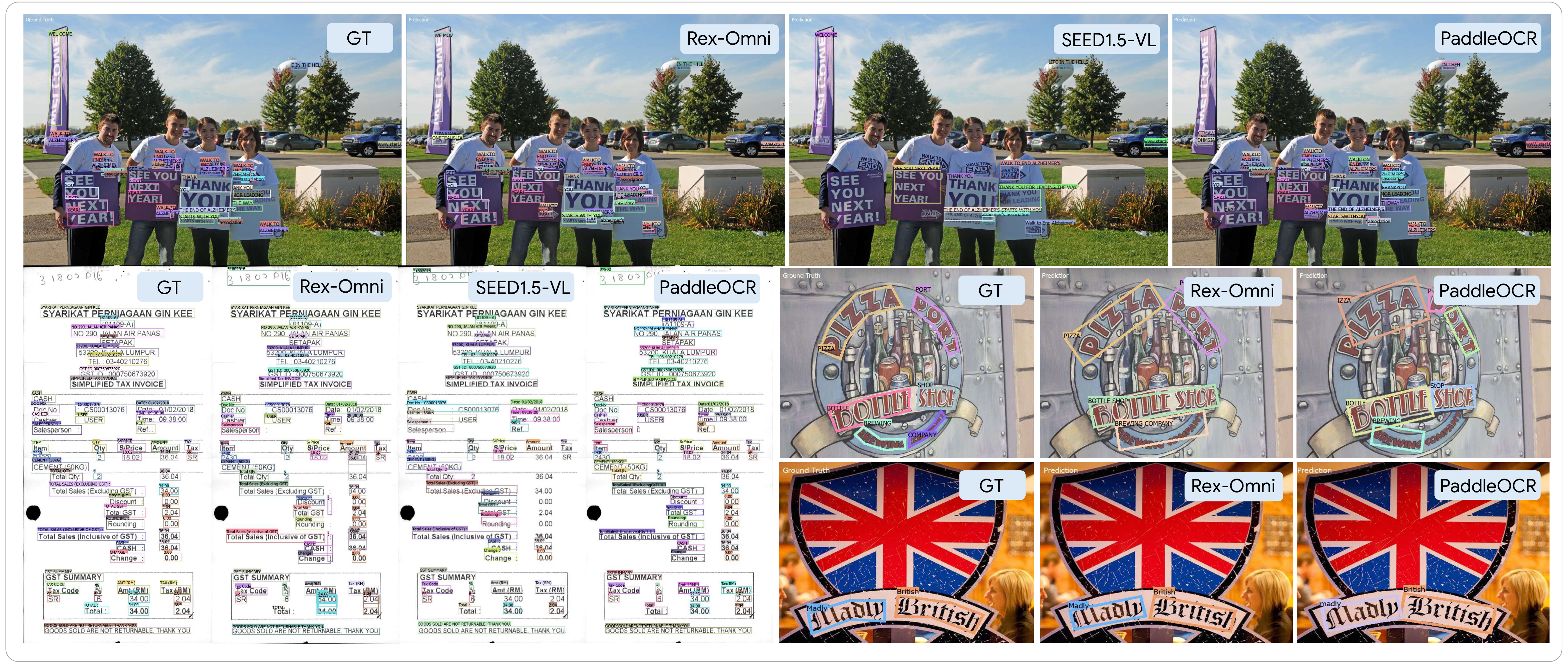}
    \captionsetup{justification=centering}
    \caption{Visualization of OCR results across models.}
    \label{fig:compare_ocr}
\end{figure}

\subsection{OCR}
Optical Character Recognition (OCR) involves both text detection and recognition, where the model identifies and extracts text from images or documents. The task requires the model to detect text regions and then recognize the characters or words within those regions, enabling the conversion of scanned documents or images into machine-readable text.

\textbf{Benchmark, Evaluation Setting:}
We evaluate the performance of PaddleOCR, SEED1.5-VL, and Rex-Omni on four diverse datasets. The core evaluation method involves detecting and recognizing all text within the images. The datasets include HierText (3,446 instances, primarily dense text), TotalText (600 instances, scene text with predominantly curved text), ICDAR2015 (1,000 instances, scene text), and SROIE (720 instances, printed receipt data with mostly horizontal text). Together, these datasets cover a broad spectrum of OCR challenges, from dense and curved scene text to structured document text. For PaddleOCR and Rex-Omni, both bounding box (BBOX) and polygonal (POLY) text regions are predicted, and we report performance for both formats to provide a comprehensive assessment of text localization.

\textbf{Metrics:} We formulate OCR as an object detection task, following the COCO evaluation protocol with categories replaced by recognized text. A prediction is considered correct if (1) the predicted and ground-truth regions match, and (2) the recognized text exactly matches the ground truth. Performance is reported using the F1 score, balancing precision and recall.

\textbf{Results:} The evaluation results for the OCR task are presented in Table \ref{tab:ocr}. For bounding box (BBOX) outputs, Rex-Omni demonstrates strong competitive performance. It significantly outperforms SEED1.5-VL across all metrics and datasets, and achieves comparable or superior results to the dedicated OCR expert model PaddleOCRv5 in several key aspects. This highlights Rex-Omni's robust capabilities in text detection and recognition using bounding boxes. In the polygonal (POLY) output format, Rex-Omni also shows competitive performance. The full Rex-Omni model, after GRPO post-training, notably achieves leading results on challenging datasets like ICDAR2015 for polygonal text region detection. This indicates the versatility of our approach in handling more complex text geometries. Consistent gains from Rex-Omni-SFT to Rex-Omni further validate the effectiveness of our two-stage training pipeline in enhancing OCR performance. Representative visualization results are provided in Figure \ref{fig:compare_ocr} and and Figure~\ref{fig:app_ocr}.

\begin{table}[!t]
\centering
\renewcommand{\arraystretch}{1.2}
\resizebox{1.0\textwidth}{!}{
\begin{tabular}{c|c|ccccc|cc}
\toprule
Multi Subtask &
\multicolumn{1}{c|}{\textit{Proprietary Models}} &
\multicolumn{5}{c|}{\textit{Referring Specialist Models}} & \multicolumn{2}{c}{\textit{Our Models}}  \\
& Gemini-2.5-Pro~\cite{comanici2025gemini} & SpaceLLAVA~\cite{foutter2024space} & RoboPoint~\cite{yuan2024robopoint} & Molmo-7B~\cite{deitke2024molmo} & Molmo-72B~\cite{deitke2024molmo} & ReboRefer-2B$^{*}$~\cite{zhou2025roborefer} & \cellcolor{lightblue!10} Rex-Omini-SFT & \cellcolor{lightblue!10} Rex-Omini \\
\hline
\textit{Location} & 46.96 & 5.82 & 22.87 & 21.91 & 45.77 & 51.00 &  \cellcolor{lightblue!10} \textbf{55.00} & \cellcolor{lightblue!10} \underline{54.00} \\
\textit{Placement} & 24.21 & 4.31 & 9.27  & 12.85 & 14.74 & \underline{49.00} & \cellcolor{lightblue!10} 45.00 & \cellcolor{lightblue!10} \textbf{50.00} \\
\textit{Useen} & 27.14 & 4.02 & 8.40  & 12.23 & 21.24 & \textbf{38.96} & \cellcolor{lightblue!10}\underline{36.36} & \cellcolor{lightblue!10}\underline{36.36}  \\
\bottomrule
\end{tabular}
}
\caption{Comparison of different models on the RefSpatial benchmark. The value is the percentage (\%) of correct predictions. *: evaluated without depth prior.}
\label{tab:spatial}
\end{table}

\begin{figure}[t]     
    \centering
    \includegraphics[width=\textwidth]{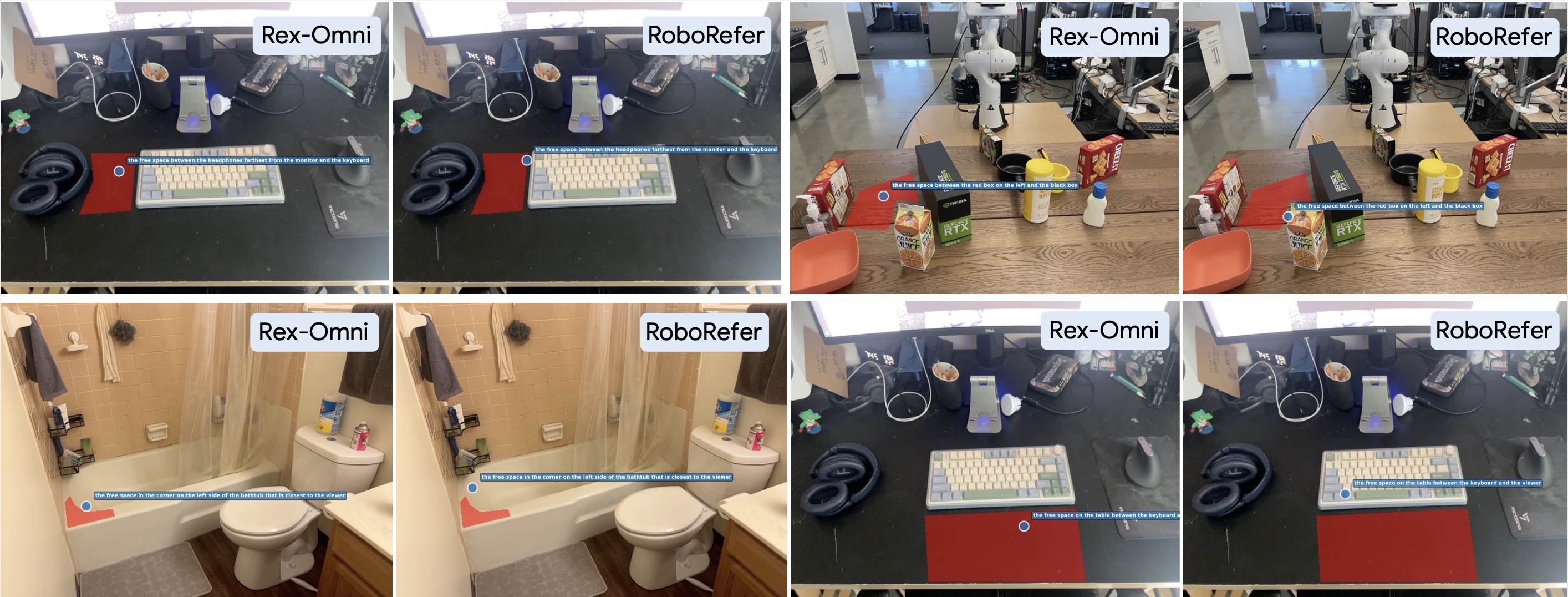}
    \captionsetup{justification=centering}
    \caption{Visualization of Spatial Pointing results across models. Masks indicate correct areas.}
    \label{fig:compare_spatial}
\end{figure}

\subsection{Spatial Pointing}

This task focuses on grounding natural language expressions that describe spatial relationships in complex scenes. Unlike standard referring object detection, which primarily matches objects to category names or simple attributes, spatial grounding requires models to interpret relational cues such as relative position, anchoring, and free-space placement.

\noindent\textbf{Benchmark and Metrics:}
RefSpatial-Bench~\cite{zhou2025roborefer} evaluates spatial referring and reasoning in complex indoor scenes across two tasks: \textit{location} and \textit{placement}, each with 100 curated samples. Each sample includes an image, a natural language referring expression, and precise mask annotations. The location task requires models to predict a 2D point corresponding to a target object based on referring expressions that may involve attributes such as color, shape, spatial order, or anchor-based references. The placement task requires identifying a suitable 2D point within a free space described in the expression, often involving multiple anchors or hierarchical spatial relations. To assess generalization, the benchmark additionally provides 77 unseen samples containing novel combinations of spatial relations not present in training. Evaluation is performed using ground-truth masks, with accuracy defined as the percentage of predictions falling within the mask region.

\noindent\textbf{Results:} As shown in Table~\ref{tab:spatial}, Rex-Omni substantially outperforms prior proprietary and referring-specialist models. Its strong performance on both location and placement tasks suggests enhanced applicability to downstream scenarios such as robotic manipulation, where accurate grasping and placement are essential. Moreover, Rex-Omni demonstrates superior generalization to unseen cases, underscoring its robustness in handling novel spatial relations. Representative visualization results are provided in Figure \ref{fig:compare_spatial}.

\subsection{Keypoint}

\noindent\textbf{Benchmark, Evaluation Setting, and Metrics:}
COCO is a benchmark dataset designed to evaluate 2D human pose estimation and instance-level keypoint detection capabilities in unconstrained environments. It comprises a large-scale collection of images featuring people in diverse and complex natural scenes. Each annotated person instance includes a set of 17 predefined body joints, forming a standard human skeleton. AP10K is a benchmark designed to advance the field of 2D animal pose estimation, addressing the challenge of anatomical variation across species. The benchmark standardizes keypoint annotation with a unified definition of 17 body keypoints for mammals, reptiles, and birds.
Following the COCO protocol, we adopt Object Keypoint Similarity (OKS) as the evaluation metric. We report F1 score at OKS thresholds of 0.5, 0.95, and the mean over thresholds from 0.5 to 0.95 in increments of 0.05.

\begin{table*}[t]
    \centering
    \resizebox{1.0\textwidth}{!}{
       \begin{tabular}{cc|c|ccc|ccc}
\toprule
\multirow{2}{*}{Type} & \multirow{2}{*}{Model} & \begin{tabular}[c]{@{}c@{}}Score\\ Thresh.\end{tabular} 
& \multicolumn{3}{c|}{COCO Keypoint} & \multicolumn{3}{c}{AP10K Keypoint} \\
\cline{4-9}
 &  &  & \begin{tabular}[c]{@{}c@{}}F1@OKS\\ 0.5\end{tabular} & \begin{tabular}[c]{@{}c@{}}F1@OKS\\ 0.95\end{tabular} & \begin{tabular}[c]{@{}c@{}}F1@OKS\\ mOKS\end{tabular} 
 & \begin{tabular}[c]{@{}c@{}}F1@OKS\\ 0.5\end{tabular} & \begin{tabular}[c]{@{}c@{}}F1@OKS\\ 0.95\end{tabular} & \begin{tabular}[c]{@{}c@{}}F1@OKS\\ mOKS\end{tabular} \\
\midrule
Open-set & X-Pose & \begin{tabular}[c]{@{}c@{}}0.3 (COCO) or\\ 0.05 (AP10K)\end{tabular} 
& \textbf{66.3} & \textbf{39.6} & \textbf{57.2} & 17.0 & 2.1 & 8.7 \\
\midrule
\multirow{2}{*}{MLLM} 
& Rex-Omni-SFT & - & 39.9 & 17.5 & 29.3 & 27.4 & 2.2 & 13.0 \\
& Rex-Omni & - & 44.4 & 17.9 & 32.6 & \textbf{30.1} & \textbf{3.0} & \textbf{14.6} \\
\bottomrule
\end{tabular}
    }
    \captionsetup{justification=centering}
    \caption{Evaluation results for keypoint estimation on COCO and AP10K datasets.}
    \label{tab:keypointing}
\end{table*}

\begin{figure}[t]     
    \centering
    \includegraphics[width=\textwidth]{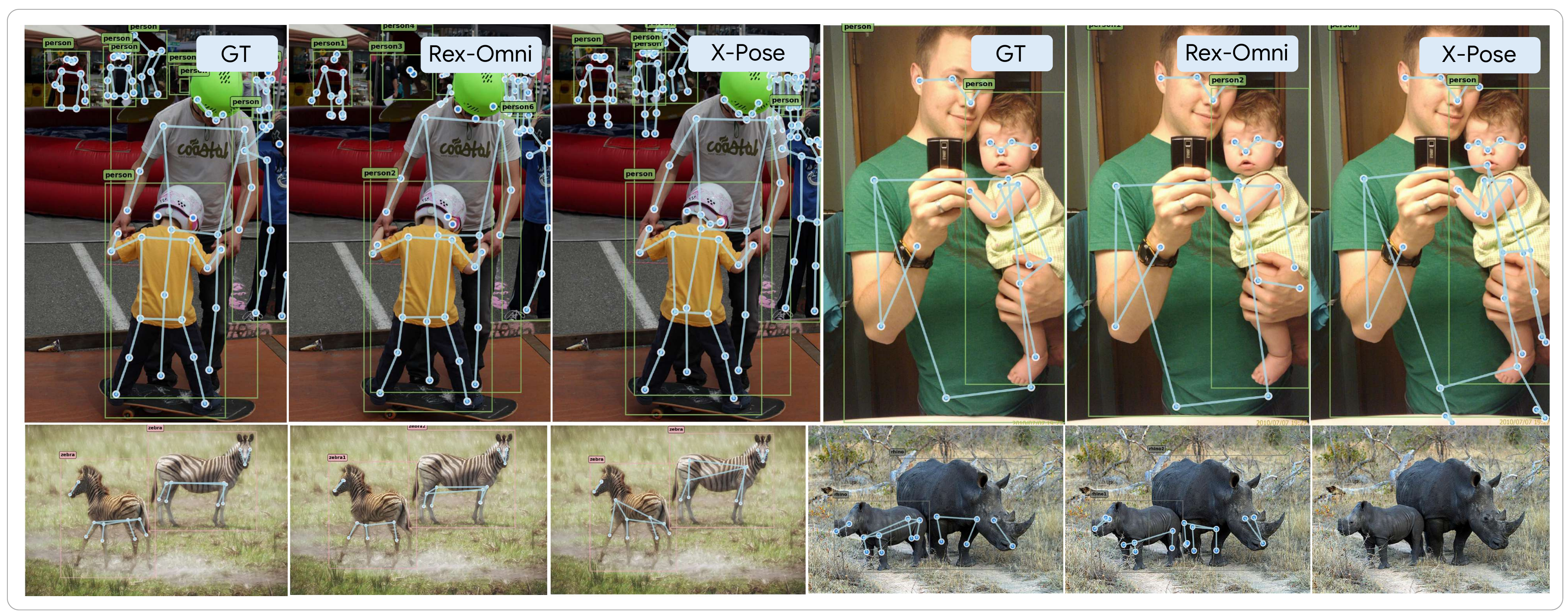}
    \captionsetup{justification=centering}
    \caption{Qualitative comparison of keypoint detection predictions from different models.}
    \label{fig:keypoint}
\end{figure}

\noindent\textbf{Results:} 
As shown in Table~\ref{tab:keypointing}, the open-set expert model X-Pose achieves the strongest performance on COCO keypoint detection, particularly at lower OKS thresholds. However, it generalizes poorly to AP10K, where its performance drops sharply. In contrast, Rex-Omni demonstrates more balanced results across both human and animal keypoint benchmarks. Although its absolute scores on COCO lag behind X-Pose, Rex-Omni substantially outperforms it on AP10K, highlighting its superior cross-domain generalization. Furthermore, the consistent improvement from Rex-Omni-SFT to the full Rex-Omni model validates the effectiveness of our two-stage training pipeline for enhancing keypoint reasoning. Representative visualization results are provided in Figure \ref{fig:keypoint}.

\section{In-depth Analysis of Rex-Omni}

In this section, we conduct a comprehensive analysis to investigate and elucidate the efficacy of Rex-Omni's key design components. Our aim is to provide a deeper understanding of how each architectural choice, training strategy (including the role of GRPO), and data design collectively influences the model’s overall performance across various visual perception tasks.

\subsection{Why GRPO Works}
Rex-Omni adopts a two-stage training strategy, beginning with supervised fine-tuning (SFT) and followed by GRPO-based reinforcement learning. Across all coordinate prediction benchmarks, the GRPO-enhanced model consistently outperforms its SFT-only counterpart. To investigate the source of these gains, we analyze the model’s behaviors and highlight key error patterns that GRPO effectively mitigates.

\begin{figure}[t]     
    \centering
    \includegraphics[width=\textwidth]{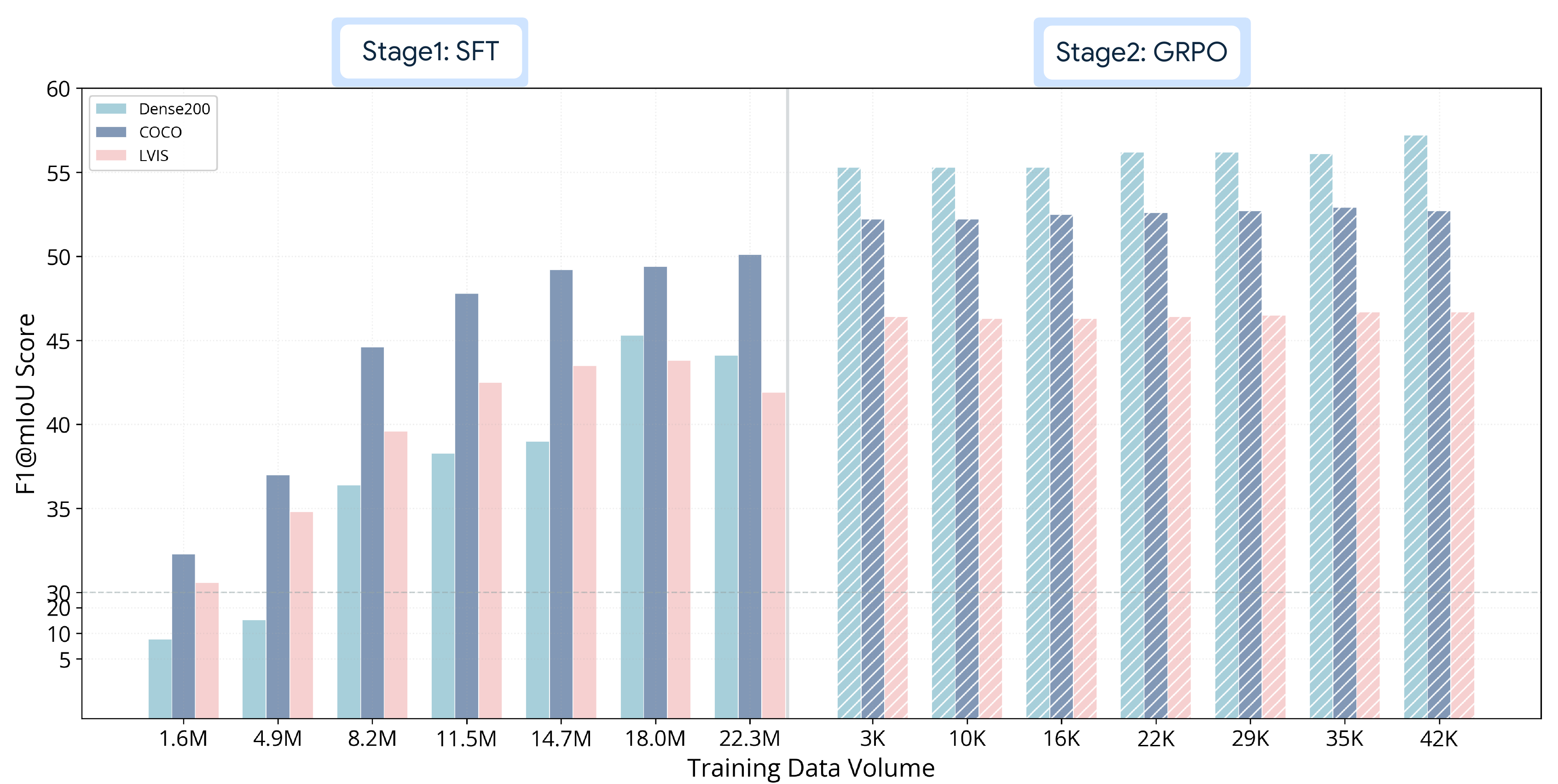}
    \caption{Model performance across SFT and GRPO training stages. F1@mIoU scores are shown for three datasets (Dense200, COCO, LVIS) with training data volume increasing.}
    \label{fig:training_dynamics}
\end{figure}

\subsubsection{Training Dynamics}
To better understand how Rex-Omni acquires its visual perception capability, we analyze the performance trajectory during both the SFT and GRPO stages as training progresses. Figure~\ref{fig:training_dynamics} illustrates model performance on representative benchmarks as a function of training steps (measured by the amount of data seen).

During the SFT stage, performance exhibits a steady and gradual improvement. As the model is exposed to more training data, it progressively learns to align visual inputs with coordinate outputs, leading to consistent but incremental gains across benchmarks. However, once SFT concludes, performance tends to plateau, suggesting that further improvements from additional supervised exposure are limited. In contrast, the GRPO stage produces a strikingly different trajectory. With only a small number of training steps, the model experiences a rapid performance jump across benchmarks. Notably, this improvement cannot be attributed simply to more data exposure, since the GRPO stage involves far fewer samples than SFT. Instead, the results suggest that the SFT-trained model already possesses strong latent capabilities that remain underutilized. GRPO, by introducing behavior-aware rewards and sequence-level feedback, effectively unlocks this hidden potential, enabling the model to achieve a substantial leap in performance with minimal additional training.

Taken together, these dynamics reveal that GRPO’s benefit lies not in extending supervised learning, but in reshaping model behavior to better exploit existing capabilities. In the following subsections, we delve deeper into the specific mechanisms behind this improvement, beginning with how GRPO corrects problematic behaviors learned during SFT.

\subsubsection{Behavioral Correction via GRPO}
\textbf{Duplicate Predictions}. A major error pattern is the tendency to generate repeated predictions. Under SFT, the model is trained with full teacher forcing, conditioning on ground-truth tokens at each step—so it rarely encounters or corrects this issue. In contrast, GRPO requires the model to generate sequences autonomously and provides reward-based feedback. Repeated coordinates receive low rewards, effectively discouraging duplication and promoting more coherent predictions.

\begin{table*}[t]
    \centering
    \resizebox{1.0\textwidth}{!}{
        \begin{tabular}{c|cccc|cccc|cccc}
\toprule
\multirow{3}{*}{\begin{tabular}[c]{@{}c@{}}Remove \\ Duplicate\end{tabular}} & \multicolumn{4}{c|}{COCO} & \multicolumn{4}{c|}{LVIS} & \multicolumn{4}{c}{VisDrone} \\ \cline{2-13} 
 & \multicolumn{2}{c|}{SFT} & \multicolumn{2}{c|}{GRPO} & \multicolumn{2}{c|}{SFT} & \multicolumn{2}{c|}{GRPO} & \multicolumn{2}{c|}{SFT} & \multicolumn{2}{c}{GRPO} \\ \cline{2-13} 
 & F1@0.5 & \multicolumn{1}{c|}{Remov.} & F1@0.5 & Remov. & F1@0.5 & \multicolumn{1}{c|}{Remov.} & F1@0.5 & Remov. & F1@0.5 & \multicolumn{1}{c|}{Remov.} & F1@0.5 & Remov. \\ \hline
No & 68.2 & \multicolumn{1}{c|}{-} & 72.0 & - & 60.3 & \multicolumn{1}{c|}{-} & 64.3 & - & 55.6 & \multicolumn{1}{c|}{-} & 61.6 & - \\ \hline
Yes & 70.1 & \multicolumn{1}{c|}{1.23\%} & 72.6 & 0.08\% & 61.3 & \multicolumn{1}{c|}{1.38\%} & 64.7 & 0.06\% & 62.3 & \multicolumn{1}{c|}{15.3\%} & 62.1 & 0.1\% \\ \bottomrule
\end{tabular}
    }
    \centering
    \caption{Performance comparison (F1@IoU=0.5) of SFT and GRPO models before and after removing duplicate predictions across COCO, LVIS, and VisDrone. This highlights GRPO's effectiveness in mitigating repetitive outputs and its impact on overall performance.}
    \label{tab:ab1_duplicate}
\end{table*}

\begin{figure}[t]     
    \centering
    \includegraphics[width=\textwidth]{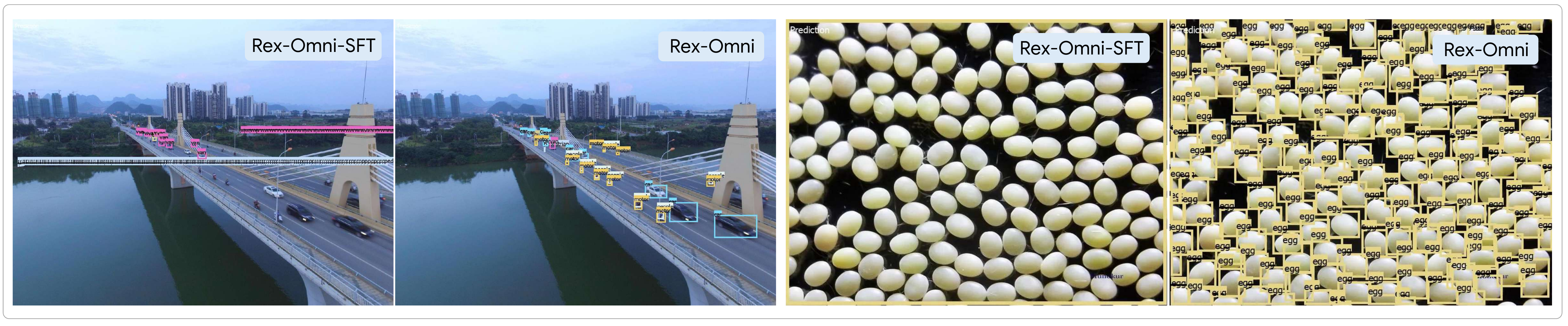}
    \caption{Illustration of challenges in MLLM-based object detection and their mitigation. The left side qualitatively compares predictions from SFT and GRPO models, highlighting GRPO's effectiveness in reducing duplicate outputs. The right side visualizes the large-box prediction failure mode, where models inappropriately predict overly large bounding boxes encompassing multiple objects.}
    \label{fig:ab1}
\end{figure}

To verify this effect, we analyzed predictions from both the SFT-only and GRPO-trained models, focusing on repeated outputs. A repeated prediction is defined as a coordinate sequence where the same value appeared consecutively at least 10 times, with the total number of predicted boxes exceeding twice the ground-truth count. We removed such duplicates and re-evaluated F1 scores. As shown in Table~\ref{tab:ab1_duplicate}, the SFT-only model exhibited substantial improvements after duplicate removal (e.g., +1.23\% on COCO, +1.38\% on LVIS, and +15.3\% on VisDrone), whereas the GRPO model showed minimal gains (e.g., +0.08\% on COCO, +0.1\% on VisDrone). This indicates that SFT-trained models produce significantly more repeated predictions than GRPO-trained models. After removing duplicates, the performance gap between SFT and GRPO narrowed, becoming nearly negligible on dense datasets like VisDrone. Visual examples of these differences are shown in Figure~\ref{fig:ab1} (left). These findings confirm that GRPO effectively suppresses duplicate predictions, which is a key factor in Rex-Omni’s overall performance improvement.

\noindent\textbf{Large-box Predictions}. Another behavioral issue observed, especially in dense object detection scenarios, is the tendency of models to predict a single large bounding box that encompasses multiple dense objects. This failure mode was also highlighted in our benchmarking of dense object detection (Section \ref{sec:dense}). To investigate this, we conducted an experiment on the Dense200 dataset. A large box prediction was defined as a scenario where only one bounding box was predicted in the image, and its area exceeded 95\% of the total image size. We then analyzed instances of such large box predictions from both the SFT-only and GRPO-trained models, removing these samples from evaluation. 
\begin{table*}[t]
    \centering
    \resizebox{0.8\textwidth}{!}{
        \begin{tabular}{c|cccccccc}
\toprule
\multirow{3}{*}{\begin{tabular}[c]{@{}c@{}}Remove \\ Large box\end{tabular}} & \multicolumn{8}{c}{Dense200} \\ \cline{2-9} 
 & \multicolumn{4}{c|}{SFT} & \multicolumn{4}{c}{GRPO} \\ \cline{2-9} 
 & \begin{tabular}[c]{@{}c@{}}F1@IoU\\ 0.5\end{tabular} & \begin{tabular}[c]{@{}c@{}}F1@IoU\\ 0.95\end{tabular} & \begin{tabular}[c]{@{}c@{}}F1@IoU\\ mIoU\end{tabular} & \multicolumn{1}{c|}{Remov.} & \begin{tabular}[c]{@{}c@{}}F1@IoU\\ 0.5\end{tabular} & \begin{tabular}[c]{@{}c@{}}F1@IoU\\ 0.95\end{tabular} & \begin{tabular}[c]{@{}c@{}}F1@IoU\\ mIoU\end{tabular} & Remov. \\ \hline
No & 59.1 & 8.8 & 44.9 & \multicolumn{1}{c|}{-} & 78.4 & 10.3 & 58.3 & - \\ \hline
Yes & 74.6 & 11.2 & 56.7 & \multicolumn{1}{c|}{20.5\%} & 81.8 & 9.4 & 60.0 & 3.5\% \\ \bottomrule
\end{tabular}
    }
    \centering
    \caption{Impact of large box prediction removal on F1-score (IoU=0.5, IoU=0.95, mIoU) for SFT and GRPO models on the Dense200 dataset. The "Remov." column indicates the percentage of large box predictions removed from the total outputs.}
    \label{tab:ab1_large_box}
\end{table*}

As shown in Table \ref{tab:ab1_large_box}, the SFT-only model exhibited a substantial 20.5\% of its total predictions as large boxes, leading to a significant performance improvement (e.g., F1@IoU=mIoU increased from 44.9 to 56.7) once these large box predictions were removed. In stark contrast, the GRPO-trained model had only 3.5\% of its predictions categorized as large boxes, and consequently, showed a much smaller performance change (F1@IoU=mIoU increased from 58.3 to 60.0) upon their removal. This clearly indicates that GRPO's behavior-aware optimization effectively discourages models from producing such overly large, encompassing bounding boxes for dense scenes. This failure mode is visually exemplified on the right side of Figure \ref{fig:ab1}.

\subsubsection{Improvement in Coordinate Precision?}
We hypothesize that the cross-entropy loss used in SFT lacks geometry-awareness, whereas GRPO can exploit geometry-aware rewards to refine coordinate precision. To validate this, we evaluate coordinate precision on COCO, LVIS, and HumanRef.

Specifically, for each test sample, We only include samples where both the SFT and GRPO models produce a number of predicted boxes that exactly matches the ground-truth count. Furthermore, for these selected samples, each predicted box from both models must achieve an IoU exceeding a predefined matching threshold with its corresponding ground-truth box. This filtering strategy allows us to effectively isolate the analysis to focus exclusively on the subtle differences in coordinate precision. As shown in Table~\ref{tab:ab3_coordinate_accuracy}, GRPO yields only modest gains over SFT. For example, F1@mIoU increases slightly from 63.0 to 63.5 on COCO and from 56.6 to 56.9 on LVIS. These results suggest that SFT already provides sufficient capacity for learning accurate coordinates and tight localization. Thus, GRPO’s primary advantage lies not in boosting raw coordinate precision, but in correcting behavioral deficiencies such as duplicate predictions and large-box outputs, as discussed earlier.

\begin{table*}[t]
    \centering
    \resizebox{1.0\textwidth}{!}{
        \begin{tabular}{c|ccc|ccc|ccc}
\hline
\multirow{2}{*}{Stage} & \multicolumn{3}{c|}{COCO}                                                                                                                                            & \multicolumn{3}{c|}{LVIS}                                                                                                                                            & \multicolumn{3}{c}{HumanRef}                                                                                                                                         \\ \cline{2-10} 
                       & \begin{tabular}[c]{@{}c@{}}F1@IoU\\ 0.5\end{tabular} & \begin{tabular}[c]{@{}c@{}}F1@IoU\\ 0.95\end{tabular} & \begin{tabular}[c]{@{}c@{}}F1@IoU\\ mIoU\end{tabular} & \begin{tabular}[c]{@{}c@{}}F1@IoU\\ 0.5\end{tabular} & \begin{tabular}[c]{@{}c@{}}F1@IoU\\ 0.95\end{tabular} & \begin{tabular}[c]{@{}c@{}}F1@IoU\\ mIoU\end{tabular} & \begin{tabular}[c]{@{}c@{}}F1@IoU\\ 0.5\end{tabular} & \begin{tabular}[c]{@{}c@{}}F1@IoU\\ 0.95\end{tabular} & \begin{tabular}[c]{@{}c@{}}F1@IoU\\ mIoU\end{tabular} \\ \hline
SFT                    & 80.5                                                 & 23.5                                                  & 63.0                                                  & 74.2                                                 & 33.0                                                  & 56.6                                                  & 85.2                                                 & 67.2                                                  & 60.0                                                  \\ \hline
GRPO                   & 81.1                                                 & 23.3                                                  & 63.5                                                  & 75.0                                                 & 32.9                                                  & 56.9                                                  & 86.4                                                 & 68.0                                                  & 61.2                                                 \\ \hline
\end{tabular}
    }
    \centering
    \caption{Impact of GRPO on coordinate precision. The table reports F1-scores (at IoU=0.5, IoU=0.95, and mIoU) for SFT and GRPO models across COCO, LVIS and HumanRef datasets, specifically for instances where both models achieve consistent ground-truth matching. This analysis highlights GRPO's modest contribution to refining coordinate precision.}
\label{tab:ab3_coordinate_accuracy}
\end{table*}

\begin{table*}[t]
    \centering
    \resizebox{1.0\textwidth}{!}{
        \begin{tabular}{c|ccc|ccc|ccc}
\hline
\multirow{2}{*}{Method} & \multicolumn{3}{c|}{COCO} & \multicolumn{3}{c|}{LVIS} & \multicolumn{3}{c}{Dense200} \\ \cline{2-10} 
 & \begin{tabular}[c]{@{}c@{}}F1@IoU\\ 0.5\end{tabular} & \begin{tabular}[c]{@{}c@{}}F1@IoU\\ 0.95\end{tabular} & \begin{tabular}[c]{@{}c@{}}F1@IoU\\ mIoU\end{tabular} & \begin{tabular}[c]{@{}c@{}}F1@IoU\\ 0.5\end{tabular} & \begin{tabular}[c]{@{}c@{}}F1@IoU\\ 0.95\end{tabular} & \begin{tabular}[c]{@{}c@{}}F1@IoU\\ mIoU\end{tabular} & \begin{tabular}[c]{@{}c@{}}F1@IoU\\ 0.5\end{tabular} & \begin{tabular}[c]{@{}c@{}}F1@IoU\\ 0.95\end{tabular} & \begin{tabular}[c]{@{}c@{}}F1@IoU\\ mIoU\end{tabular} \\ \hline
SFT & 68.2 & 15.8 & 50.4 & 60.3 & 20.7 & 44.2 & 60.2 & 10.6 & 46.4 \\
GRPO & 72.0 & 15.9 & 52.9 & 64.3 & 20.7 & 46.9 & 78.4 & 10.3 & 58.3 \\
\multicolumn{1}{l|}{SFT-Sampling-Best} & 64.6 & 9.0 & 44.0 & 56.6 & 13.8 & 38.7 & 38.2 & 2.0 & 24.6 \\
SFT-Sampling-Vote & 72.6 & 16.8 & 54.0 & 59.8 & 14.8 & 41.3 & 50.6 & 3.9 & 34.7 \\ \hline
\end{tabular}
    }
    \centering
    \caption{Impact of GRPO on the likelihood of sampling correct predictions. This table compares the F1-scores (IoU=0.5, IoU=0.95, mIoU) of SFT, GRPO, SFT-Sampling-Best, and SFT-Sampling-Vote models across COCO, LVIS, and Dense200 datasets. It illustrates GRPO's role in enhancing the probability and inherent quality of correct outputs, and explores SFT's potential under various sampling strategies.}
    \label{tab:ab4_sampling}
\end{table*}

\subsubsection{\textbf{Elevating the Likelihood of Correct Predictions}}
Beyond behavioral correction and coordinate refinement, we examine GRPO’s impact from a sampling-probability perspective. We hypothesize that SFT models inherently possess the ability to generate accurate predictions, but their inference randomness reduces the likelihood of consistently sampling optimal outputs. GRPO, by contrast, leverages reward-guided exploration to increase this likelihood.

To empirically test this, we conducted high-temperature sampling experiments using the SFT model on COCO, LVIS, and Dense200. We simulated GRPO's rollout by sampling 8 candidate predictions per test instance (using temperature 1.2, top-k 50, top-p 0.99). From these, we derived two SFT-based metrics: \textbf{SFT-Sampling-Best}: The highest F1-score achieved across 8 independent full-dataset test runs of the SFT model. \textbf{SFT-Sampling-Vote}: For each test sample, the best prediction (highest F1-score against ground truth) is chosen from its 8 sampled outputs. These sample-wise best predictions are then aggregated for overall performance. This estimates SFT's maximal performance if optimal predictions were reliably selected at the sample level.

As shown in Table~\ref{tab:ab4_sampling}, on COCO the SFT-Sampling-Vote score (72.6 F1@0.5) exceeds both GRPO (72.0) and base SFT (68.2), indicating that SFT has a latent capacity for accurate predictions and GRPO mainly improves sampling consistency on simpler datasets. However, on LVIS and Dense200, neither SFT-Sampling-Best nor SFT-Sampling-Vote approaches GRPO’s performance, showing that for complex tasks GRPO plays a deeper role by enabling inherently more coherent and precise predictions. These findings suggest that GRPO’s benefits vary by task complexity: increasing sampling probability on simpler settings, and fundamentally enhancing prediction quality in more challenging ones.

\begin{table*}[t]
    \centering
    \resizebox{1.0\textwidth}{!}{
        \begin{tabular}{c|ccc|ccc}
\hline
 & \multicolumn{3}{c|}{COCO@100} & \multicolumn{3}{c}{Dense200@100} \\ \hline
Model & \begin{tabular}[c]{@{}c@{}}boxes/img\\ (Avg.)\end{tabular} & \begin{tabular}[c]{@{}c@{}}output tokens/img\\ (Avg.)\end{tabular} & \begin{tabular}[c]{@{}c@{}}tokens/box\\ (Avg.)\end{tabular} & \begin{tabular}[c]{@{}c@{}}boxes/img\\ (Avg.)\end{tabular} & \begin{tabular}[c]{@{}c@{}}output tokens/img\\ (Avg.)\end{tabular} & \begin{tabular}[c]{@{}c@{}}tokens/box\\ (Avg.)\end{tabular} \\ \hline
SEED1.5-VL & 4.2 & 631.0 & 148.8 & 73.1 & 5446.3 & 74.5 \\ \hline
Rex-Omni & 5.9 & 45.3 & 7.6 & 86.7 & 439.0 & 5.1 \\ \hline
\end{tabular}
    }
    \centering
    \caption{Comparison of output tokenization efficiency between SEED1.5-VL and Rex-Omni. This table reports the average number of boxes per image, output tokens per image, and tokens per box on 100 randomly sampled images from COCO and Dense200, highlighting Rex-Omni's superior token efficiency.}
    \label{tab:ab5_inference_efficiency}
\end{table*}

\begin{figure}[t]     
    \centering
    \includegraphics[width=0.6\textwidth]{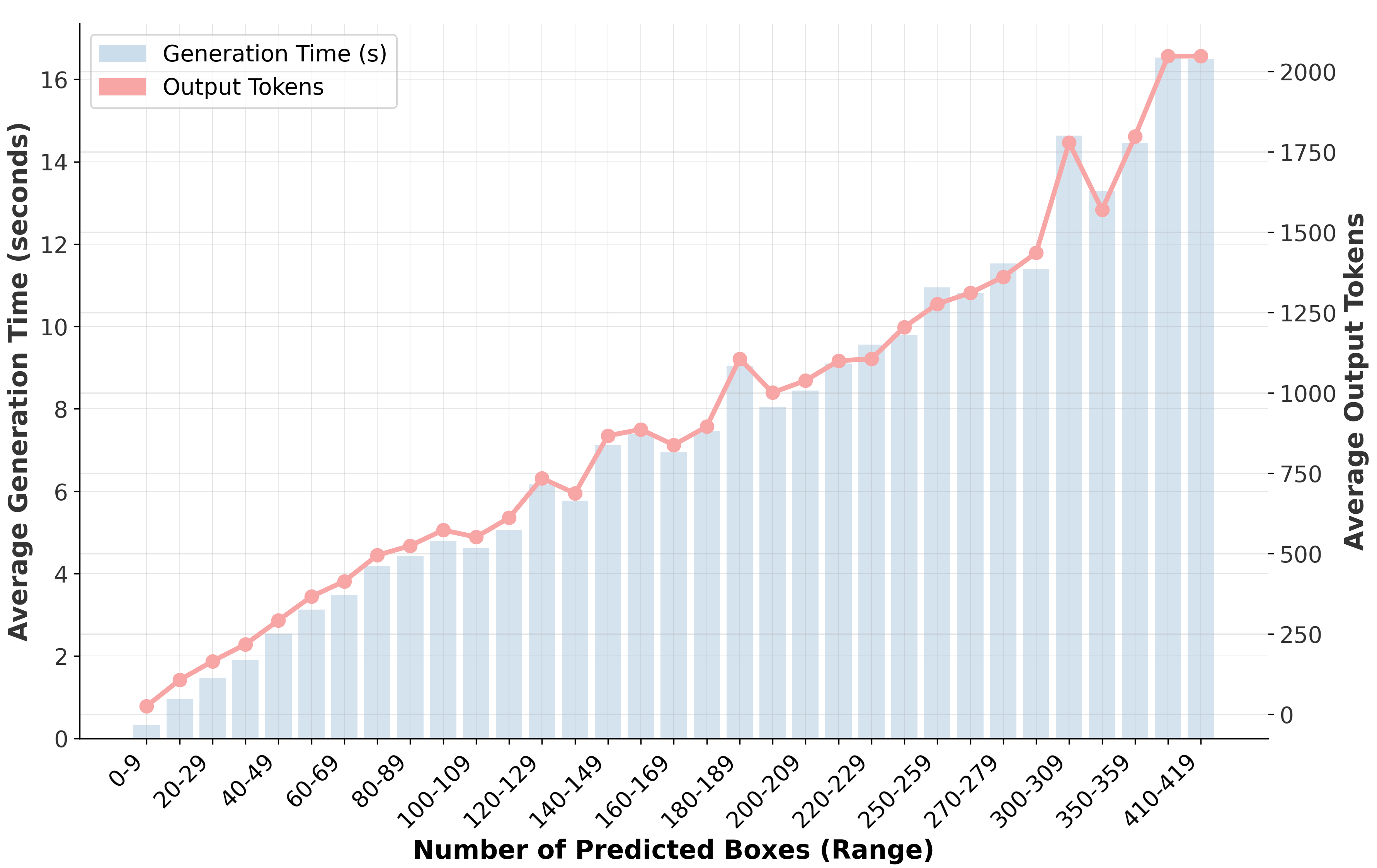}
    \caption{Inference speed and output token analysis. The plot shows the average generation time (seconds) and the average number of output tokens as a function of the number of predicted boxes. The experiment was conducted on a single NVIDIA A100 GPU with vLLM deployment in BF16 precision, without model acceleration or compression.}
    \label{fig:inference_speed}
\end{figure}

\subsection{Inference Efficiency and Speed}
The efficiency of coordinate representation is crucial, as it directly affects output length and inference speed. We compare Rex-Omni, which encodes quantized coordinates using special tokens, with SEED1.5-VL, which represents relative coordinates without special tokens. To evaluate this, we sampled 100 images each from COCO and Dense200, and measured the average boxes per image, total output tokens per image, and tokens per box. As summarized in Table~\ref{tab:ab5_inference_efficiency}, Rex-Omni achieves far greater tokenization efficiency. For example, on COCO it requires only 7.6 tokens per box on average, compared to 148.8 for SEED1.5-VL, with total output length reduced from 631.0 to 45.3 tokens per image. Similar improvements are observed on Dense200, confirming that dedicated special tokens substantially enhance efficiency, especially in dense object settings.

Beyond tokenization efficiency, we further examine practical inference speed. Figure~\ref{fig:inference_speed} illustrates the relationship between the number of predicted boxes, output token length, and average generation time, measured on a single NVIDIA A100 GPU using vLLM with BF16 precision (no acceleration or compression applied). Both generation time and token count grow approximately linearly with the number of predicted boxes: detecting a few objects (0–29) takes under 2 seconds, whereas detecting hundreds of objects (e.g., 410–419) exceeds 16 seconds. These findings indicate that current MLLM-based detectors are slower than traditional optimized detectors, with speed scaling directly with the number of detected objects. Nevertheless, this limitation could be mitigated through acceleration strategies such as quantization or distillation.
\section{Conclusion}
\label{sec:Conclusion}

In this work, we have introduced Rex-Omni, a 3B-parameter MLLM that systematically addresses the challenges of MLLM-based object detection. Through efficient coordinate tokenization with special tokens, large-scale data generation via custom engines, and a novel SFT+GRPO two-stage training pipeline, we bridge the gap between precise localization and deep language understanding. Our extensive experiments demonstrate that Rex-Omni achieves state-of-the-art or highly competitive zero-shot performance across a wide array of visual perception tasks. Crucially, our analysis validates that while SFT provides a strong foundation, GRPO-based post-training is essential for correcting SFT-induced behavioral deficiencies, such as duplicate and large-box predictions, a key contribution towards robust MLLM-based detectors. Despite its strong performance, limitations such as inference speed remain. We believe that future work in model acceleration and advanced reward-guided sampling will be critical next steps. In summary, Rex-Omni represents a significant step forward, demonstrating that the behavioral and geometric limitations of MLLMs can be systematically overcome, thereby paving the way for the next generation of versatile, language-aware perception systems.
\section{Related Work}
\label{sec:related_work}

\textbf{Regresson-based Object Detection Methods.}
Object detection has long been a cornerstone task in computer vision, with regression-based methods historically dominating the field. The core principle of these methods is to predict a bounding box by regressing its properties typically including center coordinates (x, y) and dimensions (width, height) as normalized offsets from a predefined reference.
Over the years, these methods have undergone significant evolution, progressing from early anchor-based CNN models like YOLO~\cite{redmon2016you}, SSD~\cite{liu2016ssd}, and Faster R-CNN~\cite{ren2016faster}, to anchor-free approaches such as CornerNet~\cite{law2018cornernet}, CenterNet~\cite{duan2019centernet}, and FCOS~\cite{tian2019fcos}. A major paradigm shift occurred with the introduction of Transformer-based detectors like DETR~\cite{carion2020end}, which framed object detection as a direct set prediction problem. This line of work was further advanced by models such as Deformable DETR~\cite{zhu2020deformable} and DINO~\cite{zhang2022dino}, which significantly improved performance and convergence speed.
Beyond these paradigmatic changes, the continuous improvement of regression-based detectors has been fueled by numerous incremental yet crucial innovations. These include architectural enhancements like Feature Pyramid Networks (FPN)\cite{lin2017feature}, advancements in loss functions like Focal Loss\cite{lin2017focal}, and sophisticated data augmentation techniques such as MixUp~\cite{zhang2017mixup} and Mosaic. It is the cumulative effect of these extensive and persistent efforts that has propelled regression-based object detectors to their current state of high performance and practical usability.

\textbf{Open-set Object Detection Methods.}
A long-term objective of object detection is to develop models capable of identifying an arbitrary number of object categories without task-specific fine-tuning, thereby addressing the challenges of real-world, dynamic scenarios. Open-set object detection represents a significant paradigm shift towards this goal, transcending the limitations of closed-set detection by empowering models to identify objects beyond a predefined set of categories. The prevalent approach to this challenge is text-prompted open-vocabulary object detection~\cite{li2022grounded, liu2023grounding, VLP:MDETR, yao2022detclip, zhong2022regionclip, ren2024dino, ghiasi2022scaling, cheng2024yolo, ma2023codet}. These methods typically leverage powerful pre-trained vision-language models like CLIP~\cite{VLP:CLIP} or BERT~\cite{devlin2018bert} to align textual descriptions with visual representations, demonstrating impressive zero-shot recognition capabilities. However, these models struggle with complex or nuanced descriptions due to their limited language understanding.
To overcome this, visual prompts~\cite{jiang2025t, jiang2023t, ren2024dino, wang2025yoloe, ravi2024sam, TransF:SAM, zou2023segment, li2024segment} have been introduced, allowing models to recognize objects using visual examples like boxes or points. Visual prompts are effective for rare or hard-to-describe objects but are less general than text prompts. Recent models like T-Rex2~\cite{jiang2025t} combine both text and visual prompts, using contrastive learning to leverage the strengths of each. This integration allows models to perform well across a wider range of object categories and real-world scenarios. While traditional open-set detectors achieve category-level generalization, they still lack deeper language understanding, making it challenging to handle context-rich real-world scenarios.

\textbf{MLLM-based Object Detection Methods.}
To overcome the shallow language understanding of traditional open-set detectors, a promising direction is to directly leverage the powerful reasoning capabilities of Multimodal Large Language Models (MLLMs) for object-level perception. The core idea is to reframe object detection as a language modeling task. Inspired by Pix2Seq~\cite{chen2021pix2seq}, a significant body of work has emerged that represents bounding box coordinates as a sequence of discrete, quantized tokens~\cite{peng2023kosmos, chen2023shikra, you2023ferret, wang2023cogvlm, zhan2025griffon}. These models, including Kosmos-2, Shikra, Ferret, and CogVLM, directly generate coordinate sequences through the standard next-token prediction mechanism of LLMs. This approach elegantly unifies object detection with the native capabilities of language models. However, as discussed in our introduction, this conceptually elegant approach faces significant practical challenges. While MLLMs excel at high-level image understanding, they often struggle with the fine-grained spatial precision required for object detection. Existing methods frequently suffer from limitations such as low recall rates, coordinate drift, and spurious duplicate predictions. We posit that these issues stem from two fundamental challenges: the inherent difficulty of learning a precise mapping from discrete tokens to a continuous pixel space using cross-entropy loss, and the behavioral deficiencies induced by the teacher-guided nature of Supervised Fine-Tuning (SFT). Addressing these challenges is the primary motivation for the design of Rex-Omni.

\bibliography{main}
\clearpage
\appendix
\clearpage
\appendix
\section{Appendix}

\subsection{More Visualization Results}

\begin{figure}[!h]     
    \centering
    \includegraphics[width=0.8\textwidth]{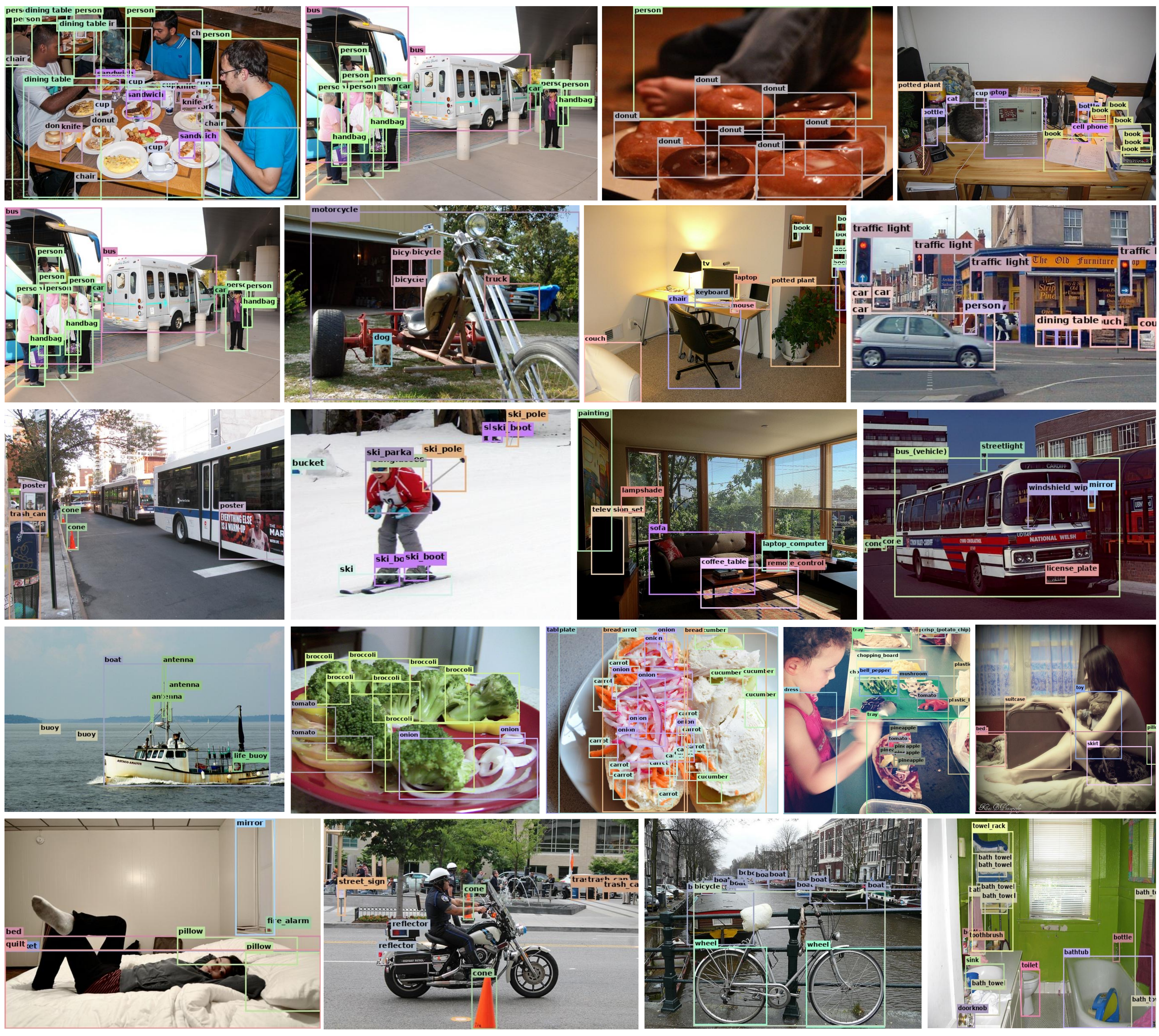}
    \captionsetup{justification=centering}
    \caption{Visualization results of Rex-Omni on common and long-tailed object detection task.}
    \label{fig:app_common_longtailed}
\end{figure}

\begin{figure}[t]     
    \includegraphics[width=1.0\textwidth]{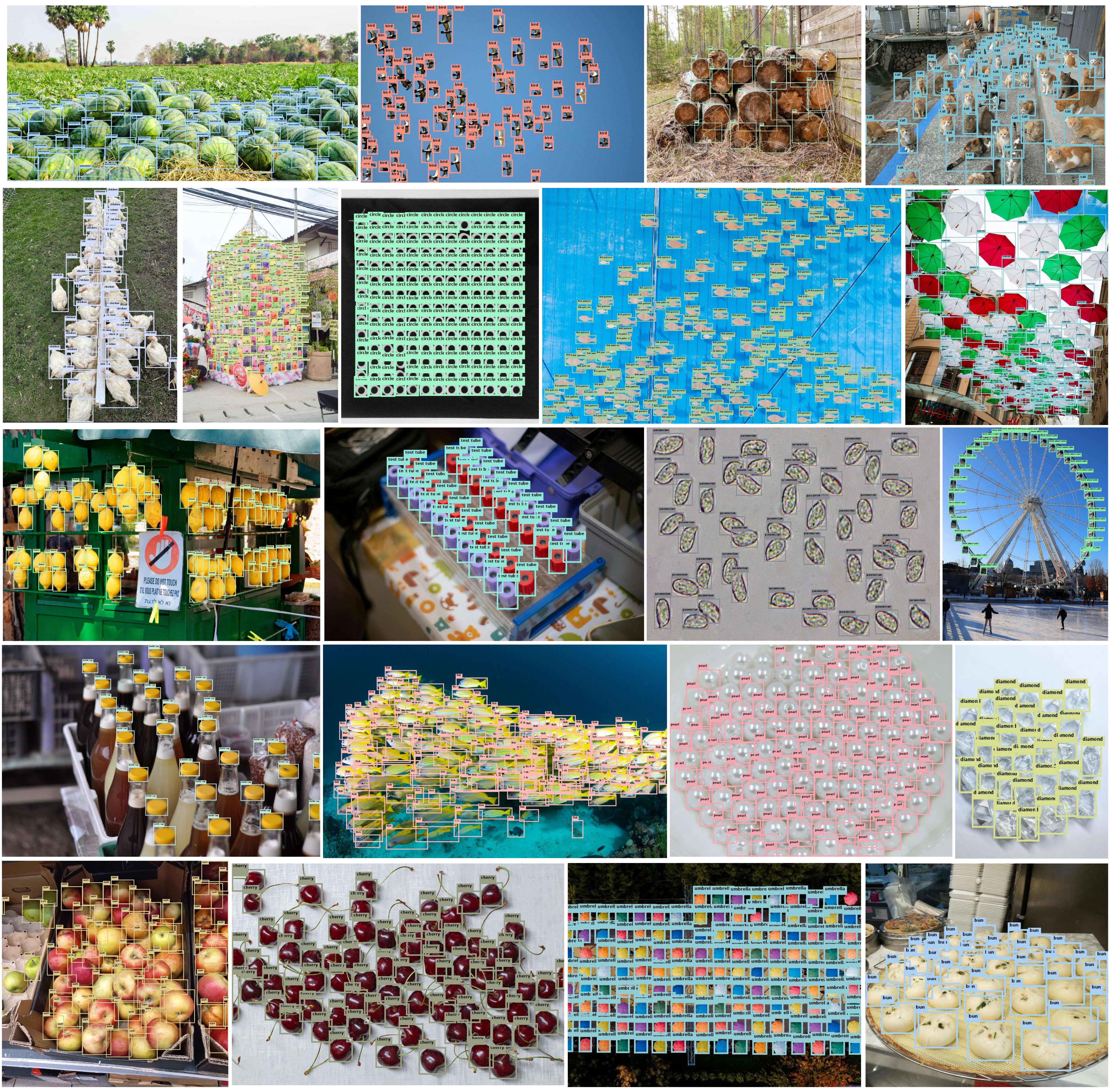}
    \captionsetup{justification=centering}
    \caption{Visualization results of Rex-Omni on dense object detection task.}
    \label{fig:app_dense}
\end{figure}

\begin{figure}[t]     
    \includegraphics[width=1.0\textwidth]{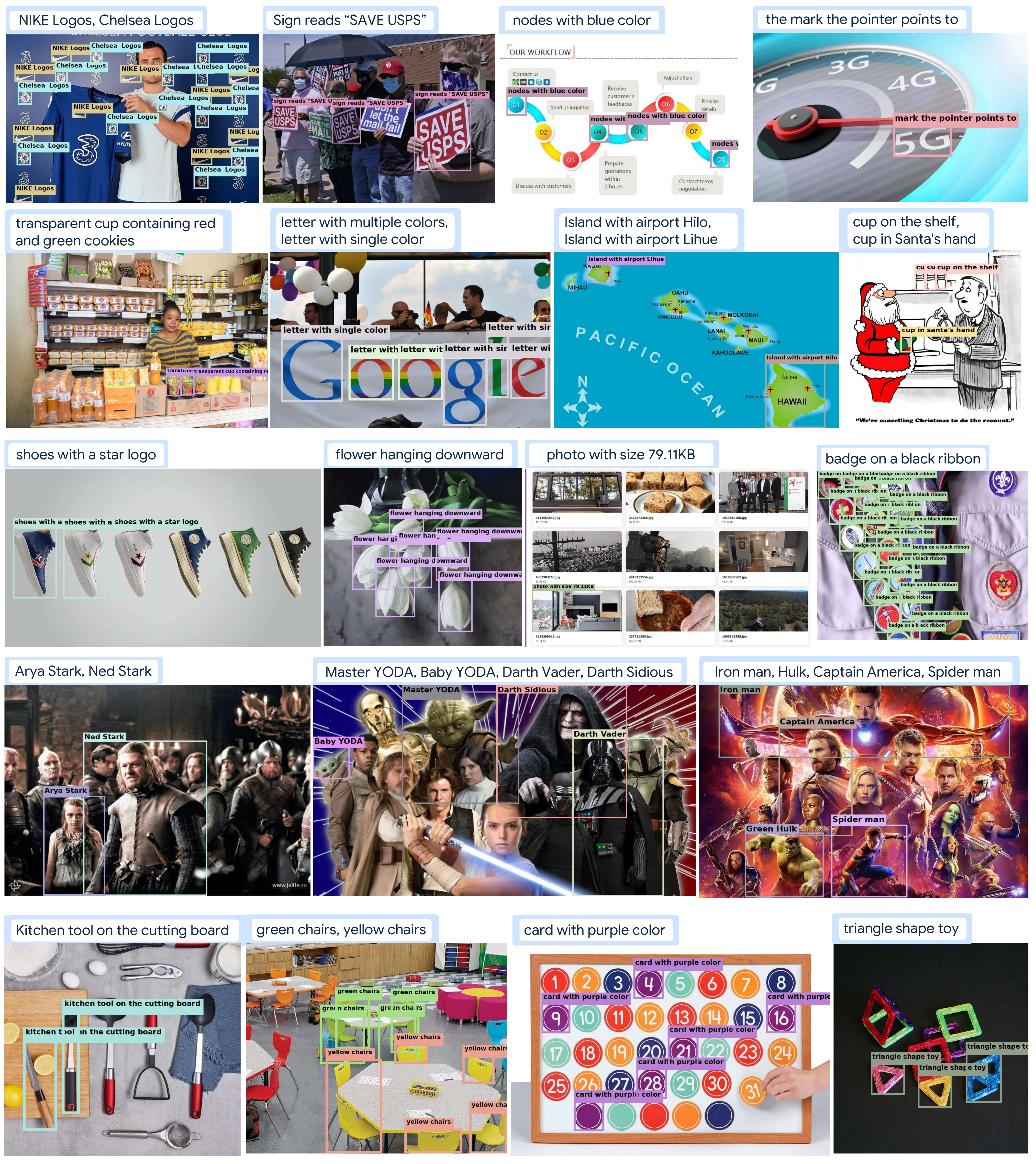}
    \captionsetup{justification=centering}
    \caption{Visualization results of Rex-Omni on object referring task.}
    \label{fig:app_referring}
\end{figure}

\begin{figure}[t]     
    \includegraphics[width=1.0\textwidth]{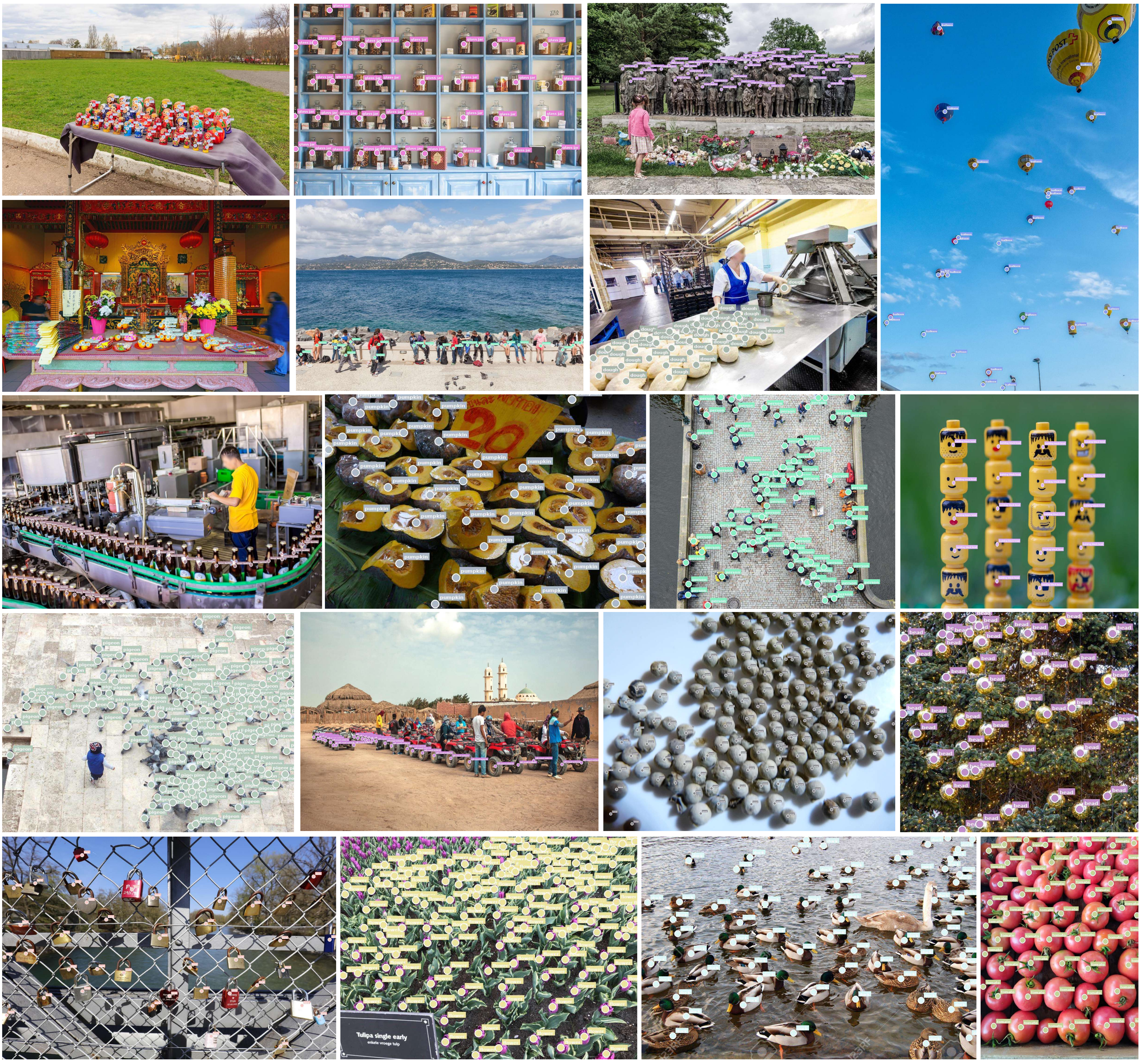}
    \captionsetup{justification=centering}
    \caption{Visualization results of Rex-Omni on object pointing task.}
    \label{fig:app_pointing}
\end{figure}

\begin{figure}[t]     
    \includegraphics[width=1.0\textwidth]{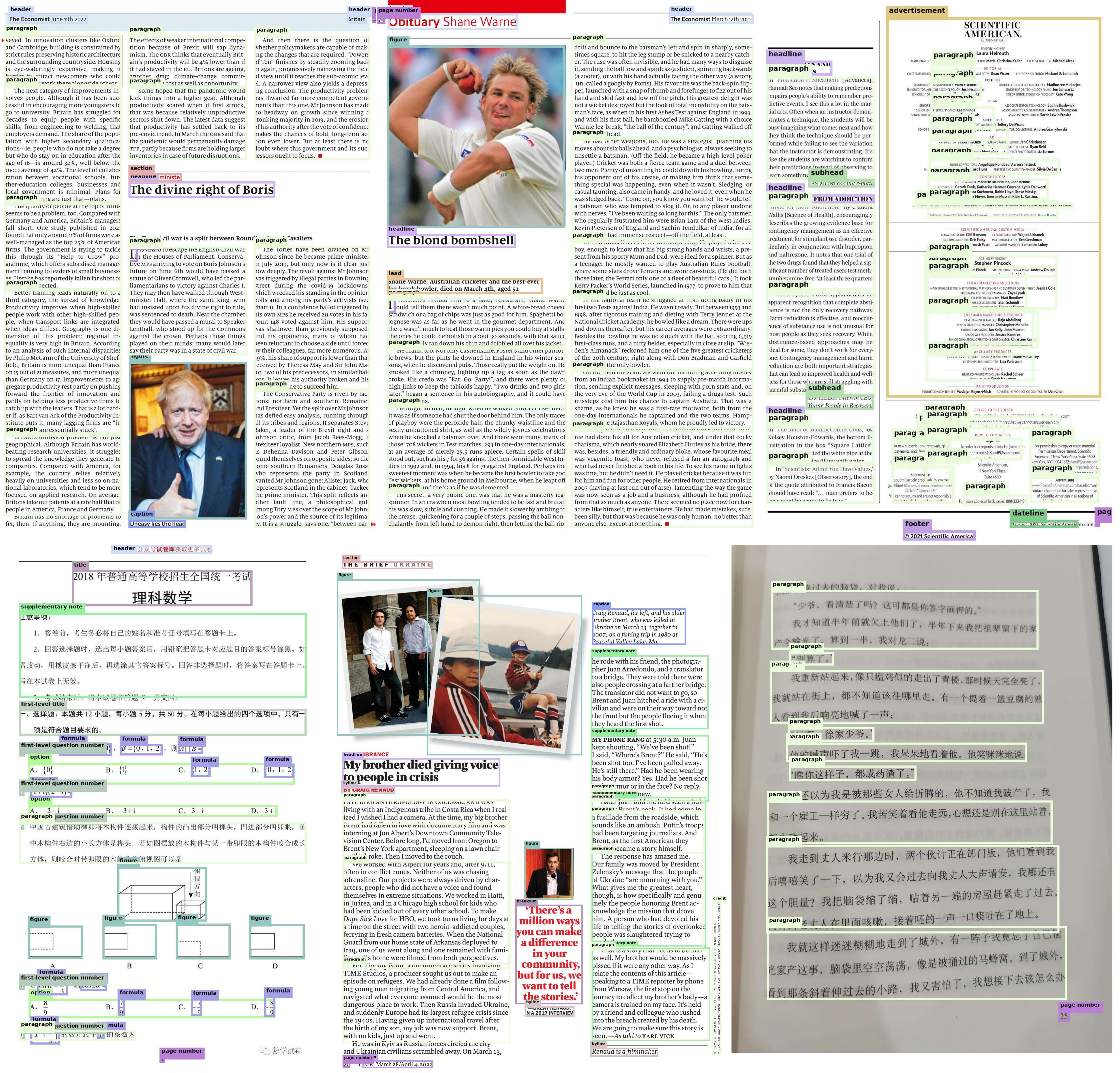}
    \captionsetup{justification=centering}
    \caption{Visualization results of Rex-Omni on layout grounding task.}
    \label{fig:app_layout}
\end{figure}

\begin{figure}[t]     
    \includegraphics[width=1.0\textwidth]{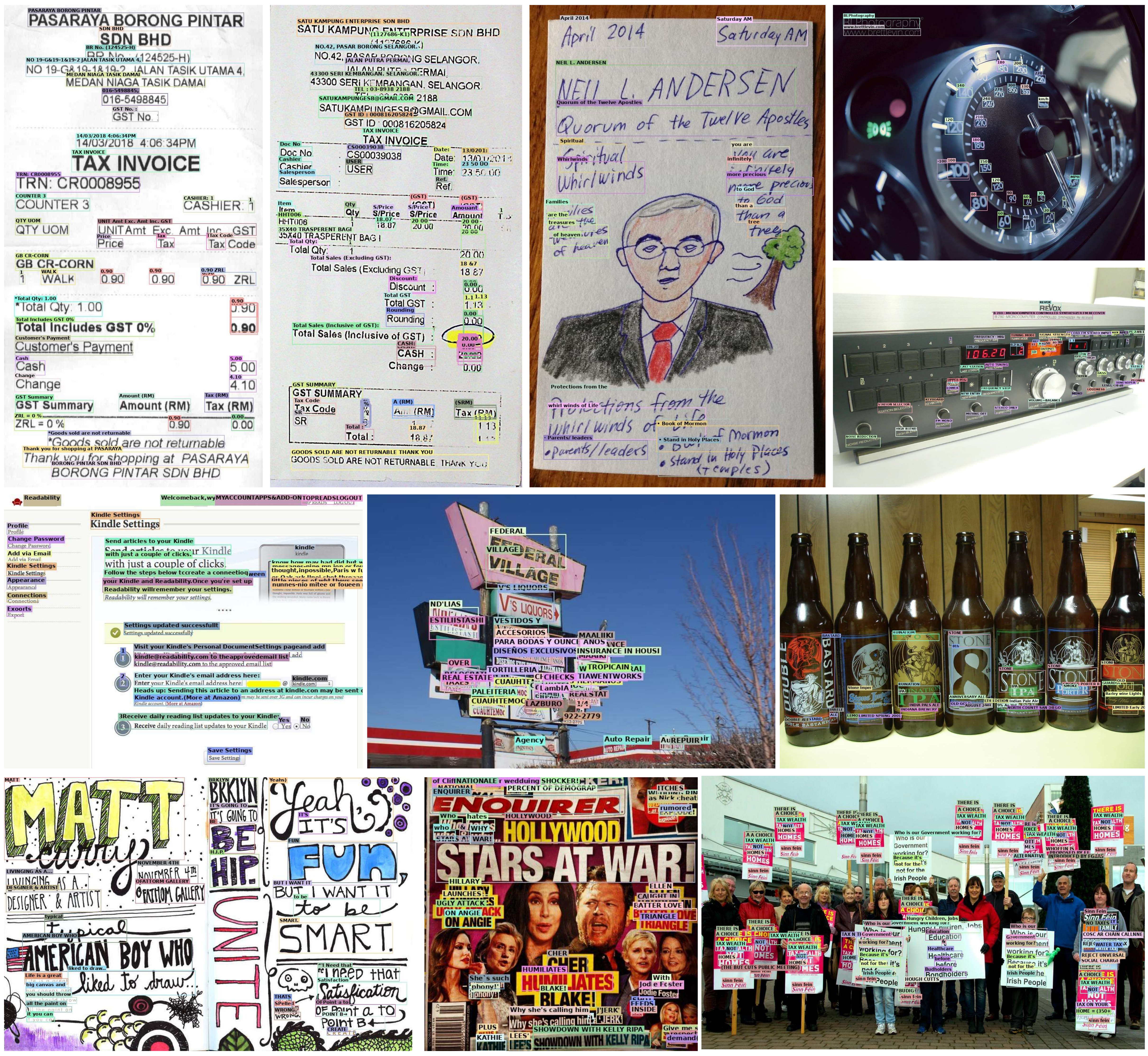}
    \captionsetup{justification=centering}
    \caption{Visualization results of Rex-Omni on OCR task.}
    \label{fig:app_ocr}
\end{figure}

To provide a more comprehensive and intuitive understanding of Rex-Omni's capabilities, this section presents additional qualitative results across a wide range of visual perception tasks. These visualizations complement the quantitative results reported in the main paper, offering further insights into the model's performance in diverse and challenging scenarios. We showcase more visualization results for the following tasks:
\begin{itemize}
    \item Common and Long-tailed Object Detection (Figure \ref{fig:app_common_longtailed})
    \item Dense Object Detection (Figure \ref{fig:app_dense})
    \item  Object Referring (Figure \ref{fig:app_referring})
    \item  Object Pointing (Figure \ref{fig:app_pointing})
    \item Layout Grounding (Figure \ref{fig:app_layout})
    \item OCR (Optical Character Recognition) (Figure \ref{fig:app_ocr})
\end{itemize}

\end{document}